%% file: tpami.tex
\documentclass[10pt,twocolumn,twoside,journal,compsoc]{IEEEtran}

\makeatletter
\def\normalsize{\@setfontsize{\normalsize}{9.5bp}{12.00pt}}
\normalsize
\makeatother

\usepackage{graphicx}
\DeclareGraphicsExtensions{.pdf,.png,.jpg}
\graphicspath{{./}{./figs/}{./fig}}

\usepackage{amsmath}
\usepackage{amssymb}
\usepackage{eucal}
\usepackage{amsthm}
\usepackage{amsfonts}

\usepackage[footnotesize,bf]{caption}
\usepackage{booktabs}

\usepackage{cite}

\usepackage{multirow}
\usepackage{dsfont}

\usepackage[labelfont=rm,font=scriptsize]{subcaption}

\usepackage{xspace,algorithmic,algorithm,amssymb}

\usepackage{float}

\usepackage[usenames,dvipsnames]{color}
\usepackage{hyperref}
\hypersetup{
  colorlinks,
  citecolor=Blue,
  linkcolor=Red,
  urlcolor=Blue}

\input{macros.tex}

\usepackage{bbm}

\def\oneslack{$ 1 $-slack\xspace}

\def\bm{\bf}

\def\Ind{\mathbbm{1}}

\def\paragraph{\subsection}

\def\argmin{\operatorname*{argmin\,}}

\newcommand{\svmstruct}{{SSVM}\xspace}
\newcommand{\ssvm}{{SSVM}\xspace}
\newcommand{\svm}{{SVM}\xspace}
\newcommand{\crf}{{CRF}\xspace}

\newcommand{\structboost}{StructBoost\xspace}
\newcommand{\nslack}{$m$-slack\xspace}
\newcommand{\mslack}{$m$-slack\xspace}

\newcommand{\U}{\mathbf{U}}
\newcommand{\V}{\mathbf{V}}

\let\phi\varphi

\usepackage[dvipsnames]{xcolor}
\definecolor{myb}{RGB}{10, 10, 240}

\begin{document}

\title{StructBoost: Boosting Methods for Predicting Structured Output Variables}

\author{
    Chunhua Shen,
    Guosheng Lin,
    Anton van den Hengel
\thanks{
        The authors are with
        The University of Adelaide,  Adelaide, SA 5005,  Australia.
        Correspondence should be addressed to C. Shen
        (e-mail: $\sf chunhua.shen@adelaie.edu.au$). This article was published in
        \textcolor{myb}
        {
        IEEE Transactions on Pattern Analysis and Machine Intelligence, vol.\ 36, no.\ 10, 2014; pages 2089--2103.
}
}
}

\IEEEcompsoctitleabstractindextext{
\begin{abstract}

        Boosting is a method for learning a single accurate predictor by linearly
        combining a set of less accurate weak learners. Recently, structured learning has
        found many applications in computer vision.
        Inspired by structured support vector machines (\svmstruct),
        here we propose a new boosting algorithm
        for structured output prediction, which we refer to as
        \structboost.  \structboost
        supports nonlinear structured
        learning by combining a set of weak structured learners.

        As
        \svmstruct generalizes \svm, our \structboost generalizes
        standard boosting
        approaches
        such as AdaBoost, or LPBoost to structured
        learning.
        The resulting optimization problem of \structboost is more
        challenging than \svmstruct in the sense that
        it may
        involve exponentially many variables and
        constraints. In contrast, for \svmstruct one usually has an
        exponential number of constraints and a cutting-plane method is
        used. In order to efficiently solve \structboost, we formulate
        an equivalent $ 1 $-slack formulation and solve it using a
        combination of cutting planes and column generation.
        We show the versatility and usefulness of \structboost on a
        range of
        problems such as optimizing the tree loss for hierarchical
        multi-class classification, optimizing the Pascal overlap criterion
        for robust visual tracking and learning conditional random
        field parameters for image segmentation.

\end{abstract}

}

\maketitle




\section{Introduction}

        Structured learning has attracted considerable attention
        in machine learning and computer vision in recent years
        (see, for example
        \cite{Torr2011ICCV,Nowozin2011Struct,Blaschko2008,Tsochantaridis2004}).
        Conventional supervised learning
        problems,
        such as classification and
        regression,
        aim to learn a
        function that predicts the best value for a response variable
        $y \in \Real$ for an input
        vector $\x \in \Real^d $
        on the basis of a set of example
        input-output pairs. In many applications, however, the outputs are
        often complex and cannot be well represented by a
        single
        scalar
        because the classes may have inter-class dependencies, or
        the most appropriate outputs
        are objects (vectors, sequences, trees,
        etc.).
        Such
        problems are referred to as {\em structured
        output prediction}. Structured support vector machines
        (\svmstruct) \cite{Tsochantaridis2004}
        generalize the multi-class \svm of \cite{Weston1999Multi} and
        \cite{Crammer2001Algorithmic} to the much broader problem of
        predicting interdependent and structured outputs. \svmstruct
        uses discriminant functions that take advantage of the
        dependencies and structure of outputs.
        In \svmstruct, the general form of the learned discriminant
        function is $ F(\x, \y; \w) : \calX \times \calY \mapsto \Real $
        over input-output pairs and the prediction is achieved
        by maximizing $ F (\x, \y; \w)$ over all possible $ \y \in
        \calY$.
        Note that to introduce non-linearity,
        the discriminant function can be defined by an
        {\em implicit} feature mapping function that is only
        accessible
        as a particular
        inner product in
        a reproducing kernel Hilbert space.
        This is the so-called kernel trick.

        On the other hand, boosting algorithms linearly combine a set
        of moderately accurate weak learners to form a nonlinear
        strong predictor, whose prediction performance is usually highly accurate.
        Recently, Shen and Hao \cite{Shen2011Direct} proposed a direct formulation for
        multi-class boosting using the loss functions of
        multi-class {\svm}s \cite{Weston1999Multi,Crammer2001Algorithmic}.
        Inspired by the general boosting framework of
        Shen and Li \cite{Shen2011Totally}, they implemented multi-class boosting
        using column generation.
        Here we go further by generalizing multi-class boosting
        of Shen and Hao to broad structured output prediction
        problems. \structboost
        thus enables
        nonlinear structured learning by
        combining a set of weak structured learners.

        The effectiveness of \ssvm has been limited by the fact that
        only the linear kernel is typically used.  This limitation arises largely as
        a result of the computational expense of training and applying
        kernelized {\ssvm}s.
        Nonlinear kernels often
        deliver improved prediction accuracy over that of linear kernels, but at the cost
        of significantly higher memory requirements and computation time.
        This is particularly the case when
        the training size is
        large, because the number of support vectors is linearly
        proportional to the size of training data
        \cite{Steinwart2003}.
        Boosting, however, learns models which
        are much faster to
        evaluate.  Boosting can also select relevant features
        during the course of learning by using particular weak
        learners such as decision stumps or decision trees, while
        almost all nonlinear kernels are defined on the entire feature
        space.
        It thus remains difficult (if not impossible) to see
        how kernel methods can select/learn explicit
        features.  For boosting, the learning
        procedure also selects or induces relevant features. The final
        model learned by boosting  methods are
        thus  often significantly simpler and
        computationally cheaper.
        In this sense, the proposed \structboost possesses  the
        advantages of both nonlinear \ssvm and boosting methods.

    \subsection{Main contributions}
        The main contributions of this work are three-fold.
    \begin{itemize}
    \item[1.]

        We propose \structboost, a new fully-corrective boosting method
        that combines a set of weak structured
        learners for predicting a broad range of structured outputs.
        We also discuss special cases of this general structured learning
        framework, including multi-class classification, ordinal
        regression, optimization of complex measures such as the
        Pascal image overlap criterion and conditional random field
        (CRF) parameters learning for image segmentation.

    \item[2.]

        To implement \structboost, we adapt the efficient
        cutting-plane method---originally designed for efficient  linear \svm
        training
        \cite{JoachimsSVM}---for our purpose. We equivalently
        reformulate the $ m$-slack optimization to $ 1$-slack
        optimization.

    \item[3.]

        We apply the proposed \structboost to a
        range of
        computer vision
        problems
        and show that  \structboost can indeed
        achieve state-of-the-art performance in some of the key problems in the field.
        In particular, we demonstrate a state-of-the-art
        object tracker trained by \structboost.
        We also demonstrate an application for CRF and
        super-pixel based image segmentation,
        using \structboost
        together with graph cuts for
        CRF parameter learning.

    \end{itemize}

        Since
        \structboost builds upon the fully corrective
        boosting of Shen and Li \cite{Shen2011Totally},
        it inherits the desirable properties of column generation
        based boosting, such as a fast convergence rate and
        clear explanations from the primal-dual convex optimization
        perspective.

        \subsection{Related work}

        The two state-of-the-art structured learning methods are CRF \cite{Lafferty01Conditional}
        and \svmstruct \cite{Tsochantaridis2004}, which capture the interdependency among
        output variables.
        Note that CRFs formulate
        global training for structured
        prediction as a convex optimization problem. \svmstruct
        also
        follows this path but employs
        a different loss function (hinge loss) and optimization methods. Our \structboost
        is directly inspired by \svmstruct. \structboost can be seen as an extension of
        boosting methods to structured prediction. It therefore builds upon the
        column generation approach to
         boosting
         from
         \cite{Shen2011Totally} and the direct formulation for multi-class
        boosting
        \cite{Shen2011Direct}. Indeed, we show that the multi-class boosting of
        \cite{Shen2011Direct} is a special case of the general framework presented here.

        CRF and \svmstruct have been applied to various problems in machine learning and computer
        vision mainly because the learned models can easily integrate prior knowledge given a
        problem of interest. For example, the linear chain CRF has been widely used in natural language
        processing
        \cite{Lafferty01Conditional,crffnt}.
        \svmstruct takes the context into account
        using the joint feature maps over the input-output pairs, where features can be
        represented equivalently as in CRF \cite{JoachimsSVM}.
        CRF is particularly of interest
        in computer vision for its success in semantic image segmentation \cite{crf09icml}. A
        critical issue of semantic image segmentation is to integrate local and global features
        for the prediction of local pixel/segment labels. Semantical segmentation is achieved by
        exploiting the class information with a CRF model. \svmstruct can also be used for similar
        purposes as demonstrated in \cite{BertelliYVG11}. Blaschko and Lampert \cite{Blaschko2008}
        trained \svmstruct models to predict the bounding box of objects in a given image, by
        optimizing the Pascal bounding box overlap score. The work in \cite{Torr2011ICCV} introduced
        structured learning to real-time object detection and tracking, which also optimizes the
        Pascal box overlap score. \svmstruct has also been used to learn statistics that capture
        the spatial arrangements of various object classes in images \cite{Desai2011}. The trained
        model can then simultaneously predict a structured labeling of the entire image. Based
        on the idea of large-margin learning in \svmstruct, Szummer et al.\ \cite{SzummerKH08}
        learned optimal parameters of a CRF, avoiding tedious cross validation. The survey of
        \cite{Nowozin2011Struct}  provided a comprehensive review of structured learning and
        its application in computer vision.

        Dietterich et al.\ \cite{Dietterich2004TCR} learned the CRF energy functions using gradient
        tree boosting.
        There the functional gradient of the CRF {\em conditional likelihood} is calculated,
        such that a regression tree (weak learner) is induced as in gradient boosting.
        An ensemble of trees is produced by iterating this procedure.
        In contrast, we learn CRF within the {\em large-margin} framework,
        by generalizing the work of \cite{SzummerKH08,NowozinGL10} where CRF parameters are learned using
        SSVM.
        In our case, we do not require approximations such as pseudo-likelihood.
        Another relevant work is \cite{munoz09}, where
        Munoz et al.\ used the functional gradient boosting methodology to discriminatively
        learn max-margin
        Markov networks (M3N), as proposed by Taskar et al.\ \cite{TaskarGK03}.
        The random fields' potential functions are learned
        following gradient boosting \cite{Mason1999}.

        There are a few structured boosting methods in the literature.
        As we discuss here, all of them are based on gradient boosting,
        which are not as general as that which
        we propose here.
        Ratliff et al.\ \cite{Ratliff07boostingstructured,Ratliff2009AR}
        proposed
        a boosting-based approach for imitation learning based on structured prediction, called maximum
        margin planning (MMP).
        Their method is named as MMPBoost. To train MMPBoost, a demonstrated
        policy is provided as example behavior as the input, and the
        problem
        is to learn a
        function over features of the environment that produce policies with similar behavior.
        Although MMPBoost is structured learning in that the output is a vector, it differs
        from ours fundamentally.
        First, the optimization procedure of MMPBoost is not
        directly defined on the joint function $ F(\x, \y; \w ) $.
        Second, MMPBoost is based on
        gradient descent boosting \cite{Mason1999}, and  \structboost is built upon fully
        corrective boosting of Shen and Li \cite{Shen2011Totally,Shen2013Fully}.

        Parker et al.\ \cite{Parker2006}
        have also successfully applied gradient tree boosting to learning sequence alignment.
        Later, Parker \cite{Parker2007}
        developed a margin-based structured perceptron update and
        showed that it can incorporate general notions of
        misclassification cost as well as kernels.
        In these methods, the objective function typically consists of an exponential
        number of terms that correspond to  all possible pairs of $ (\y, \y') $.
        Approximation is made to make the computation of gradient tractable \cite{Parker2006}.
        Wang et al.\ \cite{Wang07simpletraining}
        learned a local predictor using standard
        methods, e.g., \svm, but then achieved improved structured classification
        by exploiting the influence of misclassified components
        after structured prediction, and iteratively re-training the
        local predictor.
        This approach is heuristic and it is more like a
        post-processing procedure---it does not directly optimize the
        structured learning objective.

\subsection{Notation}

        A bold lower-case letter ($ \boldsymbol u $, $ \boldsymbol v$) denotes
        a column vector.
        An element-wise inequality between two vectors or matrices
        such as
        ${\boldsymbol u} \geq {\boldsymbol v}$ means
        $ u_i \geq  v_i $ for all $i$.
        Let $\x$ be an input; $\y$ be an output and the input-output pair be $( \x, \y ) \in
 {\cal X} \times {\cal Y}$, with $ {\cal X} \subset \Real^d $. Unlike classification ($ {\cal Y}
 = \{1,2,\dots, k \} $) or regression ($ {\cal Y}
 \subset \Real$) problems,
 we
 are interested in the case where elements of $ \cal Y$ are {\it
 structured variables} such as vectors, strings, or graphs. Recall that the proposed \structboost is
 a structured boosting method, which combines a set of {\em weak structured learners }
 (or weak compatibility functions). We denote
 by ${\cal H}$ the set of weak structured learners.
 Note that ${\cal H}$ is typically very large, or even infinite.
 Each weak structured learner: $ \wstruct ( \cdot, \cdot ) \in
 {\cal H}$, is a function that maps an input-output pair $ ( \x, \y ) $ to a scalar value which measures
 the compatibility of the input and output. We define column vector $ \wstructs( \x, \y ) = [
 \wstruct_1 ( \x, \y ), \cdots, \wstruct_n(\x, \y) ]^\T$ to be the outputs of all weak structured
 learners.
 Thus $ \wstructs( \x, \y ) $ plays the same sole as the joint mapping vector in \svmstruct, which relates
 input $ \x $ and output $ \y $.
        The form of a weak structured learner is
        task-dependent.
        We show some examples of $\wstruct(\cdot, \cdot)$ in Section~\ref{sec:app}.
        The discriminant function that we aim to learn is
        $ F: \calX \times \calY \mapsto \Real $, which measures the compatibility over input-output
        pairs.
        It has the form of
        \begin{equation}
        F( \x, \y; \w  )  =  \w^\T  \wstructs(\x, \y) = {\textstyle \sum}_j w_j \wstruct_j( \x, \y ),
        \label{EQ:1}
        \end{equation}
        with $ \w \geq  0 $.
        As in other structured learning models,
        the
        process
        for predicting a structured output (or inference) is
        to find an output $\y$ that maximizes the joint compatibility function: \begin{equation}
            \label{eq:predict}
            \y^\star = \argmax_\y F ( \x, \y; \w ) =
            \argmax_\y
            \w^\T \wstructs( \x, \y).
        \end{equation}
        We denote by $ \ones $ a column vector of all $ 1$'s, whose dimension shall be clear
        from the context.
        We describe the \structboost approach in
        Section~\ref{sec:main}, including how
        to efficiently solve the resulting optimization problem. We then highlight
        applications in various domains in Section~\ref{sec:app}. Experimental results are shown in
        Section~\ref{sec:exp} and we conclude the paper in the last section.

\section{Structured boosting}
\label{sec:main}

    We first introduce the general structured boosting framework, and
	then
	apply it to a range of specific problems:
	classification, ordinal regression, optimizing special criteria
	such as the  area under the ROC curve and the Pascal
    image area overlap ratio, and learning CRF parameters.

To measure the accuracy of prediction we use a loss function, and as is the case with \svmstruct, we accept
 arbitrary loss functions $\loss: \calY \times \calY \mapsto \Real$.
    $ \loss( \y, \z ) $ calculates the loss associated with a prediction $ \z $ against the
    true label value $ \y $. Note that in general we assume that $ \loss( \y, \y ) = 0 $,
    $ \loss(\y, \z ) > 0 $ for any $ \z \neq \y $ and the loss is upper bounded.

The formulation of \structboost can be written as (\nslack primal):
\begin{subequations}
  \label{eq:structboost1}
\begin{align}
  \min_{ {\w \geq 0,\boldsymbol \xi \geq 0}}   \;\; &
        \wnorm +  {\tfrac{C}{m}} \, \one ^\T \bxi \\
  \st  \;\; &
    \w^\T \biggl[ \wstructs( \x_i, \y_i ) -  \wstructs (\x_i, \y )  \biggr] \geq
    \loss ( \y_i, \y) - \xi_i, \notag \\
   & \quad\quad\quad
   \forall i=1, \dots, m;
   \text{ and }\forall \y \in  {\cal Y}. \label{EQ:1b}
\end{align}
\end{subequations}
    Here we have used the $ \ell_1 $ norm as the regularization function
    to control the complexity of the learned model.
    To simplify the notation, we introduce
    \begin{align}
    \dwstruct_i ( \y ) = \wstruct ( \x_i, \y_i ) - \wstruct ( \x_i, \y );
    \end{align}
and,
    \begin{align}
    \dwstructs_i ( \y ) = \wstructs( \x_i, \y_i ) - \wstructs( \x_i, \y );
\end{align}
    then the constraints in \eqref{eq:structboost1} can be re-written as:
    \[   \w ^ \T  \dwstructs_i ( \y )  \geq \loss(\y_i, \y ) - \xi_i.  \]
    There are two major obstacles to solve problem \eqref{eq:structboost1}.
    First, as in conventional boosting, because the
    set
    of
    weak structured learners $ \wstruct(\cdot, \cdot) $ can be exponentially
	large or even infinite, the dimension
    of $ \w $ can be exponentially large or infinite.  Thus, in general,
    we are not able to directly solve for $ \w $.
    Second,
    as in \svmstruct, the number of constraints \eqref{EQ:1b}
    can be extremely or infinitely large. For example, in the case of
    multi-label or multi-class classification, the label $ \y $ can be
    represented as a binary vector (or string) and clearly
    the possible number of
    $ \y $ such that $ \y $ is exponential in the length of the vector,
    which is $ 2 ^ { | \calY | }$.
    In other words, {\em problem \eqref{eq:structboost1} can have an
    extremely or infinitely
    large number of variables and constraints.
    }
    This is significantly  more challenging than solving standard boosting or \svmstruct
    in terms of optimization. In standard boosting, one has a large number of
    variables while in \svmstruct, one has a large number of constraints.

    For the moment, let us put aside the difficulty of the large
    number of constraints, and focus on how to iteratively solve for $\w $
    using column generation as in boosting methods \cite{Demiriz2002LPBoost,Shen2011Totally,Shen2013NN}.
    We derive the Lagrange dual of the optimization of \eqref{eq:structboost1} as:
    \begin{subequations}
      \label{EQ:struct-dual1}
        \begin{align}
        \max_{\bmu \geq 0}   \;\; &
        \sum_{i, \y} \mu_{ (  i, \y) } \loss( \y_i, \y ) \\
        \st \;\; &
	\textstyle \sum_{i, \y  } \mu_{ (  i, \y) }\dwstructs_i( \y )  \leq \one,  \label{eq:dual_con} \\
	&  0 \leq \textstyle \sum_{ \y } \mu_{ (  i, \y) } \leq  \tfrac{C}{m}, \forall i=1,\dots, m.
    \end{align}
    \end{subequations}
   Here $ \bmu $ are the Lagrange dual variables (Lagrange multipliers). We denote by $ \mu_{
 ( i,\y ) }$ the dual variable associated with the margin
 constraints \eqref{EQ:1b} for label  $ \y $
    and training pair $ (\x_i, \y_i ) $.

    The idea of column generation is  to split the original primal problem in \eqref{eq:structboost1}
    into two problems: a master
    problem and a subproblem. The master problem is the original problem
    with only a subset of variables (or constraints for the dual form) being considered.
	The subproblem is to add new variables (or constraints for the dual form) into the master problem.
    With the primal-dual pair of \eqref{eq:structboost1}
    and \eqref{EQ:struct-dual1} and following the general framework of
    column generation based boosting
    \cite{Demiriz2002LPBoost,Shen2011Totally,Shen2013NN}, we can obtain our
    \structboost as follows:

    {\noindent \underline{Iterate the following two steps until converge} }:
    \begin{enumerate}
        \item
    Solve the following subproblem,
    which generates the best weak structured
	learner by finding the most violated constraint in the dual:
    \begin{align}
    \label{eq:wl_org}
    \wstruct^\star ( \cdot, \cdot ) =
       \argmax_{ \wstruct( \cdot, \cdot ) }
         \sum_{i, \y  }  \mu_{ ( i, \y ) }
         \dwstruct_i(\y).
    \end{align}
    \item
    Add the selected structured weak learner $\wstruct^\star ( \cdot, \cdot )$
    into the master problem (either the primal form or the dual form)
	and re-solve for the primal solution $ \w $ and dual solution $ \bmu $.
    \end{enumerate}
    The stopping criterion can be that no violated weak learner can be found. Formally, for the
    selected $  \wstruct^\star(\cdot, \cdot) $  with \eqref{eq:wl_org} and a preset precision
    $ \epsilon_{\rm
    cg} > 0 $, if the following relation holds:
    \begin{equation}
        \label{EQ:Stopping1}
    \sum_{i, \y } \mu_{ (i, \y) }
          \dwstruct^\star_i(\y)
     \leq 1 - \epsilon_{\rm cg},
    \end{equation}
    we terminate the iteration.
    Algorithm \ref{ALG:alg1} presents the details of column generation for \structboost.
	This approach, however, may not be practical because it is
    very expensive to solve the master problem
    (the reduced problem of \eqref{eq:structboost1}) at
    each column generation (boosting iteration), which still
    can have extremely many constraints due to the set of $ \{ \y \in
    \calY \} $.
    The direct formulation for multi-class boosting in \cite{Shen2011Direct} can be seen as a
    specific instance of this approach, which is in general very slow.
    We therefore propose to employ the \oneslack formulation for efficient
    optimization,  which is described in the next section.

    \subsection{$ 1 $-slack formulation for fast optimization}

    Inspired by the cutting-plane method for fast training of linear SVM
    \cite{JoachimsSVM} and SSVM \cite{Joachims2009},
    we  rewrite the above problem into
    an equivalent ``$1$-slack'' form so that the efficient cutting-plane method can
    be employed to solve the optimization problem  \eqref{eq:structboost1}:
\begin{subequations}
\label{eq:structboost-oneslack}
\begin{align}
  \min_{\w \geq 0, \xi \geq 0}  \;\;
  &
  \wnorm + C \xi \\
  \st  \;\; &
   \frac{1}{m} \w^ \T
  \biggl[
             \sum_{i=1}^m c_i
            \cdot
                    \dwstructs_i ( \y )
    \biggr]
            \geq
            \frac{1}{ m }  \sum_{i=1}^m c_i\loss(\y_i,\y ) - \xi, \notag \\
    &
    \quad\quad
            \forall \c \in\{0,1\}^m;
  \forall \y \in \calY, i=1,\cdots, m. \label{eq:1s_con}
\end{align}
\end{subequations}

The following theorem shows the equivalence of problems
\eqref{eq:structboost1} and \eqref{eq:structboost-oneslack}.
\begin{theorem}
    \label{THM:1}
        A solution of problem \eqref{eq:structboost-oneslack} is also a
        solution of   problem \eqref{eq:structboost1} and {\em vice versa}.
        The connections are:
        $  \w_\eqref{eq:structboost-oneslack}^\star
           = \w_\eqref{eq:structboost1}^\star
        $
        and
        $  \xi_\eqref{eq:structboost-oneslack}^\star
        = \tfrac{1}{m} \ones^\T \bxi_\eqref{eq:structboost1}^\star
        $.
\end{theorem}

\begin{proof}
            This proof adapts the proof in \cite{JoachimsSVM}. Given a
            fixed $ \w $, the only variable $ \bxi_\eqref{eq:structboost1} $ in
            \eqref{eq:structboost1} can be solved by
            \[
            \xi_{i,\eqref{eq:structboost1} }  = \max_{ \y
            }
            \Bigl\{ 0,  \loss( \y_i, \y ) - \w ^\T \dwstructs_i ( \y)
            \Bigr\}, \forall i.
            \]
            For \eqref{eq:structboost-oneslack}, the
            optimal $ \xi_\eqref{eq:structboost-oneslack} $ given a $
            \w $ can be computed as:
            \begin{align*}
                \xi_\eqref{eq:structboost-oneslack}
                &=
                \frac{1}{m}
                \max_{ \c , \y  }
                \left\{
                          \sum_{i=1}^m c_i \loss(\y_i, \y) -
                            \w ^\T
                                \Bigl[  \sum_{i=1}^m c_i \dwstructs_i (
                          \y )   \Bigr]
                \right\}
                \\
                &
                = \frac{1}{m}
                \sum_{i=1}^m
                \left\{
                \max_{ c_i \in \{ 0, 1\}, \y  }
                        c_i \loss(\y_i, \y)  - c_i \w^\T
                            \dwstructs_i ( \y )
                \right\}
                \\
                &
                = \frac{1}{m}  \sum_{i=1}^m  \max_{ \y  }
                                                  \Bigl\{
                                                             0,  \loss( \y_i, \y ) - \w ^\T \dwstructs_i ( \y)
                                                  \Bigr\}
                \\
                &
                = \frac{1}{m} \ones ^\T \bxi_\eqref{eq:structboost1}.
            \end{align*}
            Note that $ \c \in \{ 0, 1 \}^m $ in the above equalities.
            Clearly the objective functions of both problems coincide
            for any fixed $ \w $ and the optimal $  \bxi_\eqref{eq:structboost1} $
            and $ \xi_\eqref{eq:structboost-oneslack}$.
\end{proof}
    As demonstrated in \cite{JoachimsSVM} and SSVM \cite{Joachims2009}, cutting-plane methods can
    be used to solve the $ 1 $-slack primal problem
    \eqref{eq:structboost-oneslack} efficiently.  This
    $ 1 $-slack formulation has been used to train linear SVM in
    linear time. When solving for $ \w $, \eqref{eq:structboost-oneslack}
    is similar to $ \ell_1 $-norm regularized {\svm}---except the
    extra non-negativeness constraint on $ \w $ in our case.

    In order to utilize column generation for designing boosting methods,
    we need to derive the Lagrange dual of the $1$-slack primal optimization problem,
    which can be written as follows:
    \begin{subequations}
      \label{eq:structboost-oneslack-dual}
  \begin{align}
    \max_{ \blambda \geq 0 }
    \;\;
    &
    \sum_{ \c , \y }
    \lambda_{ (\c, \y ) }
    \sum_{i=1}^m c_i  \loss( \y_i, \y ) \\
   \st \;\;
   &  \frac{1}{m} \sum_{ \c, \y } \lambda_{ (\c, \y) }
     \biggl[
          \sum_{i=1}^m c_i \cdot  \dwstructs_i( \y )
     \biggr]
         \leq \ones, \\
      &  0 \leq \textstyle \sum_{ \c, \y} \lambda_{ (\c, \y ) } \leq C.
\end{align}
\end{subequations}
    Here $ \c $ enumerates all possible $ \c \in \{ 0, 1 \}^m$.
    We denote by $ \lambda_{( \c, \y)} $ the Lagrange dual variable (Lagrange
    multiplier) associated with the inequality constraint in \eqref{eq:1s_con}  for  $ \c
    \in \{ 0, 1 \}^m $ and label $ \y $.
    The subproblem to find the most violated constraint in the dual form for
    generating weak structured learners is:
    \begin{align}
\label{EQ:WL2}
        \wstruct^\star ( \cdot, \cdot ) & =
       \argmax_{ \wstruct( \cdot, \cdot ) }
        \sum_{ \c, \y }  \lambda_{ ( \c, \y ) }
        \sum_i c_i
        \delta \wstruct_i(\y)
        \notag
        \\
        &= \argmax_{ \wstruct( \cdot, \cdot ) } \sum_{i, \y }
        \underbrace{
        \sum_{\c}
        \lambda_{ ( \c, \y ) } c_i
        }_{ := \mu_{ (i, \y) } }
        \delta \wstruct_i(\y).
    \end{align}
    We have changed the order of summation in order to have a similar form as in the
    $ m$-slack case.

    \input{alg_cg.tex}

\input{alg_cutting_plane.tex}

    \subsection{Cutting-plane optimization for the $1$-slack primal}

    Despite the extra nonnegative-ness constraint  $ \w \geq 0 $ in
    our case, it is straightforward to apply the cutting-plane method in
    \cite{JoachimsSVM}  for solving our problem
    \eqref{eq:structboost-oneslack}.
    The cutting-plane algorithm for \structboost
 is presented in Algorithm \ref{ALG:alg2}.
A key step in Algorithm \ref{ALG:alg2} is to solve the maximization for
finding an output $\y'$ that corresponds
to the most violated constraint for every $\x_i$ (inference step):
    \begin{align}
    \label{eq:infer}
            \y_i'= \argmax_{\y} \loss( \y_i, \y ) + \w ^ \T  \dwstructs_i ( \y ).
    \end{align}
	The above maximization problem takes a similar form as the output prediction
    problem in \eqref{eq:predict}.
	They only differ in the loss term $\loss( \y_i, \y )$.
	Typically these two problems  can be solved using the same strategy.
	This inference step usually dominates the running time for a few applications,
    e.g., in the application of image segmentation.
    In the experiment section,
    we empirically show that solving \eqref{eq:structboost-oneslack}
    using cutting planes can be significantly faster than solving
    \eqref{eq:structboost1}.
    Here improved cutting-plane methods such as \cite{Franc2008} can also be adapted to
    solve our optimization problem at each column generation boosting iteration.

	In terms of  implementation of  the cutting-plane algorithm, as mentioned in SSVM \cite{Joachims2009},
	a variety of design decisions can have substantial influence on the practical efficiency of the algorithm.
	We have considered some of  these design decisions in our implementation.
	In our case, we need to call the cutting-plane optimization at  each column generation
    iteration.
    Consideration of  warm-start initialization between two consecutive column generations can substantially
    speed up
    the training.
    We re-use the working set in the cutting-plane algorithm from  previous column generation iterations.
	Finding a new weak learner in \eqref{EQ:WL2} is based on the dual solution $\bmu$.
	We need to ensure that the solution of cutting-plane is able to reach a sufficient precision,
	such that the generated weak learner is able to ``make progress".
	Thus, we can adapt the stopping criterion parameter $\epsilon_{\rm cp}$
    (Line 10 of Algorithm 2)
	according to the cutting-plane precision
	in the last column generation iteration.

\subsection {Discussion}
Let us have a close look at the \structboost algorithm in Algorithm \ref{ALG:alg1}.
We can see that the training loop in Algorithm \ref{ALG:alg1} is almost identical to other
fully-corrective boosting methods (e.g., LPBoost \cite{Demiriz2002LPBoost} and Shen and Li
\cite{Shen2011Totally}). Line 4 finds the most violated constraint in the dual form and add a
new weak structured learner to the master problem.
The dual solution $ \mu_{ ( i, \y ) } $ defined in \eqref{EQ:WL2} plays the role as the example
weight associated to each training example in conventional boosting methods
such as AdaBoost and
LPBoost \cite{Demiriz2002LPBoost}. Then Line 5 solves the master problem, which is the reduced problem of
\eqref{eq:structboost1}.
Here we can see
that, the cutting-plane in Algorithm \ref{ALG:alg2} only serves as a solver for solving the master
problem in Line 5 of Algorithm \ref{ALG:alg1}.
This makes our \structboost framework flexible---we are able to replace the cutting-plane optimization
by other optimization methods. For example, the bundle methods in \cite{Teo2010} may
further speed up the computation.

For the convergence properties  of the cutting-plane algorithm in Algorithm \ref{ALG:alg2},
readers may refer to \cite{JoachimsSVM} and \cite{Joachims2009} for details.

	Our column generation algorithm in Algorithm \ref{ALG:alg1} is
	 a constraint generation algorithm for the dual problem in \eqref{EQ:struct-dual1}.
	We can adapt the analysis of the standard constraint generation algorithm
	for Algorithm \ref{ALG:alg1}.
    In general, for general  column generation methods, the
    global convergence can be established but it remains unclear
    about the convergence rate if no particular assumptions are made.
    See the supplementary document\footnote{Available at \url{http://arxiv.org/abs/1302.3283} }
    for details.

\input{app}

\section{Experiments}
\label{sec:exp}

To evaluate our method,
we run various experiments on applications including AUC maximization (ordinal regression),
 multi-class image classification, hierarchical image classification, visual tracking and image segmentation.
We mainly compare with the most relevant method:
Structured SVM (SSVM) and some other conventional methods
(e.g., SVM, AdaBoost). If not otherwise specified,
the cutting-plane stopping criteria ($\epsilon_{\rm cp}$) in our method is set to $0.01$.

\input{mc_exp}

\input{hmc_exp}

\input{tracking_exp}

\input{seg_exp}

\section{Conclusion}

	we have presented a structured boosting method,
    which combines a set of weak structured learners for nonlinear structured output leaning,
    as an alternative to \svmstruct \cite{Tsochantaridis2004} and CRF \cite{Lafferty01Conditional}.
    Analogous to \svmstruct, where the discriminant function is
    learned over a joint feature space of inputs and outputs,
    the discriminant function of the proposed  \structboost
    is a linear combination of weak structured learners defined over a joint space of input-output pairs.

     To efficiently solve the resulting
     optimization problems, we have introduced  a cutting-plane method,
     which was originally proposed for fast training of linear \svm.
     Our extensive experiments demonstrate that indeed
     the proposed algorithm is computationally tractable.

        \structboost is flexible in the sense that it can be used to
       optimize a wide variety of loss functions. We have exemplified
       the application of \structboost by applying to multi-class
       classification, hierarchical multi-class
       classification by optimizing the tree loss, visual tracking that
        optimizes the Pascal  overlap criterion,
        and learning \crf parameters for image segmentation.
        We show that \structboost at least is
    comparable or sometimes exceeds conventional approaches for a wide
    range of applications.
       We also observe that \structboost has improved
       performance over linear \ssvm, demonstrating the usefulness of
       our nonlinear structured learning method.
    Future work will focus on more applications of this general
    \structboost framework.

{
\footnotesize
\bibliographystyle{IEEEtran}
\bibliography{CSRef-structboost}
}

\input supplementary

\end{document}

%% file: macros.tex
\newcommand{\area}{{\rm area}}

\newcommand{\st}{{{\rm s.t.}\!:}\xspace}

\def\wnorm{\ones^\T \w}

\def\argmax{\operatorname*{argmax\,}}

\newcount\timehh\timehh=\time
\divide\timehh by 60
\newcount\timemm\timemm=\time
\multiply\count255 by -60
\advance\timemm by \count255
\newif\iftimePM
\ifnum\timehh>11 \timePMtrue\else\timePMfalse\fi
\ifnum\timehh<1 \advance\timehh by 12\fi
\ifnum\timehh>12 \advance\timehh by -12\fi
\def\now{\number\timehh:\ifnum\timemm<10 0\fi\number\timemm
         \iftimePM {$\,$pm.}  \else {$\,$am.}  \fi}
\newcount\mdYY\mdYY=\year
\divide\count255 by 100
\multiply\count255 by 100
\advance\mdYY by -\count255
\def\mdyy{\number\day/\number\month/\ifnum\mdYY<10 0\fi\number\mdYY}

\newcommand{\comment}[1]{}

\newcommand{\CSComment}[1]{{\it {\bf FIXME} (CS, \now\mdyy): \color{blue}#1}}

\renewcommand{\CSComment}[1]{}

\def\T{{\!\top}}

\def\Real{\mathbb{R}}

\def\c{{\boldsymbol c}}
\def\x{{\boldsymbol x}}
\def\w{{\boldsymbol w}}
\def\y{{\boldsymbol y}}
\def\z{{\boldsymbol y}'}
\def\v{{\boldsymbol v}}

 \def\wls{{\Phi}}
 \def\wl{\phi}

\def\wstructs{{\Psi }}
\def\wstruct{{ \psi }}

\def\blambda{{\boldsymbol \lambda}}

\def\bmu{{\boldsymbol \mu}}

\def\bxi{{\boldsymbol \xi}}

\def\calY{{\cal Y}}
\def\calX{{\cal X}}

\def\loss{\bigtriangleup}
\def\loss{{\Delta}}
\def\dwstructs{{\delta}{\wstructs}}
\def\dwstruct{{\delta}{\wstruct}}

\newcommand{\ones}{{\mathbf{1}}}
\newcommand{\one}{{\mathbf{1}}}

\newtheorem{theorem}{Theorem}[section]

\newtheorem{proposition}{Proposition}[section]

\usepackage{upgreek}
\let\delta\updelta

%% file: alg_cg.tex
\begin{algorithm}[t]
\caption{Column generation for \structboost}
\footnotesize{
1: {\bf Input:} training examples $ (\x_1; \y_1), (\x_2; \y_2) ,\cdots
    $; parameter $C$; termination threshold $\epsilon_{\rm cg}$,  and
    the maximum iteration number.

    2: {\bf Initialize:}   for each $i$, ($i=1,\dots,m $),
    randomly pick any  $\y_i^{(0)} \in \calY$,
    initialize $ \mu_{(i,\y)} = \tfrac{C}{m} $ for $ \y = \y_i^{(0)}$, and
    $ \mu_{(i,\y)} = 0 $ for all $ \y \in \calY
    \backslash \y_i^{(0)}$.

    3: {\bf Repeat}

    4:$\quad-$ Find and add a weak structured learner $\wstruct^\star ( \cdot, \cdot )$ by solving
          the subproblem \eqref{eq:wl_org} or \eqref{EQ:WL2}.

    5:$\quad-$
    Call   Algorithm \ref{ALG:alg2}
    to obtain $\w$ and $\bmu$.

    7: {\bf Until}
either
    \eqref{EQ:Stopping1} is met
    or the maximum number of iterations is reached.

    8: {\bf Output:}
    the discriminant function  $ F( \x, \y; \w  )  =  \w^\T  \wstructs(\x, \y) $.
    }
\label{ALG:alg1}
\end{algorithm}

%% file: alg_cutting_plane.tex
\begin{algorithm}[t]
\caption{ Cutting planes for solving the \oneslack primal}
\footnotesize{

1: {\bf Input:} the cutting-plane termination threshold $\epsilon_{\rm cp}
$,
and inputs from Algorithm \ref{ALG:alg1}.

2: {\bf Initialize:} working set $\mathcal{W}\leftarrow \emptyset
$;  $c_i = 1$, $\y_i' \leftarrow $ any element in $\calY $, for $
i=1,\dots, m$.

    3: {\bf Repeat}

    4: $\quad-$ $\mathcal{W}\leftarrow \mathcal{W} \cup \{ (c_1,
    \dots, c_m, \y_1',\dots,\y_m')  \} $.

    5: $\quad-$ Obtain primal and dual solutions $\w,\xi$; $\blambda $
    by solving
\begin{align}
  \min_{\w\geq 0, \xi\geq 0} & \;\;   \wnorm + C \xi \notag\\
  \st & \;\; \forall (c_1,\dots,c_m, \y_1',\dots,\y_m') \in \mathcal{W}: \notag \\
   & \frac{1}{m} \w^ \T
  \biggl[
             \sum_{i=1}^m c_i
            \cdot
                    \dwstructs_i ( \y_i' )
    \biggr]
            \geq
            \frac{1}{ m }  \sum_{i=1}^m c_i\loss(\y_i,\y_i' ) - \xi.  \notag
            \end{align}

    6: $\quad-$ {\bf For} $i=1, \dots, m$

    7: $\quad\quad\quad$ $ \y_i'= \argmax_{\y} \loss( \y_i, \y ) - \w ^ \T  \dwstructs_i ( \y ) $;

    8: $\quad\quad\quad$ $ c_i = \left\{
      \begin{array}{l l}
        1 & \quad \loss( \y_i, \y_i' ) - \w ^ \T  \dwstructs_i ( \y_i' ) > 0; \\
        0 & \quad {\rm otherwise.} \\
      \end{array}
    \right. $

    9: $\quad\;\;\,$ {\bf End for}

    10: {\bf Until}
$
      \frac{1}{m} \w^ \T
  \biggl[
             \sum\limits_{i=1}^m c_i
                    \dwstructs_i ( \y_i' )
    \biggr]
            \geq
            \frac{1}{ m } \sum\limits_{i=1}^m c_i\loss(\y_i,\y_i' ) - \xi -
            \epsilon_{\rm cp}.
$

    11:$\quad-$
    Update $ \mu_{(i,\y)} = \sum_{\c}\lambda_{ ( \c, \y ) } c_i$ for $\forall i=1, \dots, m$; $\forall \y \in  {\cal Y} \backslash \y_i$.

    12: {\bf Output:} $\w$ and $\bmu$.

}
\label{ALG:alg2}
\end{algorithm}

%% file: app.tex
\section{Examples of \structboost}
        \label{sec:app}

    We consider a few applications of the proposed general structured boosting
    in this section, namely binary classification, ordinal regression, multi-class classification,
    optimization of Pascal overlap score, and CRF parameter learning.
    We show the particular setup for each application.

    \paragraph{Binary classification}
    As the simplest example,
    the LPBoost of \cite{Demiriz2002LPBoost} for binary classification
    can be recoverd as follows.
    The label set is $ { \cal Y } = \{ +1, -1 \}  $; and
    $
        \wstructs(\x, y)  = \frac{1}{2}   y \Phi ( \x ).
    $
    The label cost can be a simple constant; for exmaple,
    $ \loss(y,y') = 1$ for $ y \neq y' $ and  $0$ for $ y = y'$.
    Here we have introduced a column vector $\Phi(\x):$
    \begin{equation}
        \Phi ( \x ) = [  \phi_1(\x), ..., \phi_n (\x)  ]^\T,
        \label{EQ:Phi}
    \end{equation}
    which is the outputs of all weak classifiers $ \phi_1(\cdot)$, $ \cdots$, $ \phi_n(\x)$
    on example $ \x$.
    The output of a weak classifier, e.g., a decision stump or tree,
    usually is a binary value: $ \phi(\cdot) \in \{+1, -1 \}$.
    In kernel methods, this feature mapping $ \Phi(\cdot) $ is only known through the so-called kernel trick.
    Here we explicitly learn this feature mapping.
    Note that if $ \Phi(\x) = \x $,
    we have the standard linear SVM.

    \paragraph{Ordinal regression and AUC optimization}
    \label{sec:auc}
    In ordinal regression, labels of the training data are ranks. Let
    us assume that the label $ y \in \Real $ indicates an ordinal
    scale, and pairs $( i, j ) $ in the set $ \cal S$
    has the relationship of example $ i $ being ranked higher than $ j
    $, i.e., $ y_i \succ  y_j $.
    The primal can be written as
        \begin{subequations}
                            \label{eq:rank1}
        \begin{align}
  \min_{ {\w \geq 0,\boldsymbol \xi \geq 0}}   \;\; &
  \wnorm +  {\tfrac{C}{m}} {\textstyle \sum}_{ (i,j) \in \cal S} \xi_{ij} \\
  \st  \;\; &
  \w^\T \biggl[ \wls ( \x_i) -  \wls (\x_{j} )  \biggr] \geq
  1 - \xi_{ij},
  \forall (i,j) \in {\cal S}.
\end{align}
\end{subequations}
Here $ \Phi(\cdot) $ is defined in \eqref{EQ:Phi}.
    Note that \eqref{eq:rank1}
    also optimizes the area under the receiver operating
    characteristic (ROC) curve (AUC) criterion.
    As pointed out in \cite{Joachims2005ICML},
    \eqref{eq:rank1} is an instance of
    the multiclass  classification problem.
    We discuss how the multiclass classification problem fits in our framework
    shortly.

    Here,
    the number of constraints is quadratic in the number of training
    examples. Directly solving \eqref{eq:rank1}
    can only solve problems with up to a few thousand training examples.
    We can reformulate \eqref{eq:rank1} into an equivalent \oneslack problem,
    and apply the proposed  \structboost framework to solve the optimization more efficiently.

    \paragraph{Multi-class boosting}
    \label{sec:mc}

    The MultiBoost algorithm of  Shen and Hao \cite{Shen2011Direct}
    can be implemented by the \structboost framework as follows.
    Let $ \calY = \{ 1,2,\dots,k\}$ and $ \w = \w_1 \odot \cdots \odot
    \w_k $. Here $ \odot $ stacks two vectors. As in
    \cite{Shen2011Direct}, $ \w_y $ is the model parameter associated
    with the $ y $-th class.
The multi-class discriminant function
    in \cite{Shen2011Direct} writes
    $ F( \x, y; \w ) = \w_y^\T  { \wls }( \x )   $.
    Now let us define the orthogonal label coding vector:
    \begin{equation}
    \Gamma (y) = [ \Ind(y,1), \cdots, \Ind(y, k) ]^\T
    \in \{0, 1\}^k.
    \label{EQ:gamma}
    \end{equation}
    Here $\Ind (  y, z  )$ is the indicator function defined as:
    \begin{equation}
        \label{EQ:INDI}
        \Ind( y, z  ) = \begin{cases} 1  & \mbox{ if } y = z, \\
                                        0  & \mbox{ if } y \neq z.
                          \end{cases}
    \end{equation}
    Then the following joint mapping function
    \[
        \wstructs( \x, y )  = \wls( \x ) \otimes \Gamma (y)
    \]
    recovers the \structboost formulation \eqref{eq:structboost1}
    for multi-class boosting. The operator $ \otimes $ calculates the tensor product.
The multi-class learning can be formulated as
\begin{subequations}
 \label{eq:mc}
        \begin{align}
  \min_{ {\w \geq 0,\boldsymbol \xi} \geq 0}   \;\; &
         \wnorm   +  {\tfrac{C}{m}} \, \one ^\T \bxi \\
  \st  \;\; &
  \w_{y_i}^\T \wls( \x_{i}) - \w_{y}^\T \wls (\x_{i} ) \geq
        1 - \xi_i, \notag \\
   &
   \forall i=1, \dots, m;
   \text{ and }\forall y \in  \{1,\dots,k\}.
\end{align}
\end{subequations}
    A new weak classifier $\wl(\cdot)$ is generated by solving the $\argmax$
problem defined in \eqref{EQ:WL2}, which can be written as:
    \begin{align}
    \label{eq:wl_mc}
    \wl^\star ( \cdot ) =&
       \argmax_{ \wl(  \cdot ), y }
         \sum_{i, y }  \mu_{ ( i, y ) }
        \Bigl[ \wl( \x_i ) \otimes \Gamma (y_i)
    - \wl( \x_i ) \otimes \Gamma (y) ) \Bigr].
    \end{align}

\begin{figure}[t]
    \centering
        \begin{subfigure}{\linewidth}
		\centering
		\includegraphics[width=.81\linewidth]{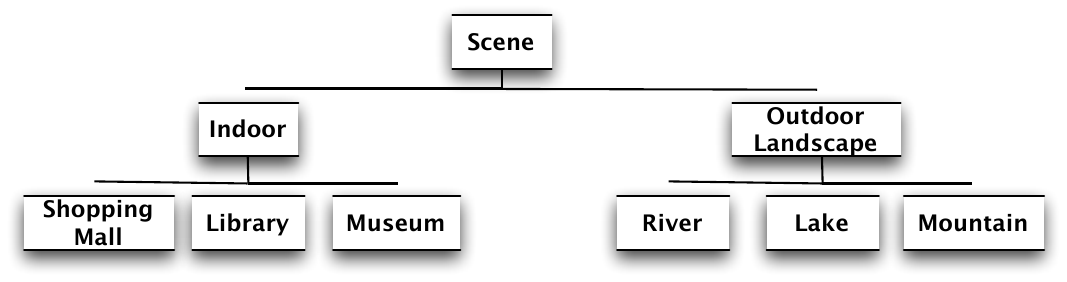}
		\caption{Taxonomy of the 6-scene dataset}
	\end{subfigure}

        \begin{subfigure}{\linewidth}
		\centering
		\includegraphics[width=.81\linewidth]{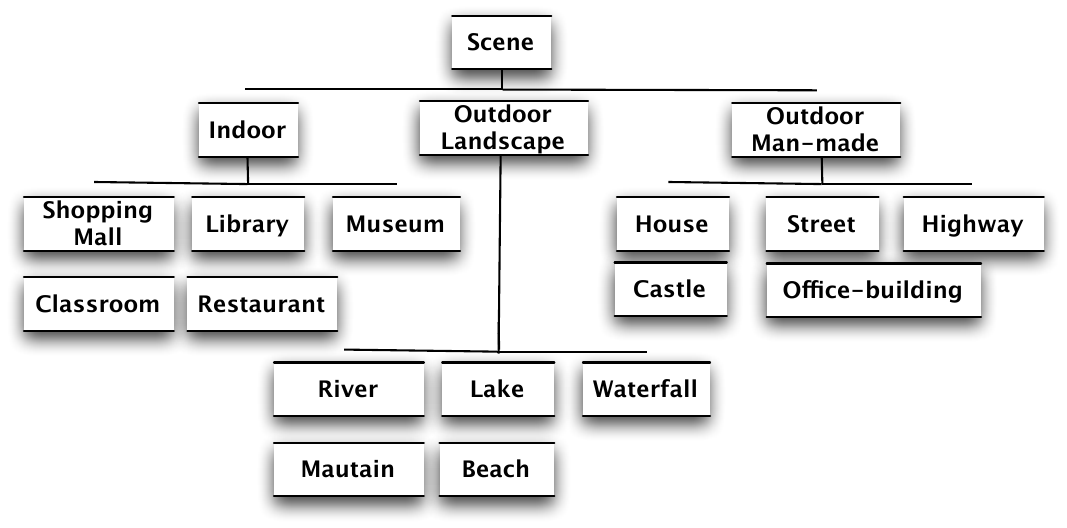}
		\caption{Taxonomy of the 15-scene dataset}
	\end{subfigure}
    \caption{The hierarchy  structures of two selected subsets of
    the SUN dataset \cite{SUN} used in our experiments for
    hierarchical image classification.}
    \label{fig:h_structure}
\end{figure}

\paragraph{Hierarchical classification with taxonomies}
\label{sec:hmc}
    In many applications such as object categorization and document classification \cite{Cai2004},
    classes of
    objects are organized in taxonomies or hierarchies. For example,
    the ImageNet dataset has organized all the classes according to
    the tree structures of WordNet \cite{wordnet}.
    This problem is a classification example that the output space has
    interdependent structures. An example tree structure (taxonomy) of image
    categories is shown in Figure \ref{fig:h_structure}.

    Now we consider the taxonomy to be a tree, with a partial order $ \prec $,
    which indicates if a class is a predecessor of another class.
    We override the indicator function, which indicates if $z$ is a
    predecessor of $y$ in a label tree:
    \begin{equation}
        \Ind(y,z)
            = \begin{cases} 1  & y \prec  z \text{ or } y = z, \\
                                        0  & \mbox{ otherwise. }
                          \end{cases}
        \label{EQ:IND2}
    \end{equation}
    The label coding vector has the same format as in the
    standard multi-class classification case:
    \begin{equation}
        \Gamma ( y ) = [ \Ind(y,1), \cdots, \Ind(y, k)
        ]^\T \in \{0,1\}^k.
    \end{equation}
    Thus
    $\Gamma(y) ^\T \Gamma(y') $ counts the number of common
    predecessors, while in the case of standard multi-class
    classification, $\Gamma(y) ^\T \Gamma(y') = 0 $ for
    $ y \neq y'$.

\begin{figure}[t]
    \centering
	\includegraphics[width=.55\linewidth]{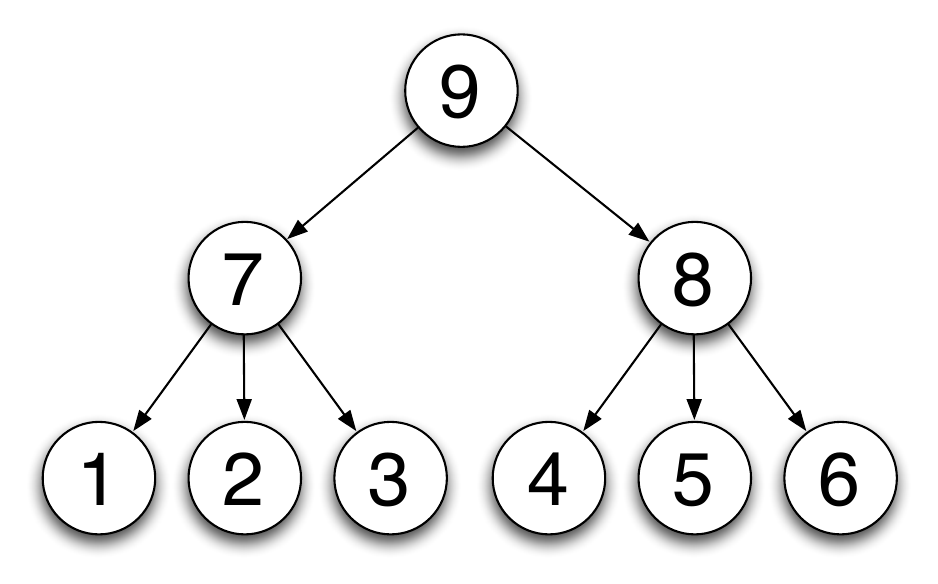}
		\caption{Classification with taxonomies (tree loss), corresponding to the first
        example in Figure \ref{fig:h_structure}.
    }
    \label{FIG:Tax1}
\end{figure}

    Figure \ref{FIG:Tax1} shows an example of the
    label coding vector for a given label hierarchy.
    In this case, for example, for class 3,
    $ \Gamma(3) = [ 0, 0, 1, 0, 0, 0, 1, 0, 1 ]^\T   $.
    The joint mapping function is
     $
        \wstructs( \x, y )  = \wls( \x ) \otimes \Gamma (y).
     $
    The tree loss function $ \loss(y,y')  $ is the height
    of the first common ancestor of the arguments $ y, y' $ in the tree.
    By re-defining $ \Gamma(y)   $ and $ \loss(y,y')$, classification
    with taxonomies can be immediately implemented using the standard multi-class
    classification shown in the last subsection.

    Here we also consider an alternative approach.
    In \cite{Paisitkriangkrai2013RandomBoost}, the authors show that
    one can train a multi-class boosting classifier by projecting data to
    a label-specific  space and then learn a single model parameter $ \w $.
    The main advantage might be that the optimization of $ \w $ is simplified.
    Similar to \cite{Paisitkriangkrai2013RandomBoost} we define  label-augmented data as
    $ \x_y' = \x \otimes \Gamma(y) $.
    The max-margin classification can be written as
        \begin{align*}
  \min_{ {\w \geq 0,\boldsymbol \xi \geq 0}}   \;\; &
   \wnorm   +  {\tfrac{C}{m}} \ones ^\T { \boldsymbol  \xi} \\
  \st  \;\; &
  \w^\T \biggl[ \wls ( \x'_{i, y_i }) -  \wls (\x'_{ i , y } )  \biggr] \geq
  \loss( y_i, y  ) - \xi_{i},
  \\
  & \quad\quad\quad    \forall i =1, \cdots, m; \text{ and } \forall y.
\end{align*}
    Compared with the first approach, now the model
    $ \w \in \Real^n $, which is independent
    of the number of classes.

    \paragraph{Optimization of the Pascal image overlap criterion}
    \label{sec:track}
     Object detection/localization has used
        the image area overlap as the loss function
        \cite{Blaschko2008,Torr2011ICCV,Nowozin2011Struct}, e.g, in
        the Pascal  object detection challenge:
        \begin{align}
   \loss( \y, \y' ) = 1 - \frac{ \area( \y \cap  \y')  }{ \area( \y
   \cup  \y')  },
        \end{align}
        with $ \y, \y' $ being the bounding box coordinates.
        $ \y \cap  \y' $ and  $ \y \cup  \y' $
        are the box intersection and union.
	Let $\x_\y$ denote an image patch defined
        by a bounding box $ \y $ on the image $\x$.
	    To apply \structboost, we define $\wstructs( \x, \y )=\wls( \x_\y )$.
	$\wls(\cdot)$ is defined in \eqref{EQ:Phi}.
	Weak learners such as classifiers or regressors $\wl(\cdot)$ are trained on the image features
	extracted from image patches. For example, we can extract histograms of oriented
        gradients (HOG) from the image patch $\x_\y $ and train a
        decision stump with the extracted HOG features by solving the $\argmax$ in \eqref{EQ:WL2}.

        Note that in this case, solving \eqref{eq:infer},
        which is to find the most violated constraint in the training step
        as well as the inference for prediction \eqref{eq:predict},
        is in general difficult. In
        \cite{Blaschko2008}, a branch-and-bound search has been
        employed to find the global optimum. Following the simple sampling strategy in \cite{Torr2011ICCV},
        we simplify this problem by evaluating a certain number of sampled image patches
        to find the most violated constraint. It is also the case for prediction.

\paragraph{\crf parameter learning}
\label{sec:seg}
    \crf has found successful applications in
many vision problems such as pixel labelling, stereo matching and image segmentation.
Previous work often uses tedious cross-validation  for setting the \crf
parameters. This approach is only feasible for a small number of parameters.
Recently, \ssvm
has been introduced to learn the parameters  \cite{SzummerKH08}.
We demonstrate here how to employ the proposed \structboost for \crf parameter learning
in the image segmentation task.
We demonstrate the effectiveness of our approach on
 the Graz-02 image segmentation dataset.

    To speed up computation, super-pixels rather than pixels
    have been widely adopted in image segmentation.
We define $\x$ as an image and $\y$ as the segmentation labels of all super-pixels in the image.
We consider the energy $E$ of  an image $\x$ and segmentation labels $\y$ over the nodes $\cal N$ and edges $\cal S$,
which takes the following form:
\begin{align}
\label{eq:seg_energy}
	E (\x, \y; \w) &
    = \sum_{p \in {\cal N} } \w^{(1)\T} \wls^{(1)} \left( \U(y^{p}, \x)
               \right) \notag \\
	& + \sum_{(p,q) \in {\cal S}}  \w^{(2)\T} \wls^{(2)} \left( \V( y^{p}, y^{q}, \x) \right).
\end{align}
Recall that $ \Ind(\cdot, \cdot)$ is the indicator function defined in \eqref{EQ:INDI}.
$p$ and $q$ are the super-pixel indices; $y^{p}, y^{q}$ are the labels of the super-pixel $p, q$.
$\U$  is a set of unary potential functions: $\U=[\, U_1, U_2, \dots ]^\T$.
$\V$ is a set of pairwise potential functions: $\V=[\, V_1, V_2, \dots ]^\T$.
When we learn the CRF parameters,  the learning algorithm sees only $ \U $ and $ \V $.
In other words $ \U $ and $ \V $ play the role as the input features.
Details on how to construct $\U$ and $\V$ are described in the experiment part.
$\w^{(1)}$ and $\w^{(2)}$ are the \crf potential weighting parameters that we aim to learn.
$\wls^{(1)}(\cdot)$ and $\wls^{(2)}(\cdot)$ are two sets of weak learners (e.g., decision stumps)
for the unary part and pairwise part respectively:
$\wls^{(1)}(\cdot)=[\, \wl_1^{(1)}(\cdot), \wl_2^{(1)}(\cdot),\dots]^\T$,
$\wls^{(2)}(\cdot)=[ \, \wl_1^{(2)}(\cdot), \wl_2^{(2)}(\cdot), \dots]^\T$.

To predict the segmentation labels $\y^\star$ of an unknown image $\x$ is
to solve the energy minimization problem:
\begin{align}
    \label{eq:seg_predict}
    \y^\star=\argmin_{\y} E(\x,\y; \w),
\end{align}
which can be solved efficiently by using
graph cuts \cite{SzummerKH08, fulkerson08localizing}.

Consider a segmentation problem with two classes (background versus foreground).
    It is desirable to keep the submodular property of the energy function in \eqref{eq:seg_energy}.
Otherwise graph cuts cannot be directly applied to achieve  globally optimal labelling.
Let us examine the pairwise energy term:
\[
    \theta_{(p,q)}(y^{p}, y^{q})= \w^{(2)\T} \wls^{(2)} \left( \V( y^{p}, y^{q}, \x) \right),
\]
and a super-pixel label $y \in \{0, 1\}$.
It is well known that, if the following is satisfied for any pair $ (p,q) \in {\cal S}$,
the energy function in \eqref{eq:seg_energy} is submodular:
\begin{align}
    \label{eq:submodular}
    \theta_{(p,q)}(0, 0) + \theta_{(p,q)}(1, 1) \leq \theta_{(p,q)}(0, 1) + \theta_{(p,q)}(1, 0).
\end{align}
We want to  keep the above property.

First, for a weak learner $ \wl^{(2)} (\cdot)  $,
we enforce it to output $0$ when two labels are identical.
This can always be achieved by multiplying $(  1 - \Ind( y^p, y^q ) )$ to a conventional  weak learner.
Now we have $\theta_{(p,q)}(0, 0) $ $=$ $  \theta_{(p,q)}(1, 1) = 0 $.

Given that the
nonnegative-ness of $\w$ is enforced in our model,
now a sufficient condition is that the output of a weak learner $\wl^{(2)}(\cdot) $ is always nonnegative,
which can always be achieved.
We can always use a weak learner $\wl^{(2)} (\cdot) $ which takes a nonnegative output,
e.g., a discrete decision stump or tree with outputs in  $\{0, 1\}$.

    By applying weak learners  on $ \U$ and $ \V $, our method introduces
    nonlinearity for the parameter
    learning, which is different from  most linear \crf learning methods such as \cite{SzummerKH08}.
    Until recently,
    Berteli et al.\ presented an image segmentation approach that uses
    nonlinear kernels for the unary energy term in the CRF model \cite{BertelliYVG11}.
    In our model, nonlinearity is introduced by applying weak learners on the potential functions'
    outputs. This is the same as the fact that an SVM introduces nonlinearity via the
    so-called kernel trick and boosting learns a nonlinear model with nonlinear weak learners.
    Nowozin et al. \ \cite{DTF} introduced decision tree fields (DTF) to overcome the
    problem of overly-simplified
    modeling of pairwise potentials in most CRF models. In DTF, local interactions between
    multiple variables are determined by means of decision trees.
    In our StructBoost, if we use decision trees as the weak learners on
    the pairwise potentials, then StructBoost and DTF share similar characteristics in that both
    use decision trees for the same purpose. However, the starting points of these two methods as well
    as the training procedures are entirely different.

  To apply \structboost, the \crf parameter learning problem in a large-margin framework can then be written as:
  \begin{subequations}
\label{eq:energy}
\begin{align}
  \min_{ {\w \geq 0, \boldsymbol \xi \geq 0}}   \;\; &
        \ones ^ \T \w
        +  {\tfrac{C}{m}} \ones^\T { \boldsymbol \xi } \\
  \st  \;\; &
    E(\x_i, \y ; \w)- E (\x_i, \y_i; \w )\geq
    \loss ( \y_i, \y) - \xi_i, \notag \\
   & \quad\quad\quad
   \forall i=1, \dots, m;
   \text{ and }\forall \y \in  {\cal Y}.
\end{align}
\end{subequations}
Here $ i $ indexes images.
Intuitively, the optimization in \eqref{eq:energy} is to encourage the
energy of the ground truth label $E(\x_i,\y_i)$ to be lower than any other
{\em incorrect} labels $E(\x_i,\y)$ by at least a margin $\loss (\y_i, \y)$, $ \forall \y $.
We simply define $\loss ( \y_i, \y)$ using the Hamming loss,
which is the sum of the differences between
the ground truth label $\y_i$ and the label $\y$ over all super-pixels in an image:
\begin{align}
\label{eq:hamming}
\loss ( \y_i, \y) = \sum_p  ( 1 - \Ind ( y_i^{p} ,   y^{p})  ).
\end{align}
We show the problem \eqref{eq:energy} is a special case of
the general formulation of \structboost  \eqref{eq:structboost1} by
defining
\[
    \w= - \w^{(1)} \odot \w^{(2)},
\]
    and, %
\begin{align*}
 \wstructs (\x, \y) & = \sum_{p \in \cal N} \wls^{(1)} \left( \U(y^p, \x )
\right)
  \odot   \sum_{(p,q) \in \cal S} \wls^{(2)} \left( \V( y^p, y^q, \x ) \right).
\end{align*}
Recall that $\odot$ stacks two vectors. With this definition, we have the relation:
\[
    \w^\T \wstructs(\x, \y)=-E(\x,\y ; \w).
\]
The minus sign here is to inverse the order of subtraction in \eqref{EQ:1b}.
At each column generation iteration (Algorithm \ref{ALG:alg1}), two new weak learners
$\wl^{(1) } (\cdot) $ and $\wl^{(2) } (\cdot) $ are added to the unary
weak learner set and the pairwise weak learner set,
respectively by solving the $\argmax$ problem defined in \eqref{EQ:WL2},
which can be written as:
 \begin{align}
        \label{eq:seg_wl}
         \wl^{(1)\star}  ( \cdot )
        =& \argmax_{ \wl( \cdot) } \sum_{i, \y  } \mu_{ ( i, \y ) }
	 \sum_{p \in \cal N} \Bigl[ \wl^{(1)} \left( \U( y^{p}, \x_i ) \right)\notag \\
     	& - \wl^{(1)} \left( \U( y^{p}_i, \x_i \right) \Bigr]; \\
        \wl^{(2)\star} ( \cdot )
        =& \argmax_{ \wl( \cdot ) }\sum_{i, \y  } \mu_{ ( i, \y ) }
	\sum_{(p,q) \in \cal S} \Bigl[ \wl^{(2)} \left(\V( y^{p}, y^{q}, \x_i) \right) \notag \\
        & -\wl^{(2)} \left( \V( y_i^{p}, y_i^{q},  \x_i) \right) \Bigr].
    \end{align}
The maximization problem to find the most violated constraint in \eqref{eq:infer} is to solve the inference:
\begin{align}
\label{eq:seg_infer3}
 \y'_i=\argmin_{\y} E(\x_i,\y) - \loss ( \y_i, \y),
\end{align}
which is similar to the label prediction inference in \eqref{eq:seg_predict},
 and the only difference is that the
labeling loss term: $\loss ( \y_i, \y)$ is involved in \eqref{eq:seg_infer3}.
Recall that we use the Hamming loss  $\loss ( \y_i, \y)$ as defined in \eqref{eq:hamming},
the term $\loss ( \y_i, \y)$ can be
absorbed into the unary term of the energy function defined in
\eqref{eq:seg_energy} (such as in \cite{SzummerKH08}).
The inference in \eqref{eq:seg_infer3} can be written as:
\begin{align}
\label{eq:seg_infer4}
    \y_i' = \argmin_{\y}  &  \sum_{p \in \cal N}
    \biggr[ \w^{(1)\T} \wls^{(1)} \left( \U(y^{p}, \x_i) \right) -    ( 1 -
    \Ind ( y_i^{p} ,   y^{p})  )
    \biggr] \notag \\
     & +  \sum_{(p,q) \in \cal S} \w^{(2)\T} \wls^{(2)} \left( \V( y^{p}, y^{q}, \x_i)
     \right). %
\end{align}
The above minimization \eqref{eq:seg_infer4}
can also  be solved efficiently by using graph cuts.

%% file: mc_exp.tex
\begin{table}
  \caption{
      AUC maximization. We compare the
    performance of \mslack \ and \oneslack\
    formulations.
    ``$-$'' means that the method is not able to converge within a memory
    and time limit. We can see that
    \oneslack\ can achieve similar
    AUC results on training and testing data as \mslack while \oneslack is
    significantly faster than \mslack.
  }
  \centering
  \resizebox{.998\linewidth}{!}
  {
  \begin{tabular}{ r | l l l l  }
    \hline
    dataset & method & time (sec) & AUC training  & AUC test \\ \hline \hline

    \multirow{2}{*}{\sf wine}
    & \mslack & 13$\pm$1  & 1.000$\pm$0.000 & 0.994$\pm$0.005 \\
    & \oneslack & 3$\pm$1 & 1.000$\pm$0.000 & 0.994$\pm$0.006 \\  \hline

    \multirow{2}{*}{\sf glass}
    & \mslack & 20$\pm$1  & 0.967$\pm$0.011 & 0.849$\pm$0.028 \\
    & \oneslack & 6$\pm$1 & 0.955$\pm$0.030 & 0.844$\pm$0.039 \\  \hline

    \multirow{2}{*}{\sf svmguide2}
    & \mslack & 332$\pm$6  & 0.988$\pm$0.003 & 0.905$\pm$0.036 \\
    & \oneslack & 106$\pm$8 & 0.988$\pm$0.003 & 0.905$\pm$0.036 \\  \hline

    \multirow{2}{*}{\sf svmguide4}
    & \mslack & 564$\pm$79  & 1.000$\pm$0.000 & 0.982$\pm$0.005 \\
    & \oneslack & 106$\pm$13 & 1.000$\pm$0.000 & 0.982$\pm$0.005 \\  \hline

    \multirow{2}{*}{\sf vowel}
    & \mslack & 4051$\pm$116  & 0.999$\pm$0.001 & 0.968$\pm$0.013 \\
   & \oneslack & 952$\pm$139 & 0.999$\pm$0.001 & 0.967$\pm$0.013 \\  \hline

    \multirow{2}{*}{\sf dna}
    & \mslack &$ -$ &$ -$ &$ -$ \\
    & \oneslack & 1598$\pm$643 & 0.998$\pm$0.000 & 0.992$\pm$0.003 \\  \hline

    \multirow{2}{*}{\sf segment}
    & \mslack &$ -$ &$ -$ &$ -$ \\
    & \oneslack & 475$\pm$42 & 1.000$\pm$0.000 & 0.999$\pm$0.001 \\  \hline

    \multirow{2}{*}{\sf satimage}
     & \mslack &$ -$ &$ -$ &$ -$  \\
     & \oneslack & 37769$\pm$6331 & 0.999$\pm$0.000 & 0.997$\pm$0.002 \\ \hline

  \end{tabular}
  }
  \label{TAB:AUC1}
\end{table}

\begin{figure*}[t]
    \centering
    \includegraphics[width=.245\linewidth]{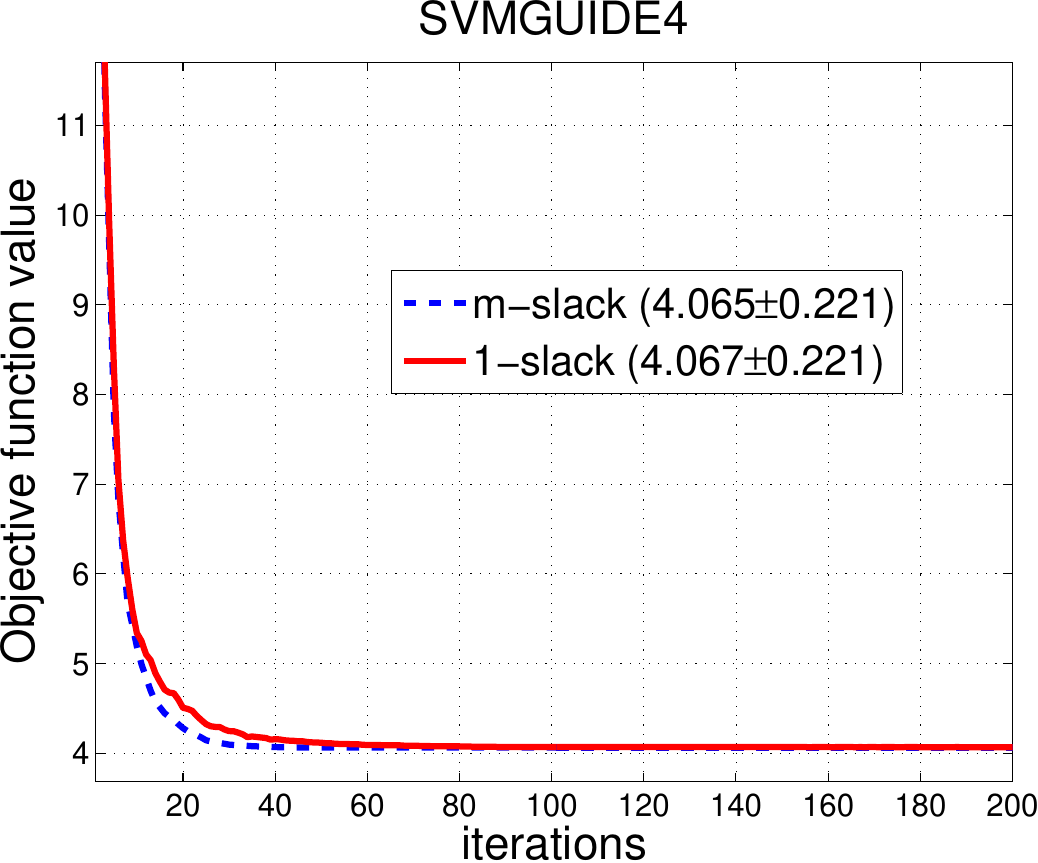}
    \includegraphics[width=.245\linewidth]{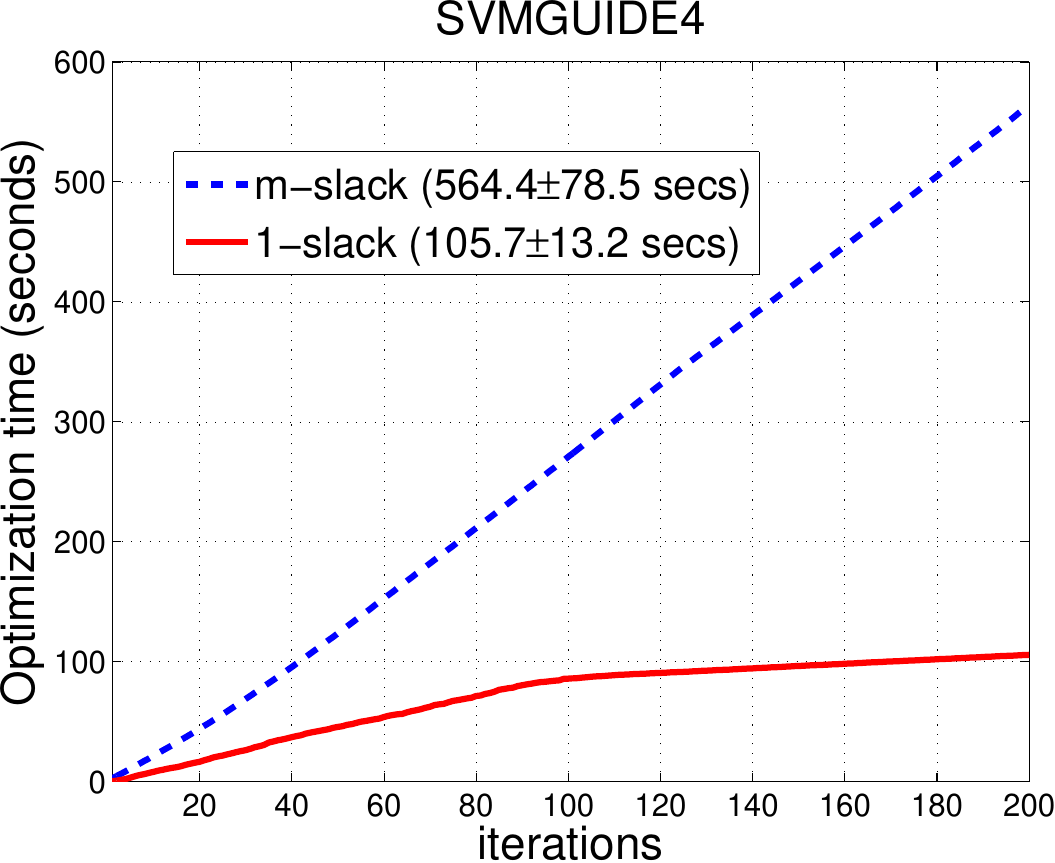}
    \includegraphics[width=.245\linewidth]{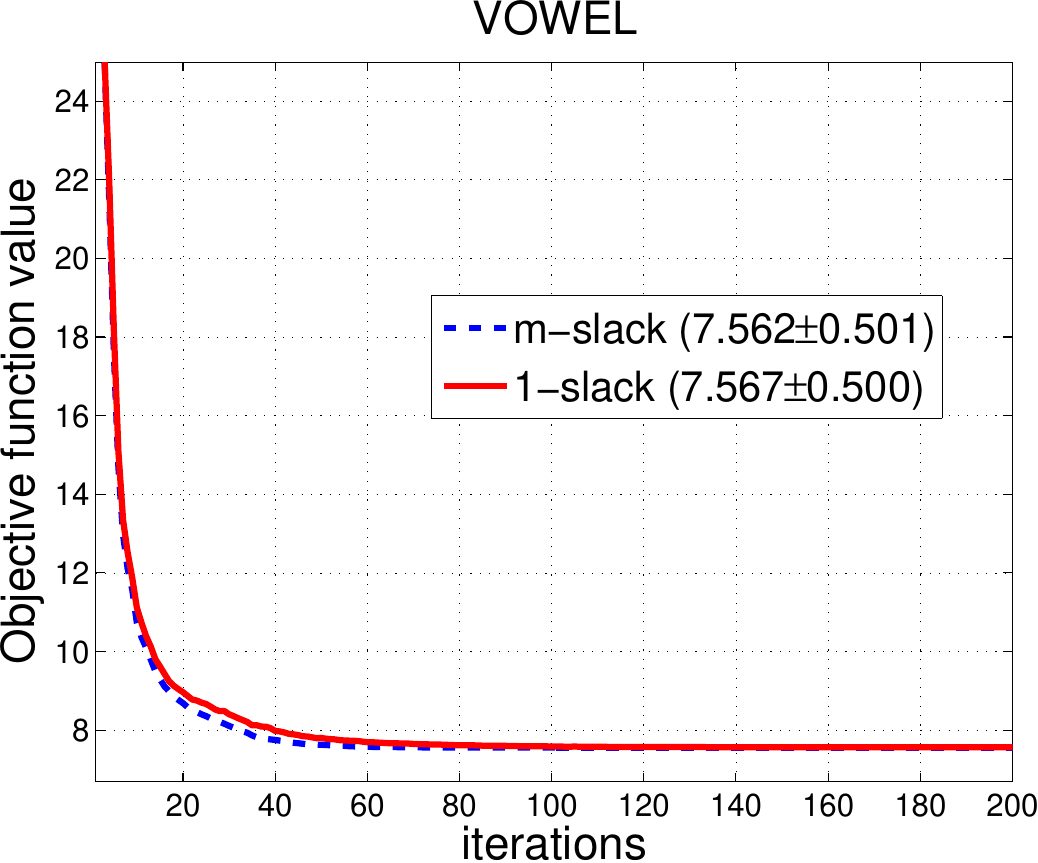}
    \includegraphics[width=.245\linewidth]{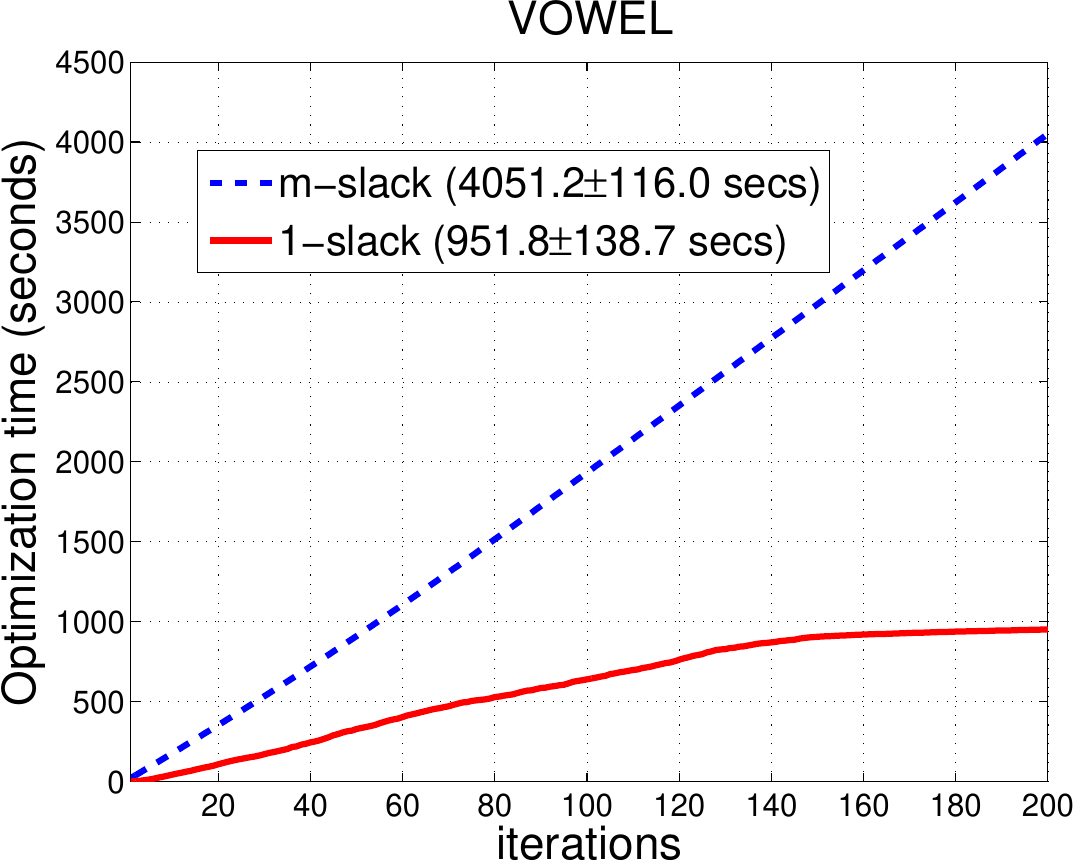}
    \caption{AUC optimization on two UCI datasets. The objective values and optimization time
are shown in the figure by varying boosting (or column generation) iterations. It shows that \oneslack achieves similar
objective values as \mslack but needs less running time.}
    \label{fig:auc_iters}
\end{figure*}

\paragraph{AUC optimization}

    In this experiment, we compare efficiency of the \oneslack (solving \eqref{eq:structboost-oneslack})
    and \mslack (solving \eqref{eq:structboost1} or its
             dual) formulations of our method \structboost.
    The details for AUC optimization are described in Section \ref{sec:auc}.
    We run the experiments on a few UCI multi-class datasets.
             To create
             imbalanced data, we use one class of the
             multi-class UCI datasets as positive data and  the
             rest   as negative data.
The boosting (column generation) iteration is set to 200;
the cutting-plane stopping criterion  ($\epsilon_{\rm cp}$) is
set to 0.001.
Decision stumps are used as weak learners ($\wls(\cdot)$ in \eqref{eq:rank1}).
For each data set,
we randomly sample 50\% for training, 25\% for validation and the rest for testing.
The regularization parameter $C$ is chosen from 6 candidates ranging from $10$ to $10^3$.
Experiments are repeated  5 times on  each dataset and the mean and standard deviation are reported.
Table \ref{TAB:AUC1} reports the results. We can see that the \oneslack\
             formulation of \structboost achieves similar AUC performance
             as the \mslack formulation, while being much faster than  \mslack.
The values of objective function and optimization time are shown in Figure \ref{fig:auc_iters} by varying
column generation iterations. Again, it shows that \oneslack achieves similar objective values
as \mslack with less running time.

             Note that RankBoost may
             also be applied to this problem \cite{Freund2003}.
             RankBoost has been designed for solving ranking problems.

\begin{table*}
\caption{Multi-class classification test errors (\%) on several UCI
 and MNIST datasets. StructBoost-stump and StructBoost-per denote our \structboost using decision stumps
and perceptrons as weak learners, respectively. \structboost outperforms \ssvm in most cases and
achieves competitive performance compared with other multi-class classifiers.
 }
\centering
\resizebox{1\linewidth}{!}
  {
  \begin{tabular}{ r | c c c c c c c c c c}
    \hline
     &  glass  & svmguide2 & svmguide4 & vowel &dna &segment & satimage & usps & pendigits & mnist \\ \hline \hline
StructBoost-stump	&35.8 $\pm$ 6.2	&21.0 $\pm$ 3.9	&20.1 $\pm$ 2.9	&17.5 $\pm$ 2.2	& \bf 6.2 $\pm$ 0.7	& \bf 2.9 $\pm$ 0.7	&12.1 $\pm$ 0.7	&6.9 $\pm$ 0.6	&3.9 $\pm$ 0.4	&12.5 $\pm$ 0.4\\
StructBoost-per	&37.3 $\pm$ 6.2	&22.7 $\pm$ 4.8	&53.4 $\pm$ 6.1	& \bf 6.8 $\pm$ 1.8	&6.6 $\pm$ 0.6	&3.8 $\pm$ 0.7	& \bf 11.4 $\pm$ 1.1	& \bf 4.1 $\pm$ 0.6	& \bf 1.8 $\pm$ 0.3	& \bf 6.5 $\pm$ 0.6\\
Ada.ECC	&32.7 $\pm$ 4.9	&23.3 $\pm$ 4.0	& \bf 19.1 $\pm$ 2.3	&20.6 $\pm$ 1.5	&7.6 $\pm$ 1.2	& \bf 2.9 $\pm$ 0.8	&12.8 $\pm$ 0.7	&8.4 $\pm$ 0.7	&8.4 $\pm$ 0.7	&15.8 $\pm$ 0.3\\
Ada.MH	& \bf 32.3 $\pm$ 5.0	&21.9 $\pm$ 4.5	&19.3 $\pm$ 3.0	&18.8 $\pm$ 2.1	&7.1 $\pm$ 0.6	&3.7 $\pm$ 0.7	&12.7 $\pm$ 0.9	&7.4 $\pm$ 0.5	&7.4 $\pm$ 0.5	&13.4 $\pm$ 0.4\\
SSVM	&38.8 $\pm$ 8.7	&21.9 $\pm$ 5.9	&45.7 $\pm$ 3.9	&25.6 $\pm$ 2.5	&6.9 $\pm$ 0.9	&5.3 $\pm$ 1.0	&14.9 $\pm$ 0.1	&5.8 $\pm$ 0.3	&5.2 $\pm$ 0.3	&9.6 $\pm$ 0.2\\
1-vs-all SVM	&40.8 $\pm$ 7.0	& \bf 17.7 $\pm$ 3.5	&47.0 $\pm$ 3.2	&54.4 $\pm$ 2.2	&6.3 $\pm$ 0.5	&7.7 $\pm$ 0.8	&17.5 $\pm$ 0.4	&5.4 $\pm$ 0.5	&8.1 $\pm$ 0.5	&9.2 $\pm$ 0.2\\
    \hline
  \end{tabular}
  }

\label{tab:mc}
\end{table*}

\begin{figure*}
    \centering
    \includegraphics[width=.27\linewidth]{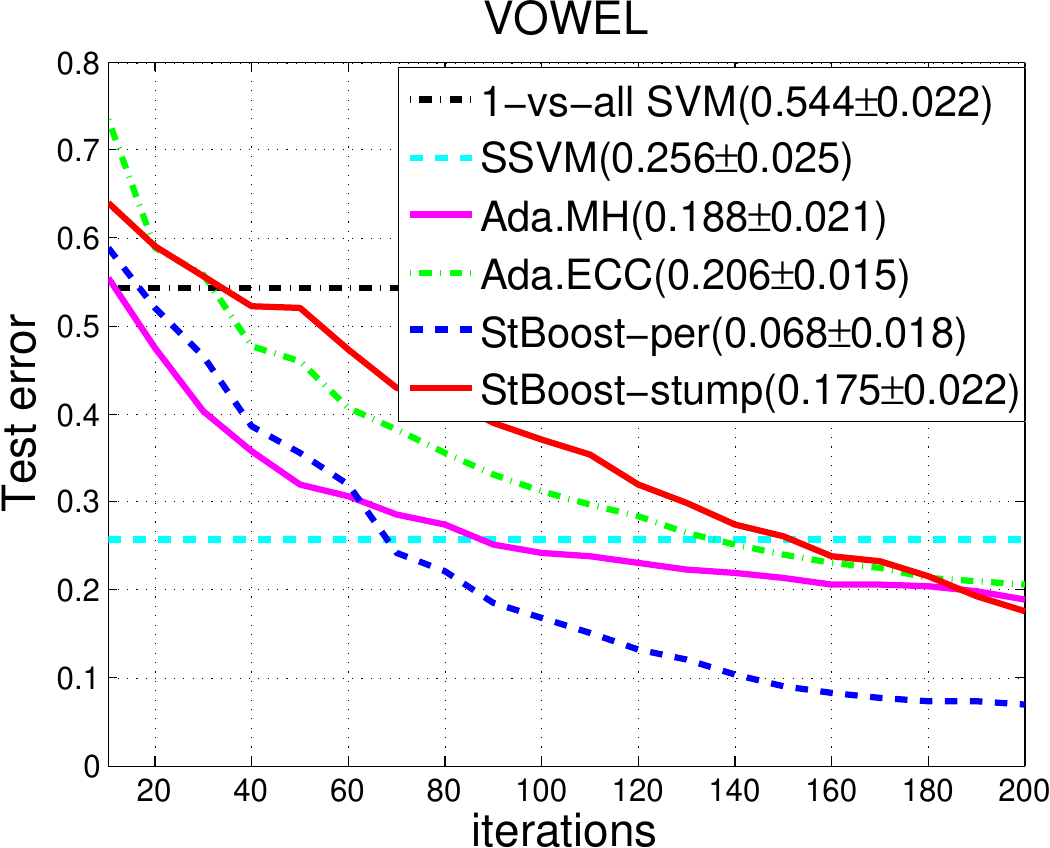}
     \includegraphics[width=.27\linewidth]{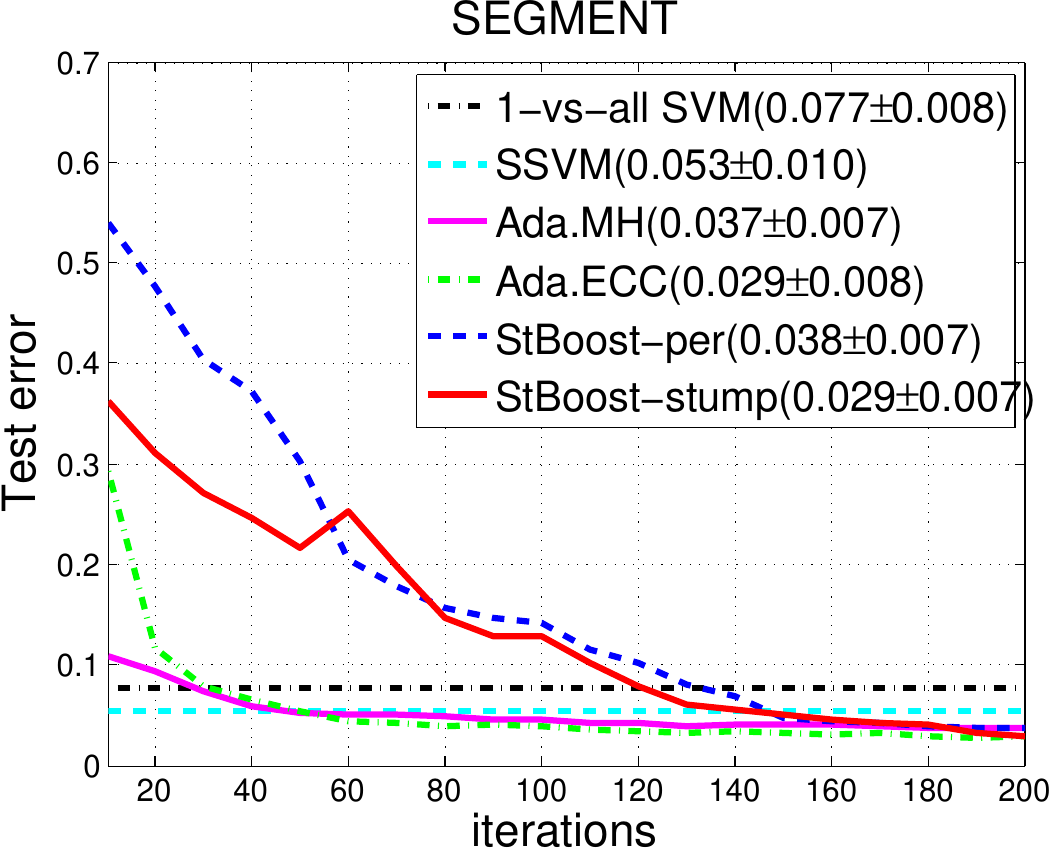}
     \includegraphics[width=.27\linewidth]{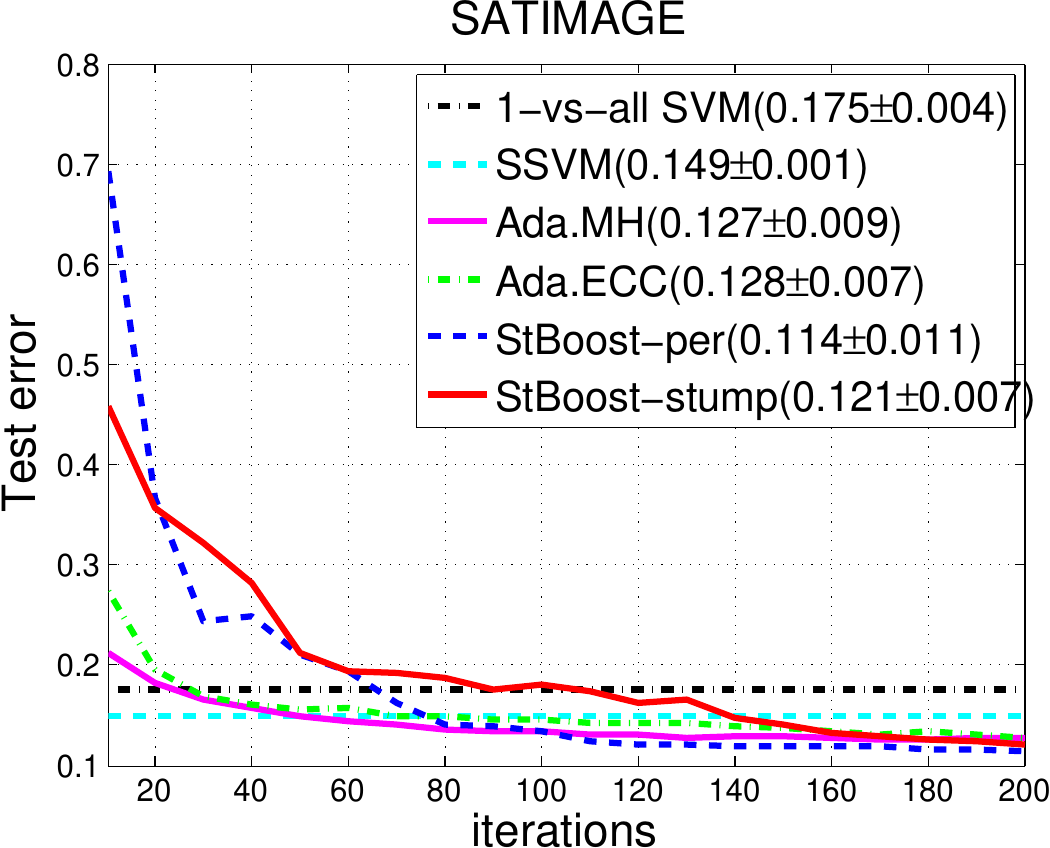}

     \vspace{0.2cm}

     \includegraphics[width=.27\linewidth]{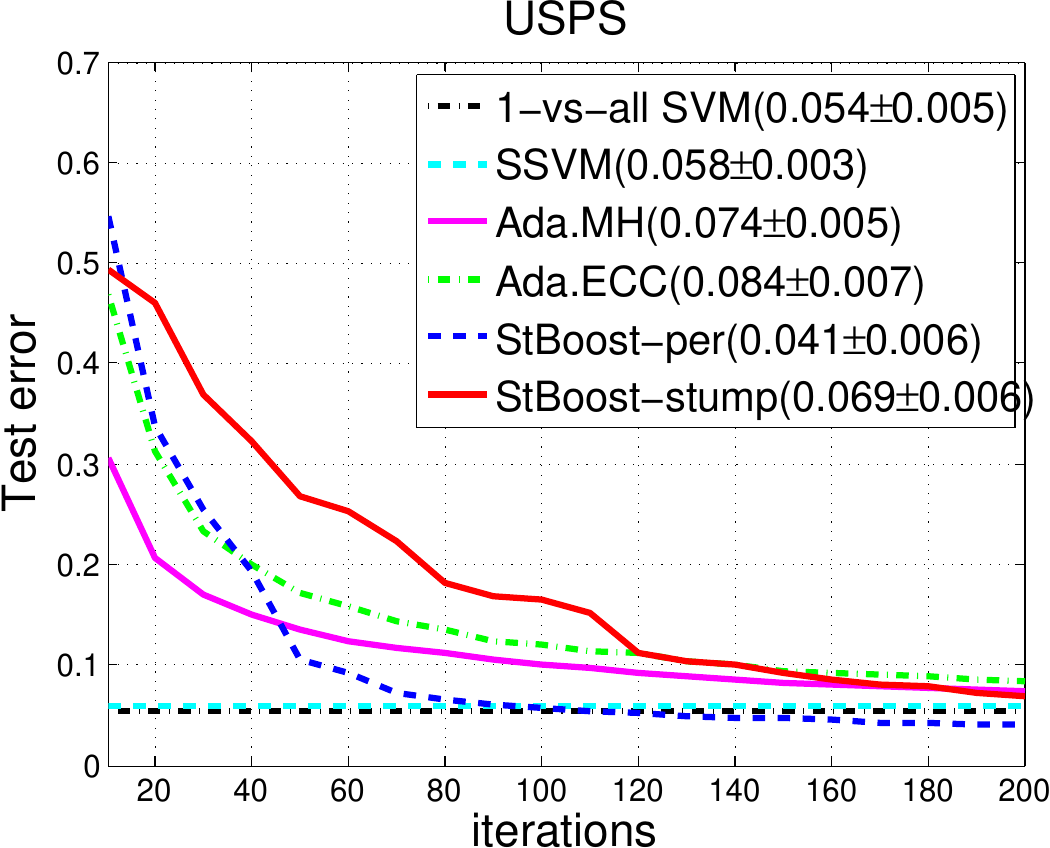}
     \includegraphics[width=.27\linewidth]{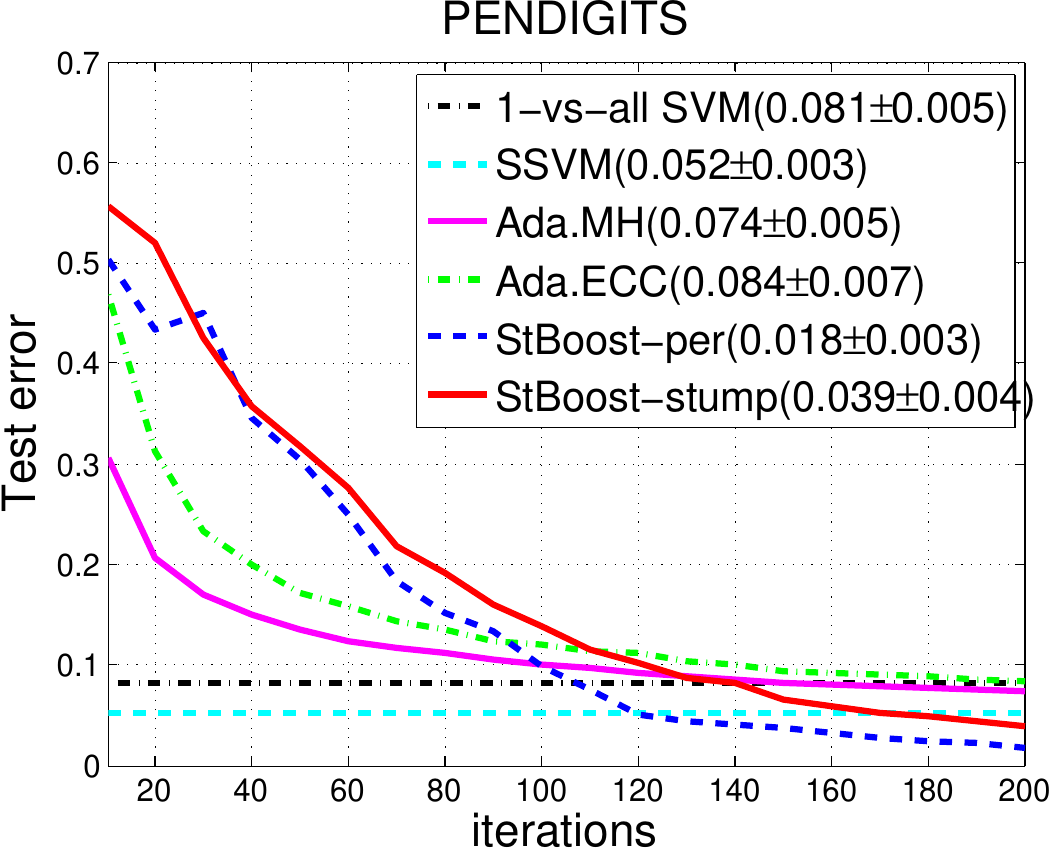}
     \includegraphics[width=.27\linewidth]{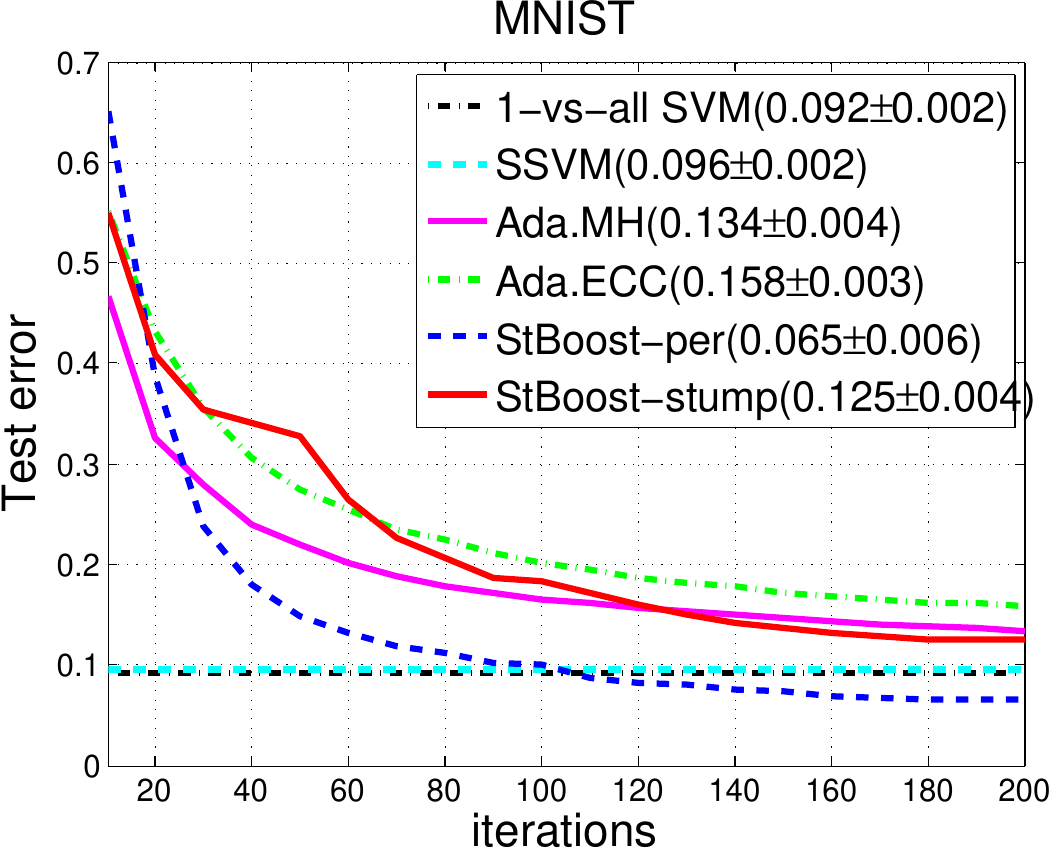}
     \caption{Test performance versus the number of
         boosting iterations of multi-class classification.
       StBoost-stump and StBoost-per denote our
       \structboost using decision stumps and perceptrons as weak learners,
       respectively.  The results of SSVM and SVM are shown as
       straight lines in the plots. The values shown in the legend are
       the error rates of the final iteration for each method. Our
       methods perform better than SSVM in most cases.
       }
    \label{fig:mc}
\end{figure*}

\paragraph{Multi-class classification}

    Multi-class classification is a special case of structured learning. Details
are described in Section \ref{sec:mc}. We carry out experiments on some UCI
multi-class datasets and MNIST. We compare with Structured SVM (\ssvm),
conventional multi-class boosting methods (namely AdaBoost.ECC and AdaBoost.MH),
and the one-vs-all SVM method. For each dataset, we randomly select 50\% data for training, 25\%
data for validation and the rest  for testing. The maximum number of boosting
iterations is set to 200; the regularization parameter $C$ is chosen from 6 candidates whose values range from
10 to $1000$. The experiments are repeated 5 times for each dataset.

To demonstrate the flexibility of our method, we use decision stumps and linear perceptron functions as
weak learners ($\wls(\cdot)$ in \eqref{eq:mc}). The perceptron weak learner can be written as:
\begin{align}
\wl(\x)=\mathrm{sign}(\v^\T\x+b).
\label{eq:wlper}
\end{align}
We use a smooth sigmoid function to replace the step function
so that gradient descent optimization can be applied.
We solve the $\argmax$ problem in \eqref{eq:wl_mc} by using the Quasi-Newton LBFGS \cite{lbfgs}
solver.
We found that  decision stumps often provide a good initialization for LBFGS learning of
the perceptron.
Compared with decision stumps,
using the perceptron weak learner usually needs fewer boosting iterations to converge.
Table \ref{tab:mc} reports the error rates.
Figure \ref{fig:mc} shows test  performance versus the number of boosting (column generation) iterations.
The results demonstrate that our method  outperforms \ssvm, and
achieves competitive performance compared with  other conventional
multi-class methods.

StructBoost performs better than \ssvm on most datasets. This might be due to the introduction of
non-linearity in StructBoost.
Results also show that using the perceptron weak learner often
achieves better performance than using decision stumps on those large datasets.

%% file: hmc_exp.tex
\begin{table*}
\caption{Hierarchical classification. Results of the tree loss and the 1/0 loss (classification error rate)
on subsets of the SUN dataset. %
StructBoost-tree uses the hierarchy class formulation with the tree loss,
and StructBoost-flat uses the standard multi-class formulation.
StructBoost-tree that minimizes the tree loss performs best.
}
\centering
  {
  \begin{tabular}{ c  r | c c c c c }
    \hline
     datasets &  &  StructBoost-tree  & StructBoost-flat  & Ada.ECC-SVM
      &  Ada.ECC & Ada.MH  \\ \hline \hline
    \multirow{2}{*}{\sf 6 scenes} & 1/0 loss & 0.322 $\pm$ 0.018 & 0.343 $\pm$ 0.028 & 0.350 $\pm$ 0.013  & 0.327 $\pm$ 0.002 & \bf 0.315 $\pm$ 0.015 \\
				 & tree loss & \bf 0.337 $\pm$ 0.014 & 0.380 $\pm$ 0.027 & 0.377 $\pm$ 0.018  & 0.352 $\pm$ 0.023 & 0.346 $\pm$ 0.018 \\ \hline
    \multirow{2}{*}{\sf 15 scenes} & 1/0 loss  & \bf 0.394 $\pm$ 0.005 & 0.396 $\pm$ 0.013 & 0.414 $\pm$ 0.012  & 0.444 $\pm$ 0.012 & 0.418 $\pm$ 0.010 \\
				 & tree loss  & \bf 0.504 $\pm$ 0.007 & 0.536 $\pm$ 0.009 & 0.565 $\pm$ 0.019  & 0.584 $\pm$ 0.017 & 0.551 $\pm$ 0.013 \\ \hline
  \end{tabular}
  }
\label{TAB:tree_loss}
\end{table*}

\subsection{Hierarchical multi-class classification}
The details of hierarchical multi-class are described in Section \ref{sec:hmc}.
    We have constructed two hierarchical image datasets from the SUN
    dataset \cite{SUN} which contains 6 classes and 15 classes of scenes. The hierarchical tree structures of these two
     datasets are shown in the Figure \ref{fig:h_structure}.
	 For each scene class, we use the first 200 images from the original SUN dataset. There are 1200 images in
    the first dataset and 3000 images in the second dataset. We have used the HOG features as described in \cite{SUN}.
For each dataset, 50\% examples are randomly selected for training and the rest for testing. We run 5 times for each dataset.
   The regularization parameter is chosen from 6 candidates ranging from 1 to $10^3$.

We use linear SVM as weak classifiers in our method.
The linear SVM weak classifier has the same form as \eqref{eq:wlper}.
At each boosting iteration, we solve the $\argmax$ problem  by training a linear SVM model.
The regularization parameter $C$ in SVM is set to a large value ($10^8$/\#examples).
To alleviate the overfitting problem of using linear SVM as weak learners,
we set 10\% of the smallest non-zero dual solutions $\mu_{(i,\y)}$  to zero.

We compare \structboost using the hierarchical multi-class formulation (StructBoost-tree) and the standard multi-class formulation (StructBoost-flat).
We run some other multi-class methods for further comparison: Ada.ECC, Ada.MH with decision stumps.
We also run Ada.ECC using linear SVM as weak classifiers (labelled as Ada.ECC-SVM).

When using SVM as weak learners, the number of boosting iterations is set to 200,
and for decision stump, it is set to 500.
 Table \ref{TAB:tree_loss} shows the tree loss and the 1/0 loss (classification error rate).
We observe that StructBoost-tree has the lowest  tree loss among all  compared methods, and it also
improves its counterpart, StructBoost-flat, in terms of both classification error rates and the tree loss.
Our StructBoost-tree makes use of the class hierarchy information and directly optimizes the tree loss.
That might be the reason why StructBoost-tree  achieves best  performance.

%% file: tracking_exp.tex
\paragraph{Visual tracking}

    A visual tracking method, termed Struck \cite{Torr2011ICCV},
    was introduced based on \svmstruct. The  core idea is to train a
    tracker by optimizing the
    Pascal image overlap score using \svmstruct.
    Here we apply \structboost to visual tracking, following the similar setting as in
    Struck \cite{Torr2011ICCV}.
    More details can be found in Section \ref{sec:track}.

    Our experiment follows the on-line tracking setting. Here we  use the first 3 labeled frames
    for initialization and
    training of our \structboost tracker.
    Then the tracker is updated by re-training the model during the course of tracking.
    Specifically, in the $i$-th frame (represented by $\x_i$), we first perform a
    prediction step (solving \eqref{eq:predict}) to output the detection box ($\y_i$),
    then collect training
    data for tracker update. For solving the prediction
    inference in \eqref{eq:predict}, we simply sample about 2000
    bounding boxes around the prediction box of the last frame (represented by
    $ \y_{i-1}$),  and search for the most confident bounding box over all
    sampled boxes as the prediction.
	After the prediction, we collect training data by
    sampling about 200 bounding boxes around the current prediction
    $\y_i$. We use the training data in recent 60 frames to re-train
    the tracker for every 2 frames.
    Solving \eqref{eq:infer} for finding the most violated constraint is similar to the prediction
    inference.

    For \structboost, decision stumps are used as the weak classifiers;
    the number of boosting iterations is set
        to 300; the regularization parameter $C$ is selected from
        $10^{0.5}$ to $10^2$. We use the down-scaled gray-scale raw
        pixels and HOG \cite{Felzenszwalb2010} as image features.

For comparison, we also run a simple binary AdaBoost tracker using the same
    setting as our \structboost tracker. The
    number of weak classifiers for AdaBoost is set to 500.
    When training the AdaBoost
    tracker, we collect positive boxes that significantly
    overlap (overlap score above $0.8$) with the prediction box of the current frame,
	and negative boxes with small overlap scores (lower or equal to $0.3$).

We also compare with a few state-of-the-art tracking methods,
including Struck \cite{Torr2011ICCV} (with a buffer size of 50),
multi-instance tracking (MIL) \cite{MITracking},
fragment tracking (Frag) \cite{Frag},
online AdaBoost tracking (OAB) \cite{OAB},
and visual tracking decomposition (VTD) \cite{VTD}.
Two different settings are used for OAB: one positive
example per frame (OAB$_1$) and five positive examples per
frame (OAB$_5$) for training. The
test video sequences: ''coke, tiger1, tiger2, david, girl and sylv"
were used in \cite{Torr2011ICCV}; ``shaking, singer"
are from \cite{VTD} and the rest are from \cite{shu2011}.

Table \ref{Tab:exp-overlap} reports the Pascal overlap scores of compared methods.
{\em Our \structboost tracker performs best on most sequences}.
Compared with the simple binary AdaBoost tracker, \structboost that optimizes
the pascal overlap criterion perform significantly better.
Note that here Struck uses Haar features. When Struck uses a Gaussian
kernel defined on raw pixels, the performance is slightly different \cite{Torr2011ICCV},
and ours still outperforms Struck in most cases.
This might be due to the fact that \structboost selects relevant
features (300 features selected here), while SSVM of Struck \cite{Torr2011ICCV}
uses all the image patch
information which may contain noises.

The center location errors (in pixels) of compared methods are shown in Table
\ref{Tab:exp-center-error}. We can see that optimizing the overlap score also helps to
minimize the center location errors. Our method also achieves the best performance.

\begin{figure*}[t!]
    \centering

        \includegraphics[width=0.5\linewidth]{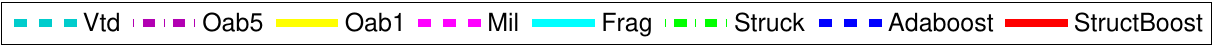}\vspace{2pt}

	\begin{subfigure}{.326\linewidth}
      		\includegraphics[width=\linewidth]{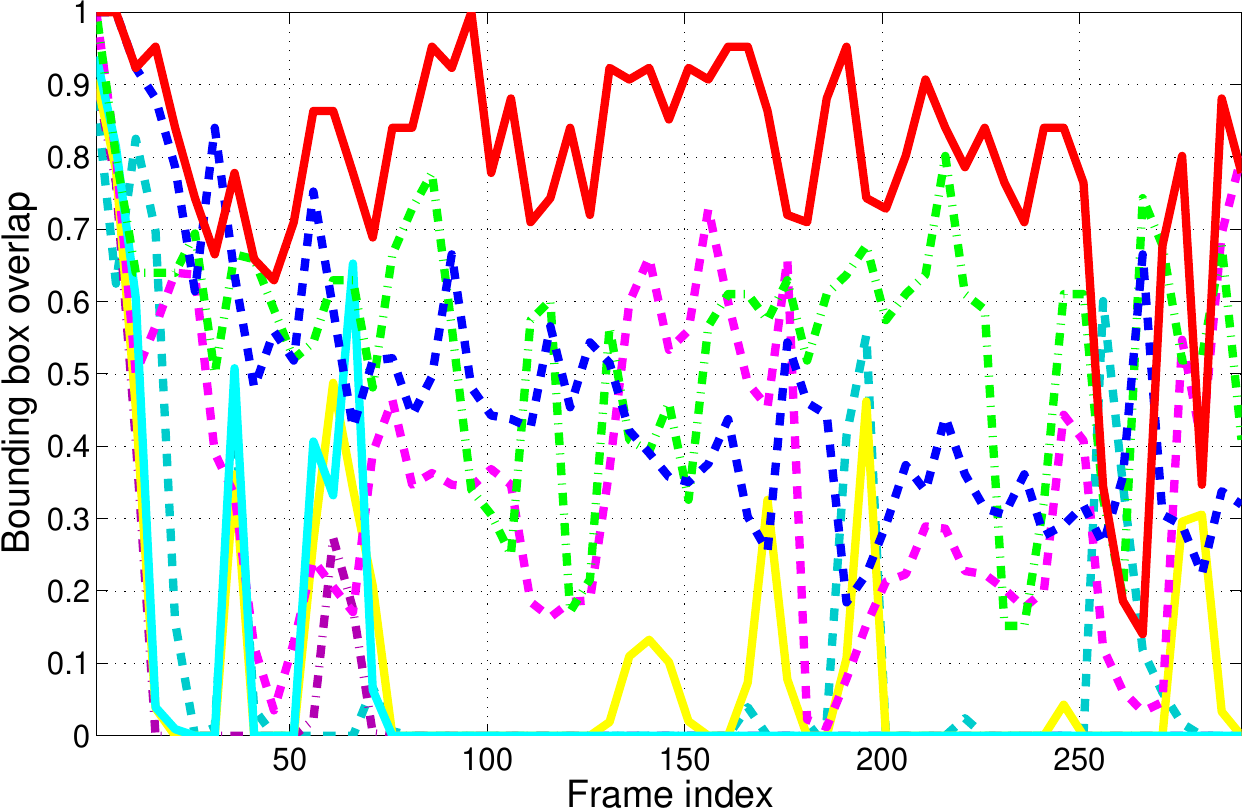}
		\caption{coke}
        \end{subfigure}
	\begin{subfigure}{.326\linewidth}
        	\includegraphics[width=\linewidth]{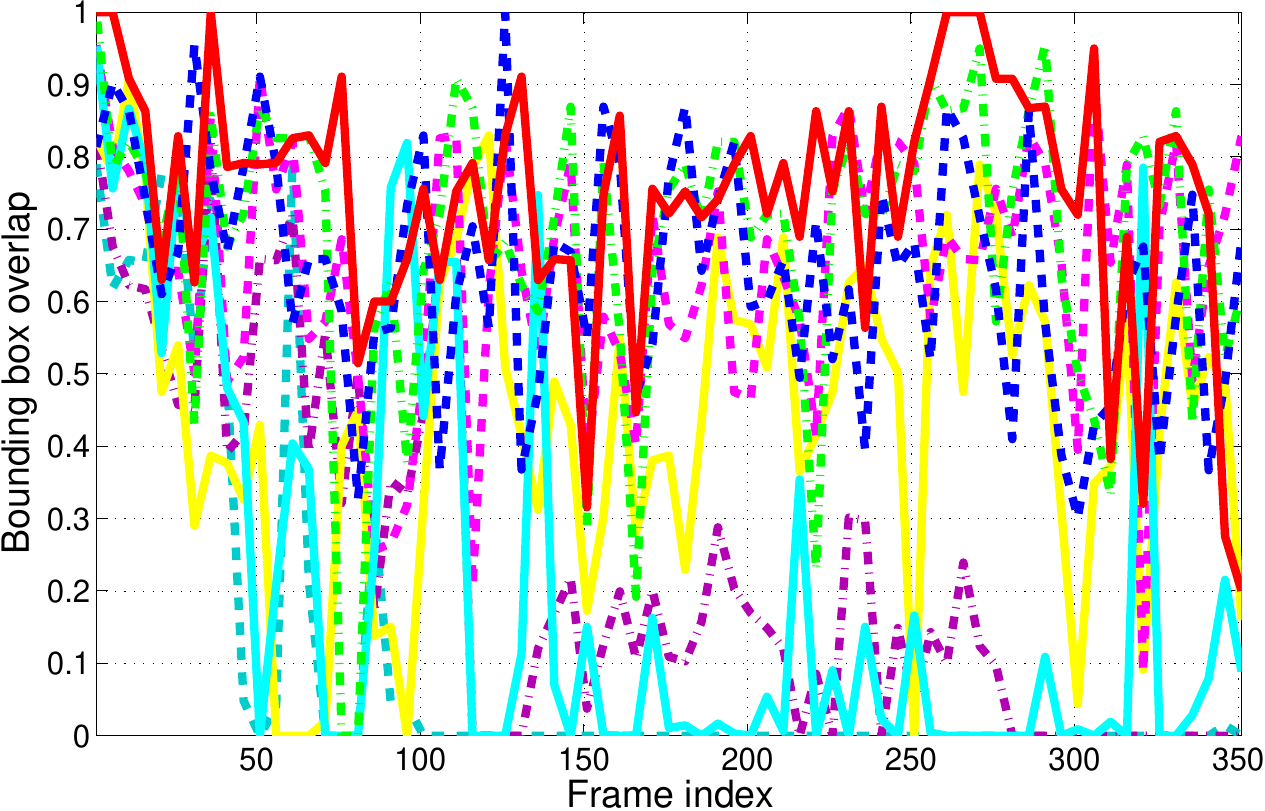}
	      	\caption{tiger1}
        \end{subfigure}
	\begin{subfigure}{.326\linewidth}
	        \includegraphics[width=\linewidth]{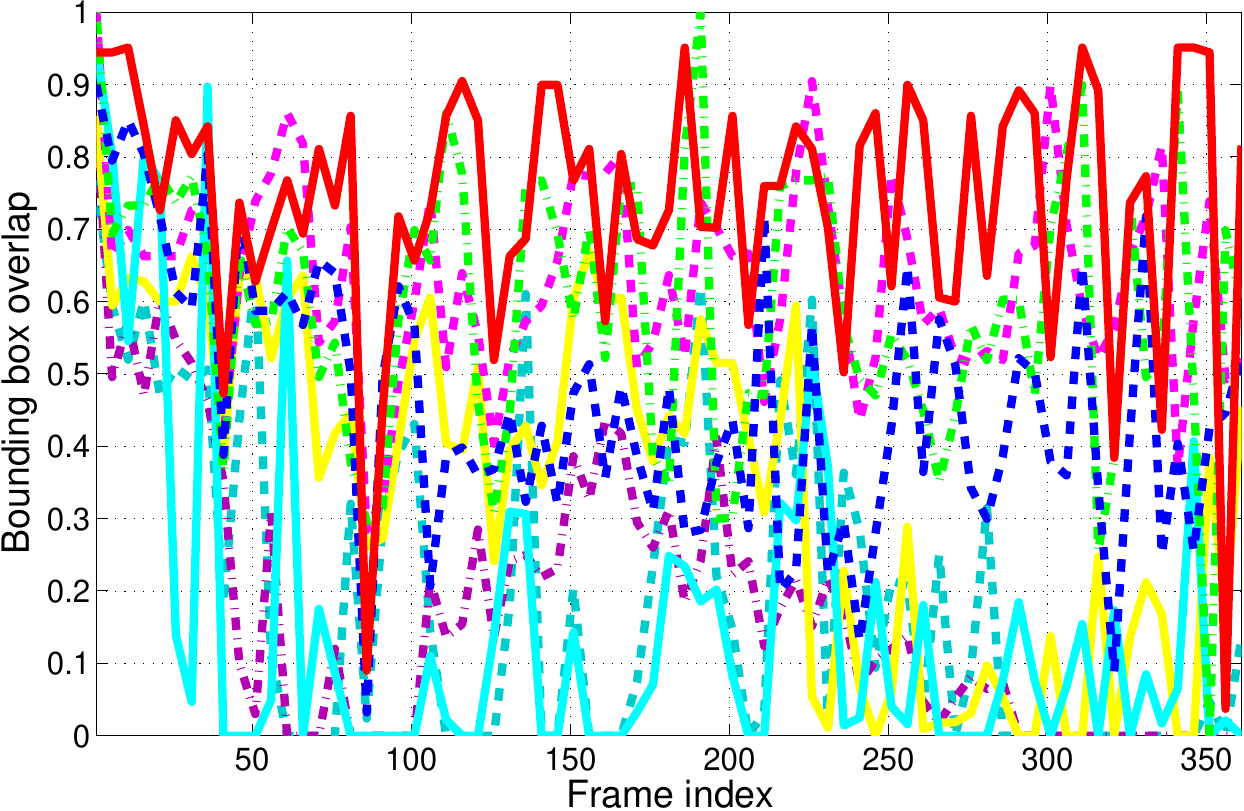}
		\caption{tiger2}
        \end{subfigure}

	\begin{subfigure}{.326\linewidth}
	        \includegraphics[width=\linewidth]{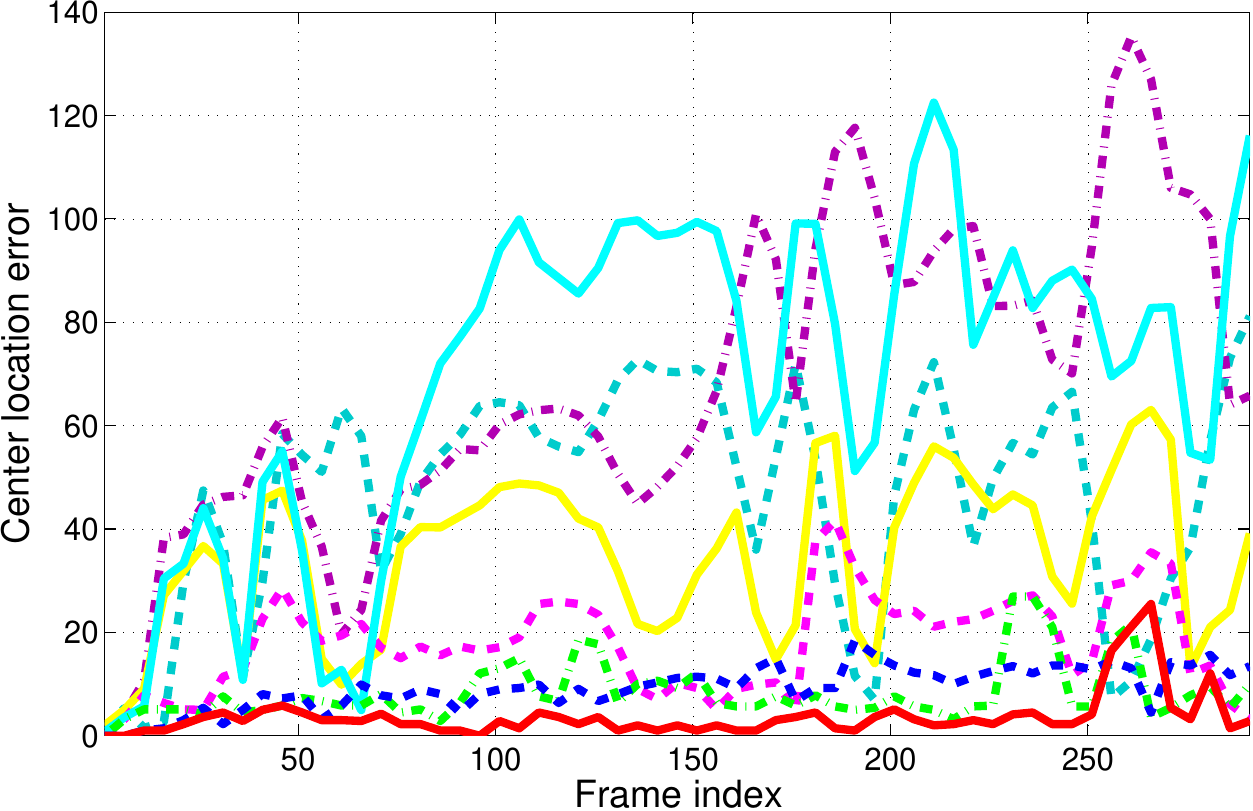}
        	\caption{coke}
        \end{subfigure}
	\begin{subfigure}{.326\linewidth}
	        \includegraphics[width=\linewidth]{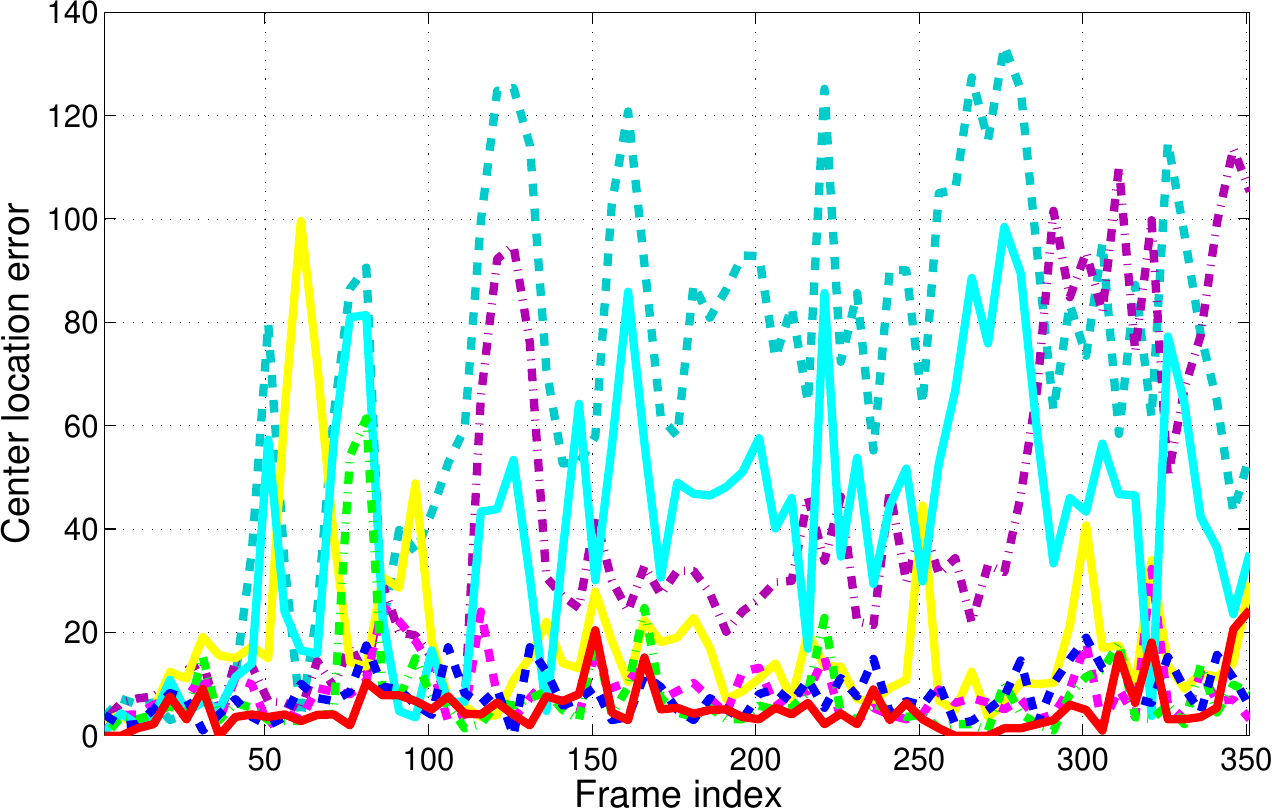}
        	\caption{tiger1}
        \end{subfigure}
	\begin{subfigure}{.326\linewidth}
	        \includegraphics[width=\linewidth]{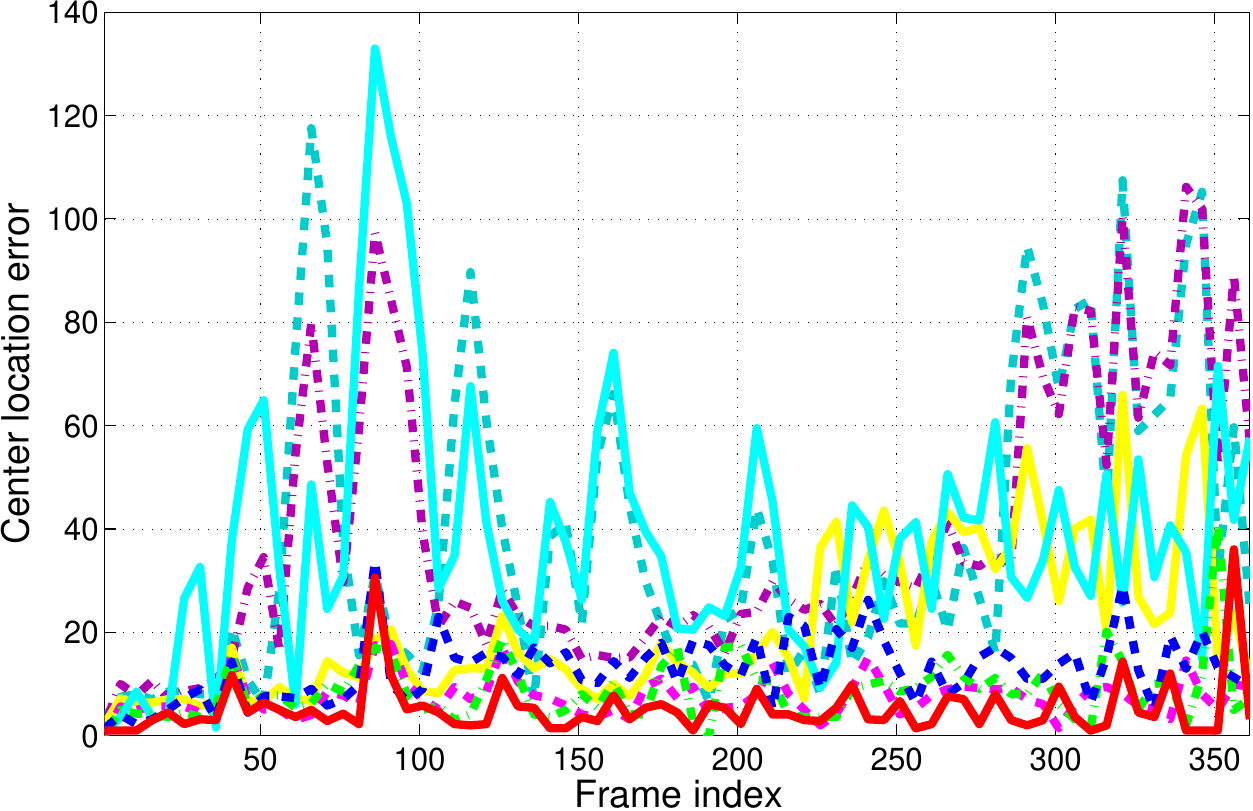}
        	\caption{tiger2}
        \end{subfigure}

    \caption{Bounding box overlap (first row) and center location error (second row) in frames of several video
    sequences. Our \structboost often achieves higher scores
    of box overlap and lower center location errors compared with
    other trackers.}
    \label{fig:frame_ovp_err}
\end{figure*}

\begin{figure*}
    \centering

        \includegraphics[width=0.6\linewidth]{track_fig/frames/legend}\vspace{2pt}

        \includegraphics[width=.177\linewidth]{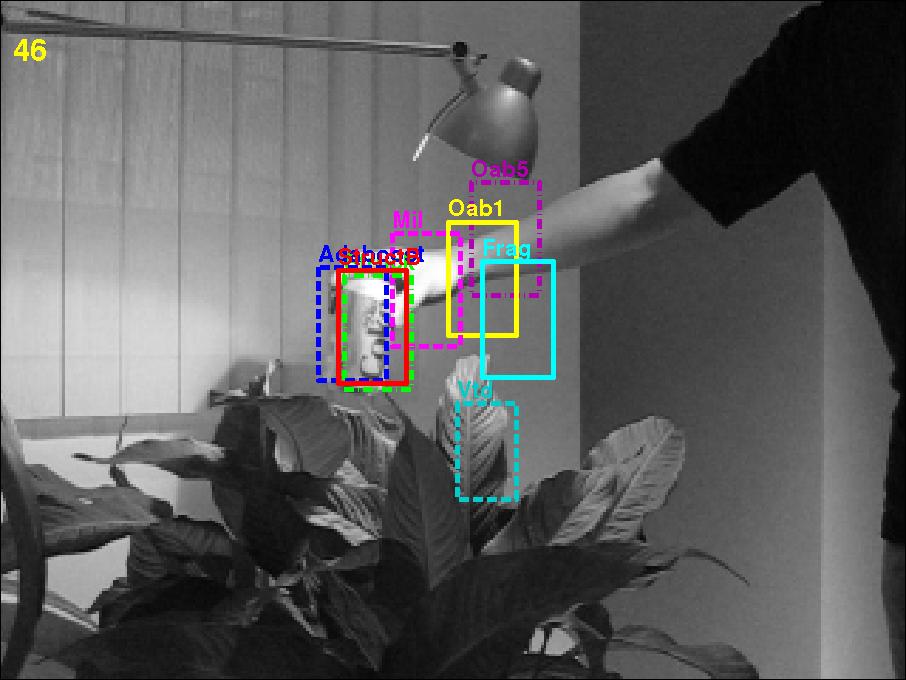}
        \includegraphics[width=.177\linewidth]{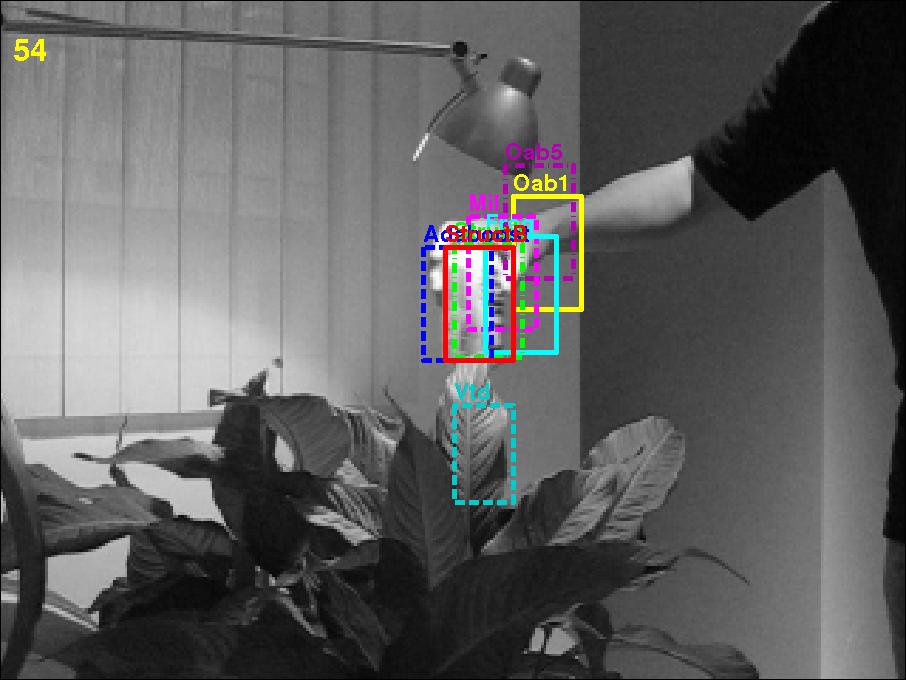}
        \includegraphics[width=.177\linewidth]{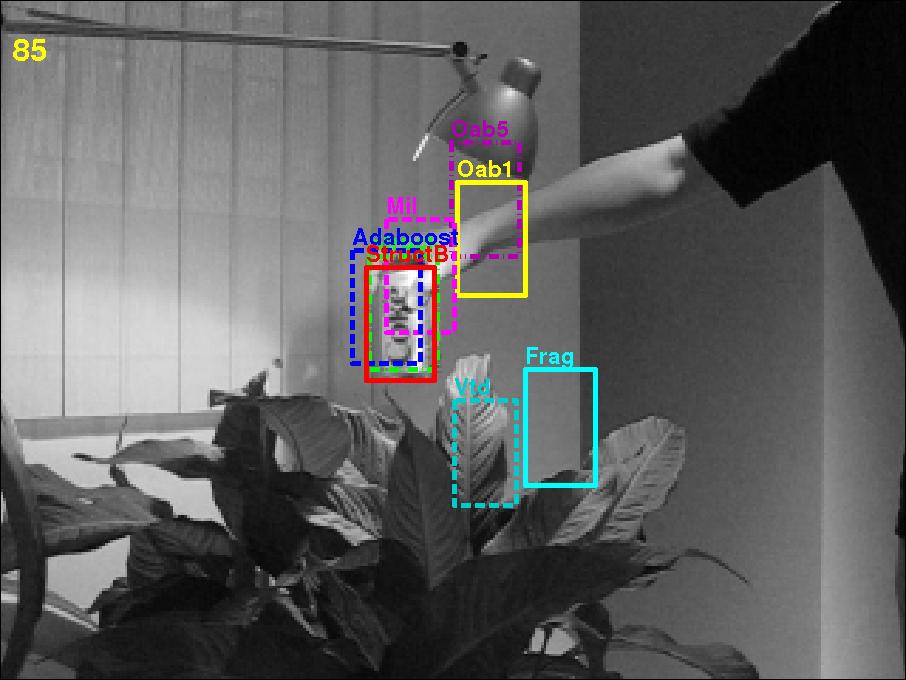}
        \includegraphics[width=.177\linewidth]{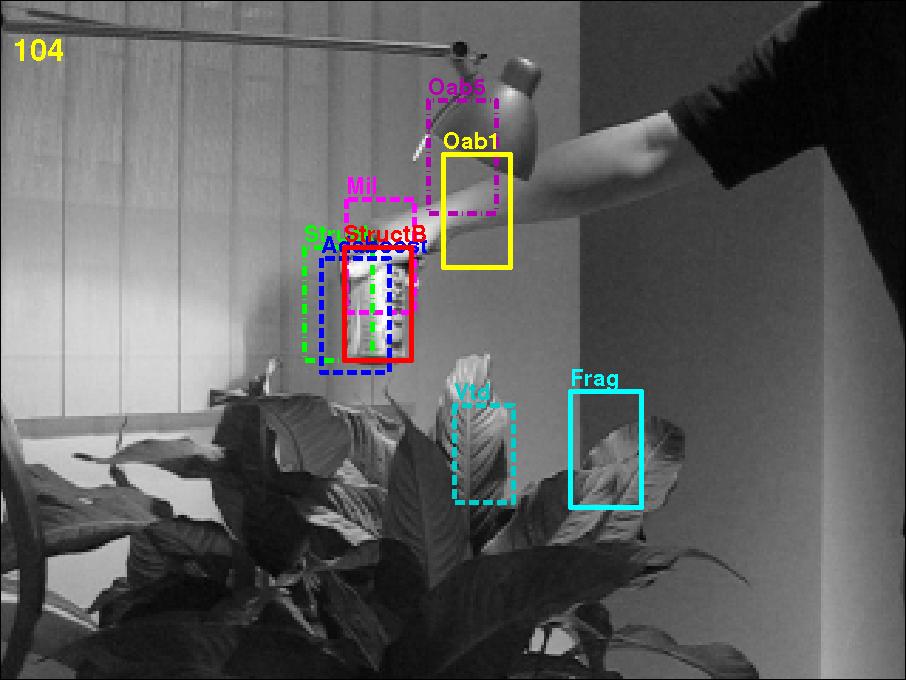}
        \includegraphics[width=.177\linewidth]{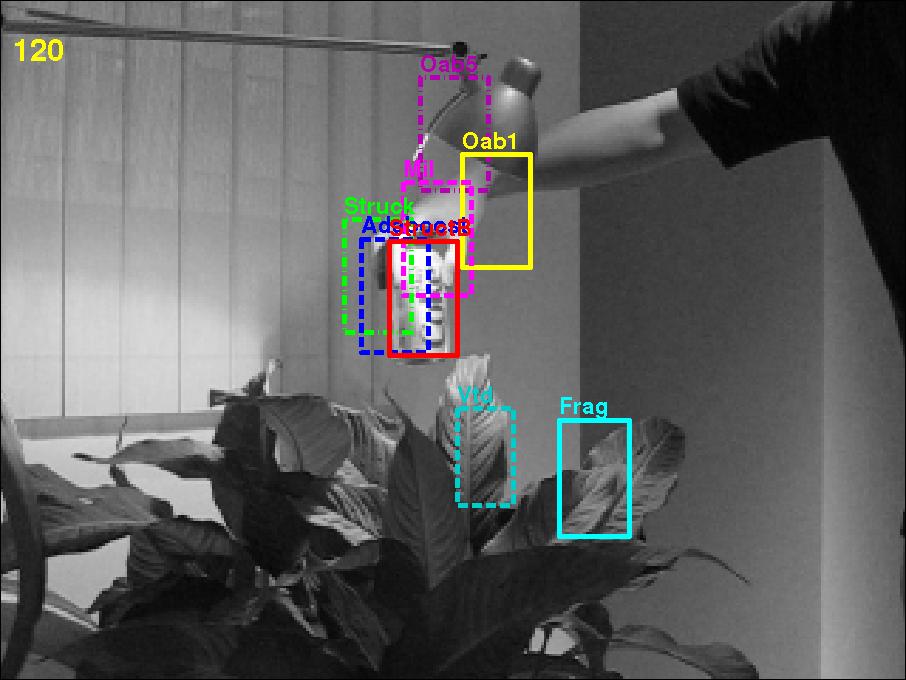}\vspace{2pt}

        \includegraphics[width=.177\linewidth]{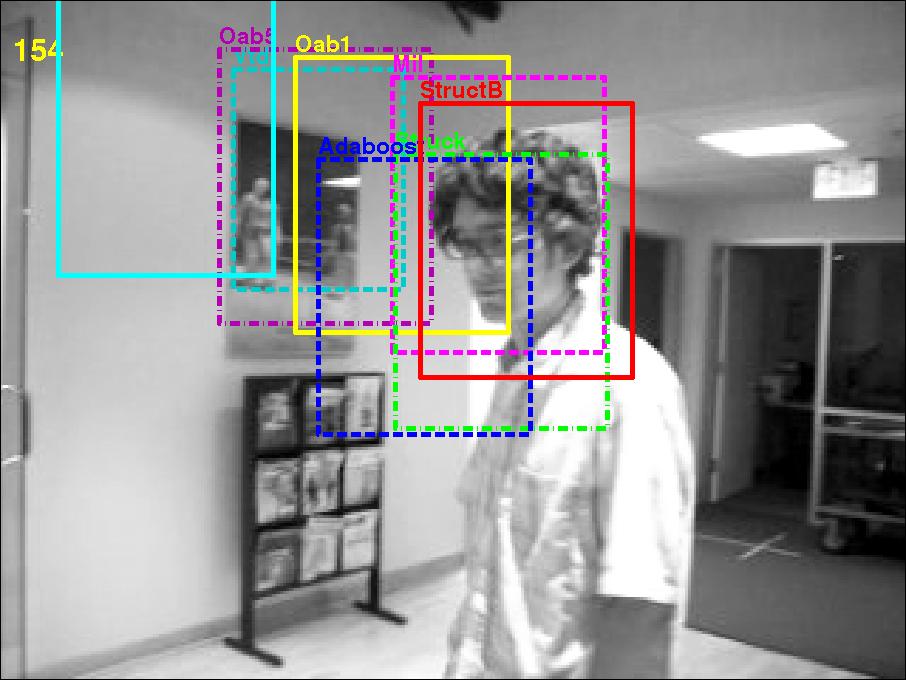}
        \includegraphics[width=.177\linewidth]{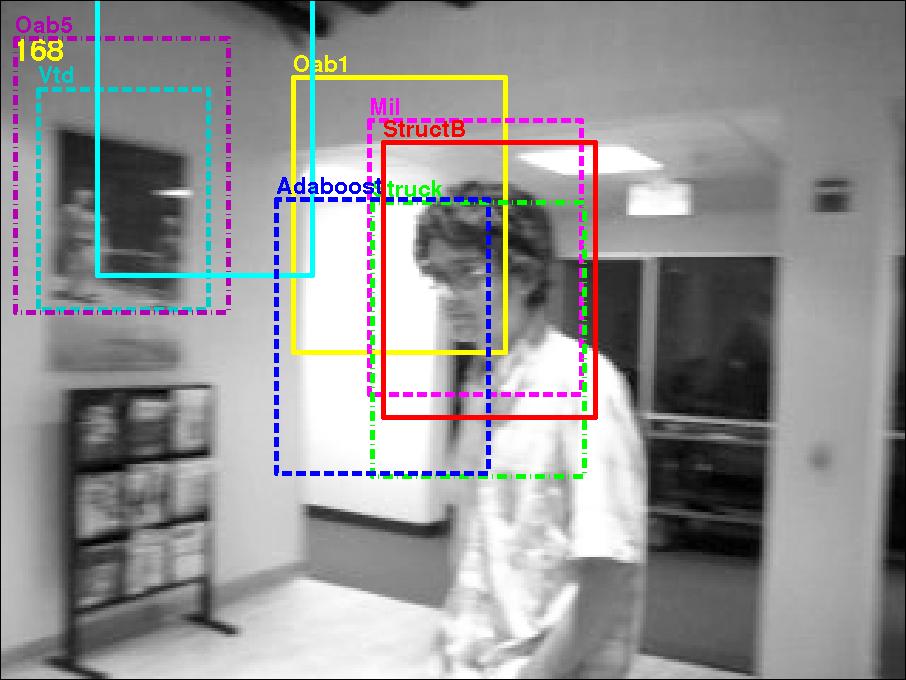}
        \includegraphics[width=.177\linewidth]{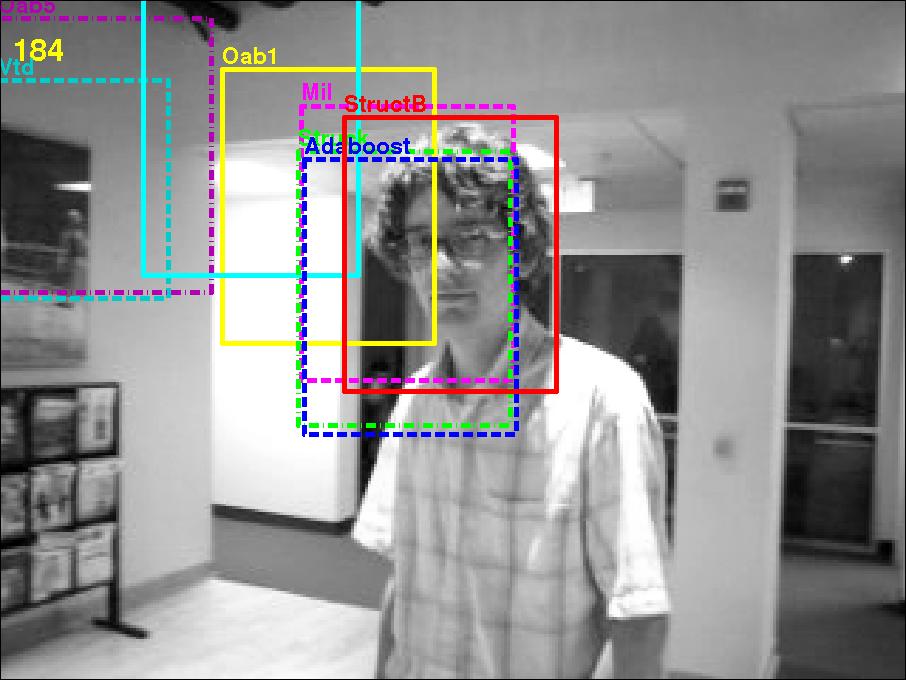}
        \includegraphics[width=.177\linewidth]{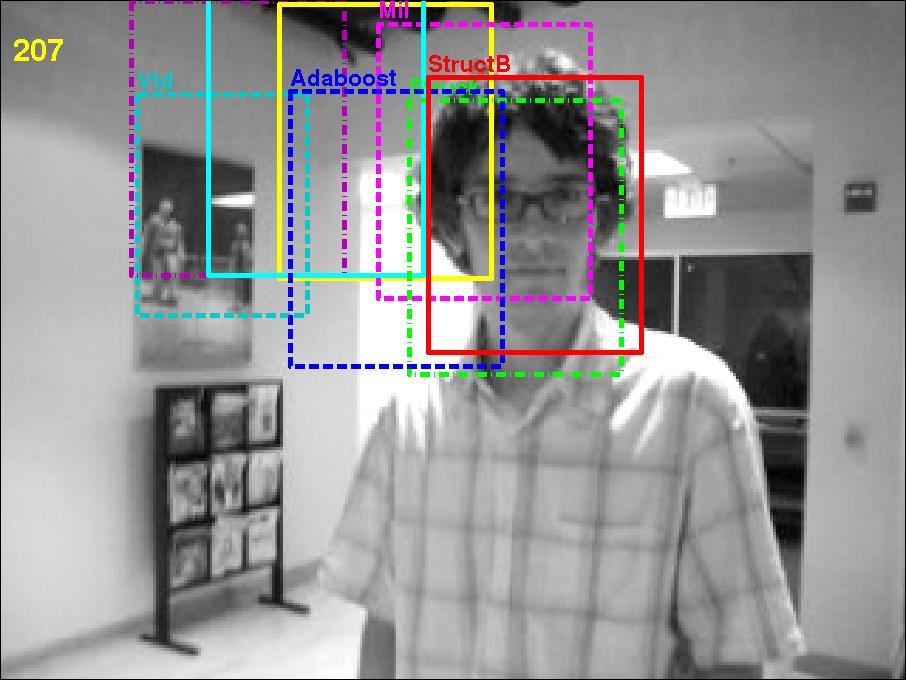}
        \includegraphics[width=.177\linewidth]{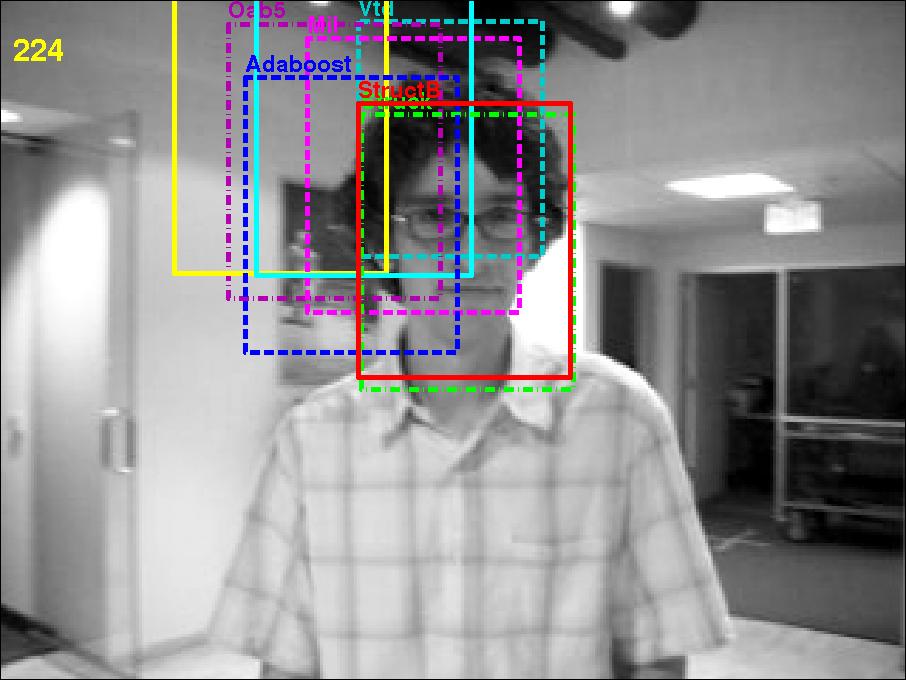}\vspace{2pt}

        \includegraphics[width=.177\linewidth]{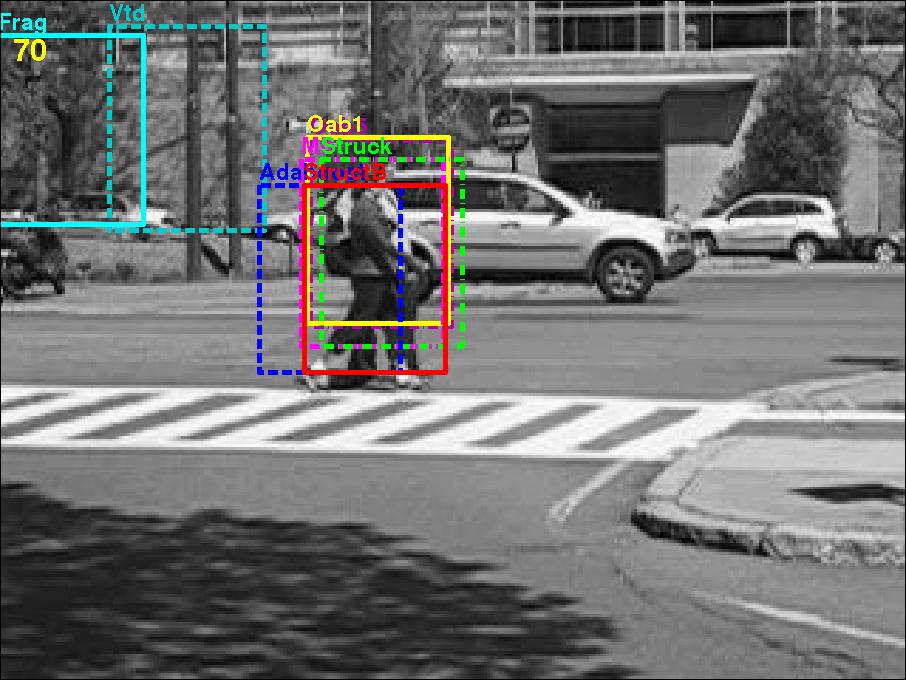}
        \includegraphics[width=.177\linewidth]{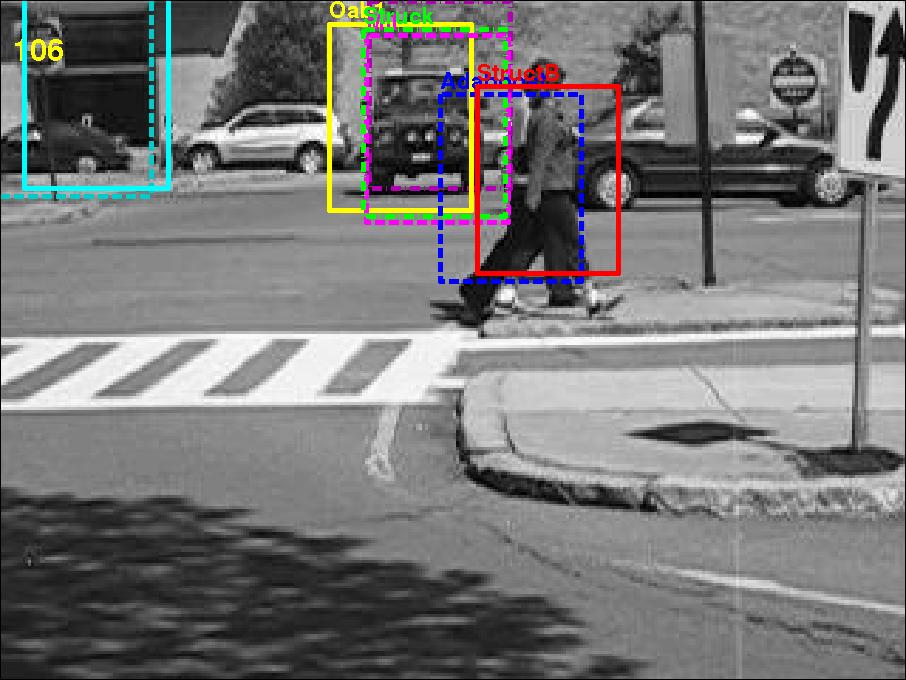}
        \includegraphics[width=.177\linewidth]{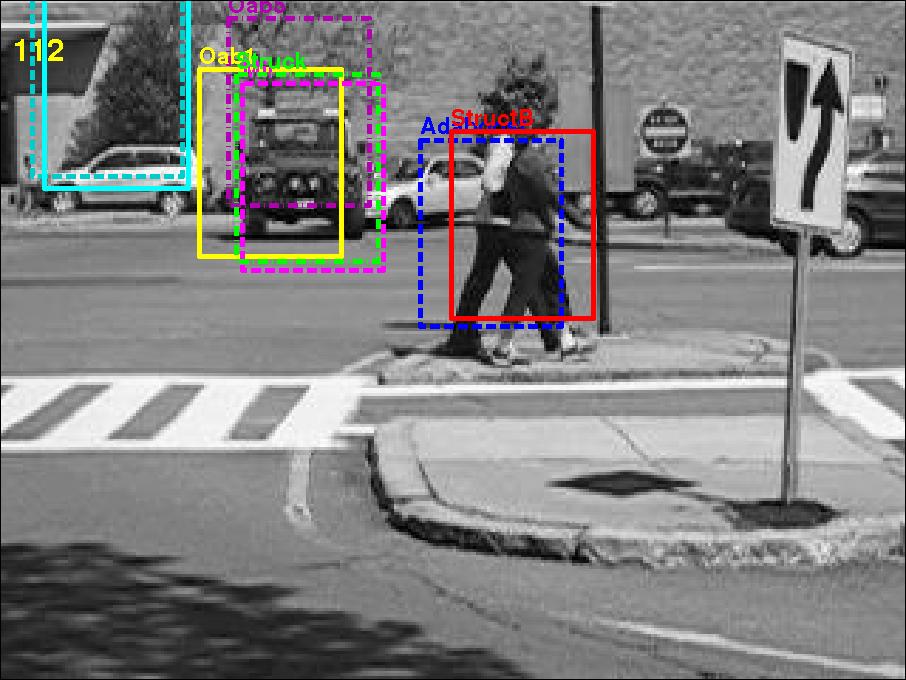}
        \includegraphics[width=.177\linewidth]{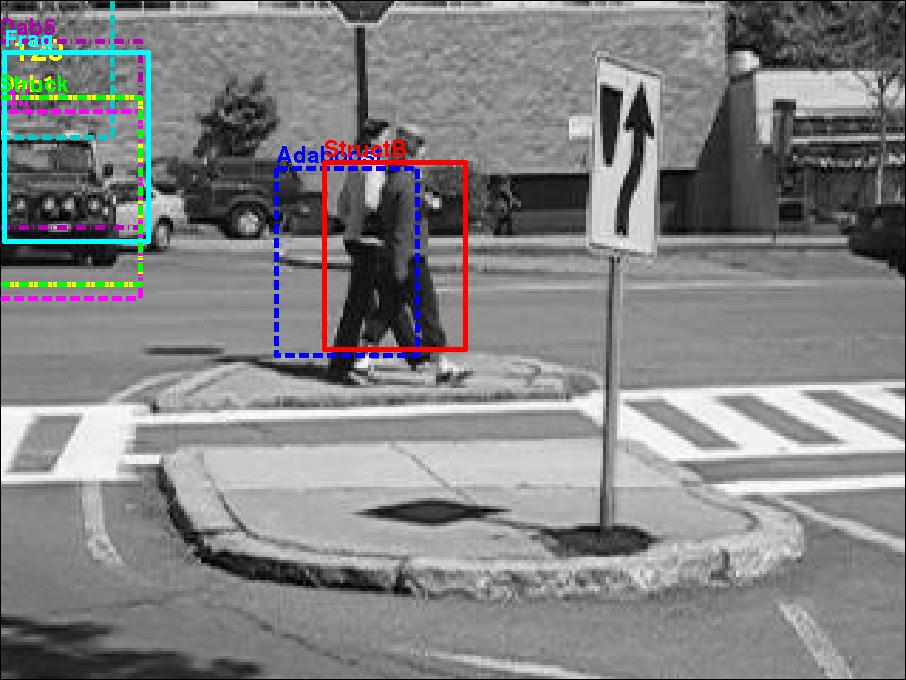}
        \includegraphics[width=.177\linewidth]{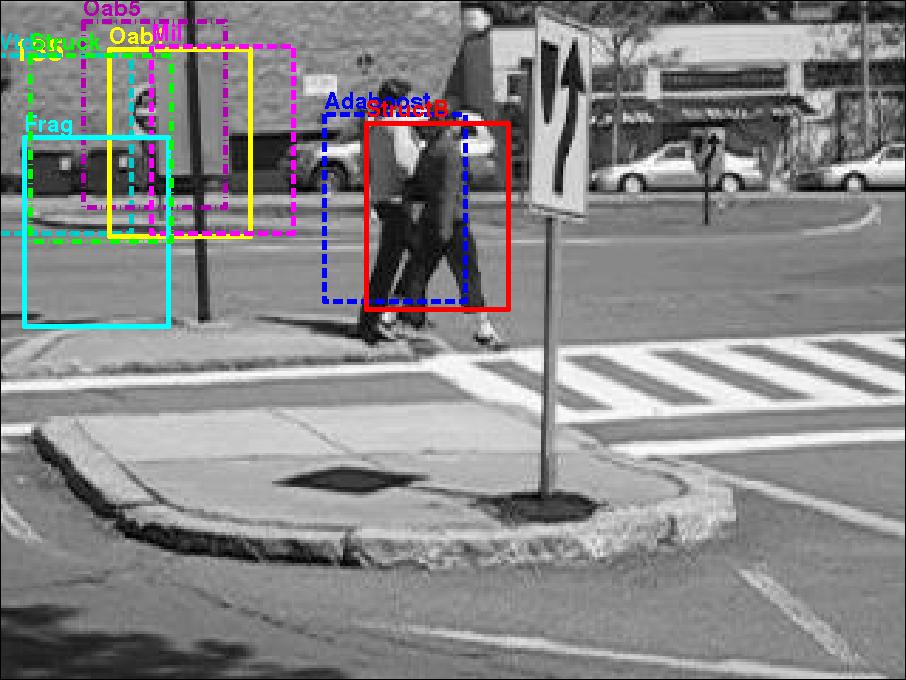}\vspace{2pt}

    \caption{Some tracking examples of several video sequences:
    ``coke", ``david", %
    and ``walk" (best viewed on screen).
    The output bounding boxes
    of our \structboost   better overlap against the ground truth than the compared methods.}
    \label{fig:track_examples}
\end{figure*}

Figure \ref{fig:frame_ovp_err} plots the Pascal overlap scores and central location errors frame by
frame on several video sequences.
Some tracking examples are shown in Figure \ref{fig:track_examples}.

\begin{table*}
  \caption{Average bounding box overlap scores on benchmark videos.
  Struck$_{50}$ \cite{Torr2011ICCV} is structured SVM tracking with a buffer size of 50.
  Our \structboost performs the best in most cases. Struck performs the second best,
  which confirms the usefulness of structured output learning.
  }
  \centering
  {
  \begin{tabular}{r | c  c  c   c  c  c   c  c}
    \hline
     &  \structboost  & AdaBoost  &  Struck$_{50}$
     & Frag & MIL &  OAB$_1$ & OAB$_5$ & VTD \\
               \hline \hline
    {coke}   & \bf 0.79  $\pm$  0.17   &0.47  $\pm$  0.19  &  0.55  $\pm$  0.18 & 0.07 $\pm$ 0.21  & 0.36 $\pm$ 0.23   & 0.10  $\pm$  0.20 & 0.04  $\pm$  0.16 & 0.10  $\pm$  0.23\\
    {tiger1}   & \bf 0.75  $\pm$  0.17   &0.64  $\pm$  0.16  &  0.68  $\pm$  0.21 & 0.21 $\pm$ 0.30  & 0.64 $\pm$ 0.18   & 0.44  $\pm$  0.23 & 0.23  $\pm$  0.24 & 0.11  $\pm$  0.24\\
    {tiger2}   & \bf 0.74  $\pm$  0.18   &0.46  $\pm$  0.18  &  0.59  $\pm$  0.19 & 0.16 $\pm$ 0.24  & 0.63 $\pm$ 0.14   & 0.35  $\pm$  0.23 & 0.18  $\pm$  0.19 & 0.19  $\pm$  0.22\\
    {david}   & \bf 0.86  $\pm$  0.07   &0.34  $\pm$  0.23  &  0.82  $\pm$  0.11 & 0.18 $\pm$ 0.24 & 0.59 $\pm$ 0.13  & 0.28 $\pm$ 0.23   & 0.21 $\pm$ 0.22 & 0.29  $\pm$  0.27\\
    {girl}   & 0.74  $\pm$  0.12   &0.41  $\pm$  0.26  & \bf  0.80  $\pm$  0.10 & 0.65 $\pm$ 0.19 & 0.56 $\pm$ 0.21  & 0.43 $\pm$ 0.18   & 0.28 $\pm$ 0.26 & 0.63  $\pm$  0.12\\
    {sylv}   & 0.66  $\pm$  0.16   &0.52  $\pm$  0.18  & \bf 0.69  $\pm$  0.14 & 0.61 $\pm$ 0.23  & 0.66 $\pm$ 0.18   & 0.47  $\pm$  0.38 & 0.05  $\pm$  0.12 & 0.58  $\pm$  0.30\\

    {bird}   & \bf 0.79  $\pm$  0.11   &0.67  $\pm$  0.14  &  0.60  $\pm$  0.26 & 0.34 $\pm$ 0.32  & 0.58 $\pm$ 0.32   & 0.57  $\pm$  0.29 & 0.59  $\pm$  0.30 & 0.11  $\pm$  0.26\\
    {walk}   & \bf 0.74  $\pm$  0.19   &0.56  $\pm$  0.14  &  0.59  $\pm$  0.39 & 0.09 $\pm$ 0.25  & 0.51 $\pm$ 0.34   & 0.54  $\pm$  0.36 & 0.49  $\pm$  0.34 & 0.08  $\pm$  0.23\\
    {shaking}  & \bf 0.72  $\pm$  0.13   &0.49  $\pm$  0.22  &  0.08  $\pm$  0.19 & 0.33 $\pm$ 0.28  & 0.61 $\pm$ 0.26   & 0.57  $\pm$  0.28 & 0.51  $\pm$  0.21 & 0.69  $\pm$  0.14\\
    {singer}   & 0.69  $\pm$  0.10   & \bf 0.74  $\pm$  0.10  &  0.34  $\pm$  0.37 & 0.14 $\pm$ 0.30  & 0.20 $\pm$ 0.34 & 0.20 $\pm$ 0.33   & 0.07  $\pm$  0.18 & 0.50  $\pm$  0.20\\
    {iceball}   & \bf 0.58  $\pm$  0.17   &0.05  $\pm$  0.16  &  0.51  $\pm$  0.33 & 0.51 $\pm$ 0.31  & 0.35 $\pm$ 0.29 & 0.08 $\pm$ 0.23   & 0.38  $\pm$  0.30 & 0.57  $\pm$  0.29\\

    \hline
  \end{tabular}
  }
  \label{Tab:exp-overlap}
\end{table*}

\begin{table*}
  \caption{Average center errors on benchmark videos.
  Struck$_{50}$ \cite{Torr2011ICCV} is structured SVM tracking with a buffer size of 50.
  We observe similar results as in Table \ref{Tab:exp-overlap}:
  Our \structboost  outperforms others on most sequences, and Struck is the second best.
  }
  \centering
  {
  \begin{tabular}{r | c  c  c   c  c  c   c  c}
    \hline
     & \structboost  & AdaBoost  &  Struck$_{50}$
     & Frag & MIL &  OAB$_1$ & OAB$_5$ & VTD \\
               \hline \hline
    {coke}   & \bf 3.7  $\pm$  4.5   &9.3  $\pm$  4.2  &  8.3  $\pm$  5.6 & 69.5 $\pm$ 32.0  & 17.8 $\pm$ 9.6   & 34.7  $\pm$  15.5 & 68.1  $\pm$  30.3 & 46.8  $\pm$  21.8\\
    {tiger1}   & \bf 5.4  $\pm$  4.9   &7.8  $\pm$  4.4  &  7.8  $\pm$  9.9 & 39.6 $\pm$ 25.7  & 8.4 $\pm$ 5.9   &  17.8  $\pm$  16.4 & 38.9  $\pm$  31.1 & 68.8  $\pm$  36.4\\
    {tiger2}   & \bf 5.2  $\pm$  5.6   &12.7  $\pm$  6.3  &  8.7  $\pm$  6.1 & 38.5 $\pm$ 24.9  & 7.5 $\pm$ 3.6   & 20.5  $\pm$  14.9 & 38.3  $\pm$  26.9 & 38.0  $\pm$  29.6\\
    {david}   & \bf 5.2  $\pm$  2.8   &43.0  $\pm$  28.2  &  7.7  $\pm$  5.7 & 73.8 $\pm$ 36.7 & 19.6 $\pm$ 8.2  & 51.0 $\pm$ 30.9   & 64.4 $\pm$ 33.5 & 66.1  $\pm$  56.3\\
    {girl}   & 14.3  $\pm$  7.8   &47.1  $\pm$  29.5  &  \bf 10.1  $\pm$  5.5 & 23.0 $\pm$ 22.5 & 31.6 $\pm$ 28.2  & 43.3 $\pm$ 17.8   & 67.8 $\pm$ 32.5 & 18.4  $\pm$  11.4\\
    {sylv}   & 9.1  $\pm$  5.8   &14.7  $\pm$  7.8  & \bf  8.4  $\pm$  5.3 & 12.2 $\pm$ 11.8  & 9.4 $\pm$ 6.5   & 32.9  $\pm$  36.5 & 76.4  $\pm$  35.4 & 21.6  $\pm$  35.7\\

    {bird}   & \bf 6.7  $\pm$  3.8   &12.7  $\pm$  9.5  &  17.9  $\pm$  13.9 & 50.0 $\pm$ 43.3  & 49.0 $\pm$ 85.3   & 47.9  $\pm$  87.7 & 48.5  $\pm$  86.3 & 143.9  $\pm$  79.3\\
    {walk}   & \bf 8.4  $\pm$  10.3   &13.5  $\pm$  5.4  &  33.9  $\pm$  49.5 & 102.8 $\pm$ 46.3  & 35.0 $\pm$ 47.5   & 35.7  $\pm$  49.2 & 38.0  $\pm$  48.7 & 100.9  $\pm$  47.1\\
	{shaking}   & \bf 9.5  $\pm$  5.4   &21.6  $\pm$  12.0  &  123.9  $\pm$  54.5 & 47.2 $\pm$ 40.6  & 37.8 $\pm$ 75.6   & 26.9  $\pm$  49.3 & 29.1  $\pm$  48.7 & 10.5  $\pm$  6.8\\
  	{singer}   & 5.8  $\pm$  2.2   & \bf 4.8  $\pm$  2.1  &  29.5  $\pm$  23.8 & 172.8 $\pm$ 95.2  & 188.3 $\pm$ 120.8   & 189.9  $\pm$  115.2 & 158.5  $\pm$  68.6 & 10.1  $\pm$  7.6\\
  	{iceball}   & \bf 8.0  $\pm$  4.1   &107.9  $\pm$  66.4  &  15.6  $\pm$  22.1 & 39.8 $\pm$ 72.9  & 61.6 $\pm$ 85.6   & 97.7  $\pm$  53.5 & 58.7  $\pm$  84.0 & 13.5  $\pm$  26.0\\
    \hline
  \end{tabular}
  }
  \label{Tab:exp-center-error}
\end{table*}

%% file: seg_exp.tex
\subsection{CRF parameter learning for image segmentation}

We evaluate our method on CRF parameter learning for image segmentation,
following the work of \cite{SzummerKH08}.
The work of \cite{SzummerKH08} applies SSVM to learn weighting parameters
for  different potentials (including multiple unary and pairwise potentials).
The goal of applying \structboost here is to learn a non-linear weighting for different potentials.
Details are described in Section \ref{sec:seg}.

We extend the super-pixels based segmentation method \cite{fulkerson08localizing} with CRF parameter learning.
The Graz-02 dataset\footnote{\url{http://www.emt.tugraz.at/~pinz/} } is used here
which contains 3 categories (bike, car and person).
Following the setting  as other methods \cite{fulkerson08localizing},
the first 300 labeled images in each category are used in the experiment.
Images with the odd indices are used for training and the rest for
testing.
We generate super-pixels using the same setting as \cite{fulkerson08localizing}.
For each super-pixel, we generate 5 types of features: visual word histogram \cite{fulkerson08localizing},
color histograms, GIST \cite{gist}, LBP\footnote{http://www.vlfeat.org/} and HOG \cite{Felzenszwalb2010}.
For constructing the visual word histogram, we follow
\cite{fulkerson08localizing} using a neighborhood size of $2$;
 the code book size is set to 200.
For GIST, LBP and HOG, we extract features from patches centered at the super-pixel with 4 increasing
sizes: $4\times4, 8\times8, 12\times12$ and $16\times16$.
The cell size for LBP and HOG is set to
a quarter of the patch size.
For GIST, we generate 512 dimensional features for each patch
by using 4 scales and the number of orientations is set to 8.
In total, we generate 14 groups of features (including features extracted on patches of different sizes).
Using these super-pixel features, we construct 14 different unary potentials ($\U=[\, U_1, \dots, U_{14} \,]^\T$)
from AdaBoost classifiers,
which are trained on the foreground and background super-pixels.
The number of boosting iterations for AdaBoost is set to $1000$.
Specifically, we define $F'(\x^{p})$ as the discriminant
function of AdaBoost on the features of the $p$-th super-pixel.
Then the unary potential function can be written as:
\begin{align}
    U(\x, y^{p})=- y^{ p } F'(\x^{p}).
\end{align}
We also construct 2 pairwise potentials ($\V=[\, V_1, V_2 \, ]^\T$):
$V_1$ is constructed using color difference, and $V_2$
using shared boundary length \cite{fulkerson08localizing} which
is able to discourage small isolated segments.
Recall that $
\Ind(\cdot, \cdot)$ is
an indicator function defined in \eqref{EQ:INDI}.
$ \Vert  \x^{p}-\x^{q}\Vert_2$ calculates the $ \ell_2$ norm
of the color difference between two super-pixels in the LUV
color-space;
$ \ell(\x^{p},\x^{q})$ is the shared boundary length between two
super-pixels.
Then $V_1, V_2$ can be written as:
\begin{align}
V_1( y^{p}, y^{q}, \x)= &
 \exp( - \Vert \x^{p}-\x^{q} \Vert_2 ) \bigl[ 1 -  \Ind(y^{ p } ,
 y^{ q }) \bigr], \\
V_2( y^{p}, y^{q}, \x)= &
 \ell(\x^{p}, \x^{q}) \bigl[ 1 -   \Ind( y^{ p } , y^{ q })
 \bigr].
\end{align}
We apply \structboost here to learn  non-linear weights for combining these potentials.
We use decision stumps
as weak learners ($\wls(\cdot)$ in \eqref{eq:seg_energy}) here.
The number of boosting iterations for \structboost is set to 50.

For comparison, we run \ssvm to learn CRF weighting parameters on exactly the same potentials as our method.
The regularization parameter $C$ in \ssvm and our \structboost is chosen from 6 candidates with
the value ranging from 0.1 to $10^3$.
We also run two simple binary super-pixel classifiers (linear SVM and AdaBoost) trained on visual word
histogram features of foreground and background super-pixels.
The regularization parameter $C$ in SVM is chosen from $10^2$ to $10^7$. The number of
boosting iterations for AdaBoost is set to $1000$.

\begin{figure*}
        \centering
        \begin{subfigure}{0.65in}
        \includegraphics[width=0.65in, height=0.6in]{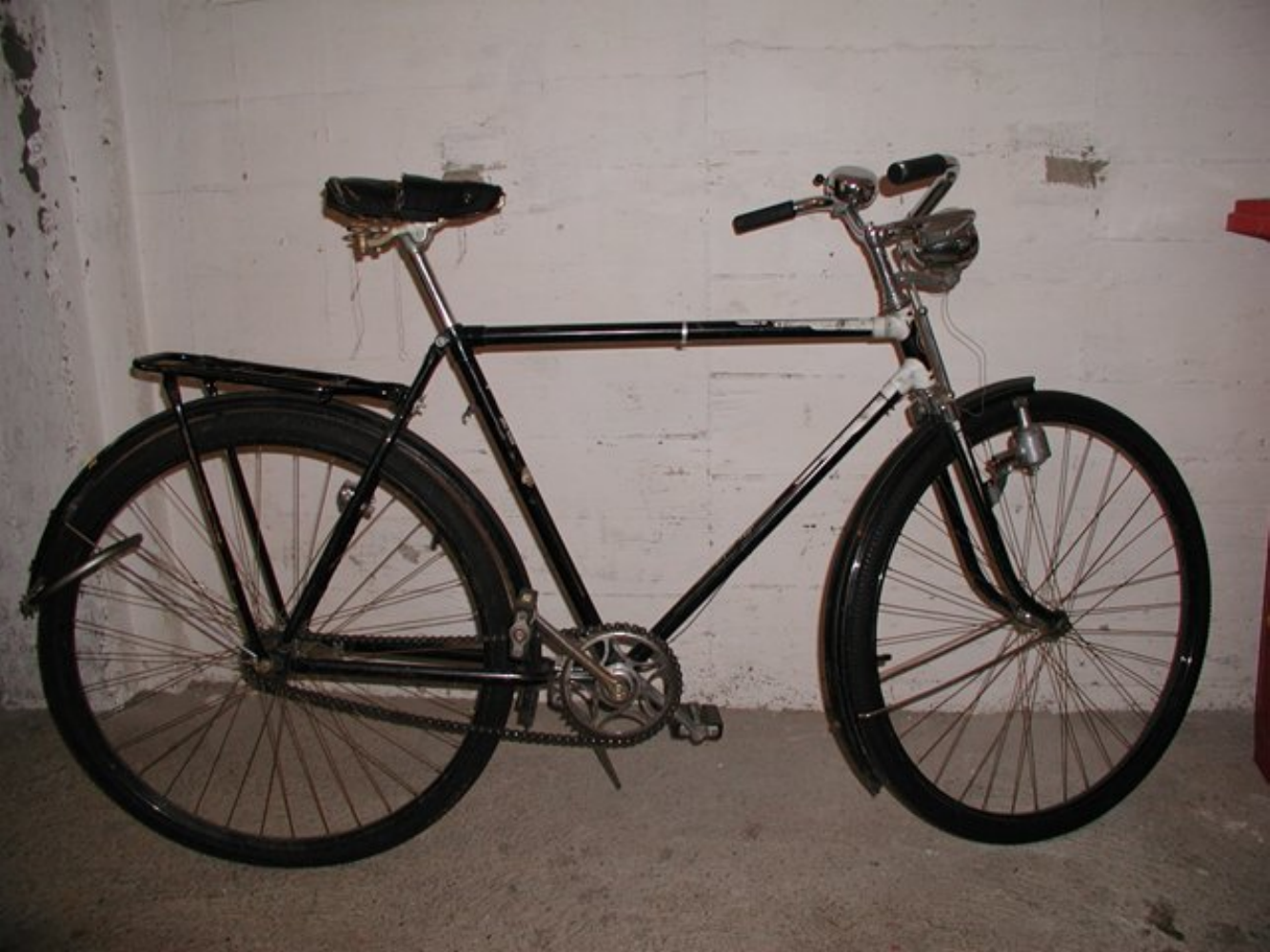}\vspace{2pt}
		\includegraphics[width=0.65in, height=0.6in]{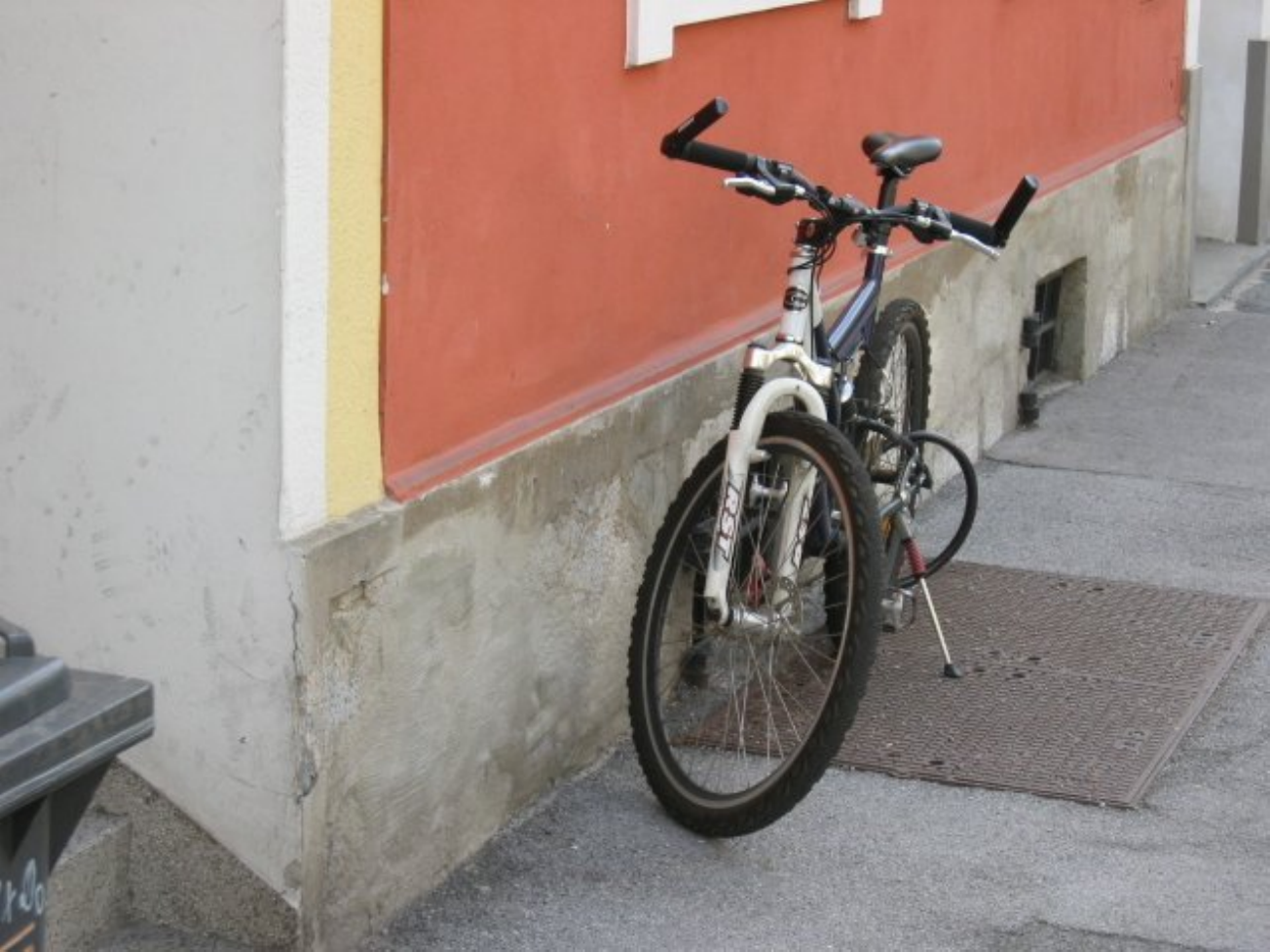}\vspace{2pt}
		\includegraphics[width=0.65in, height=0.6in]{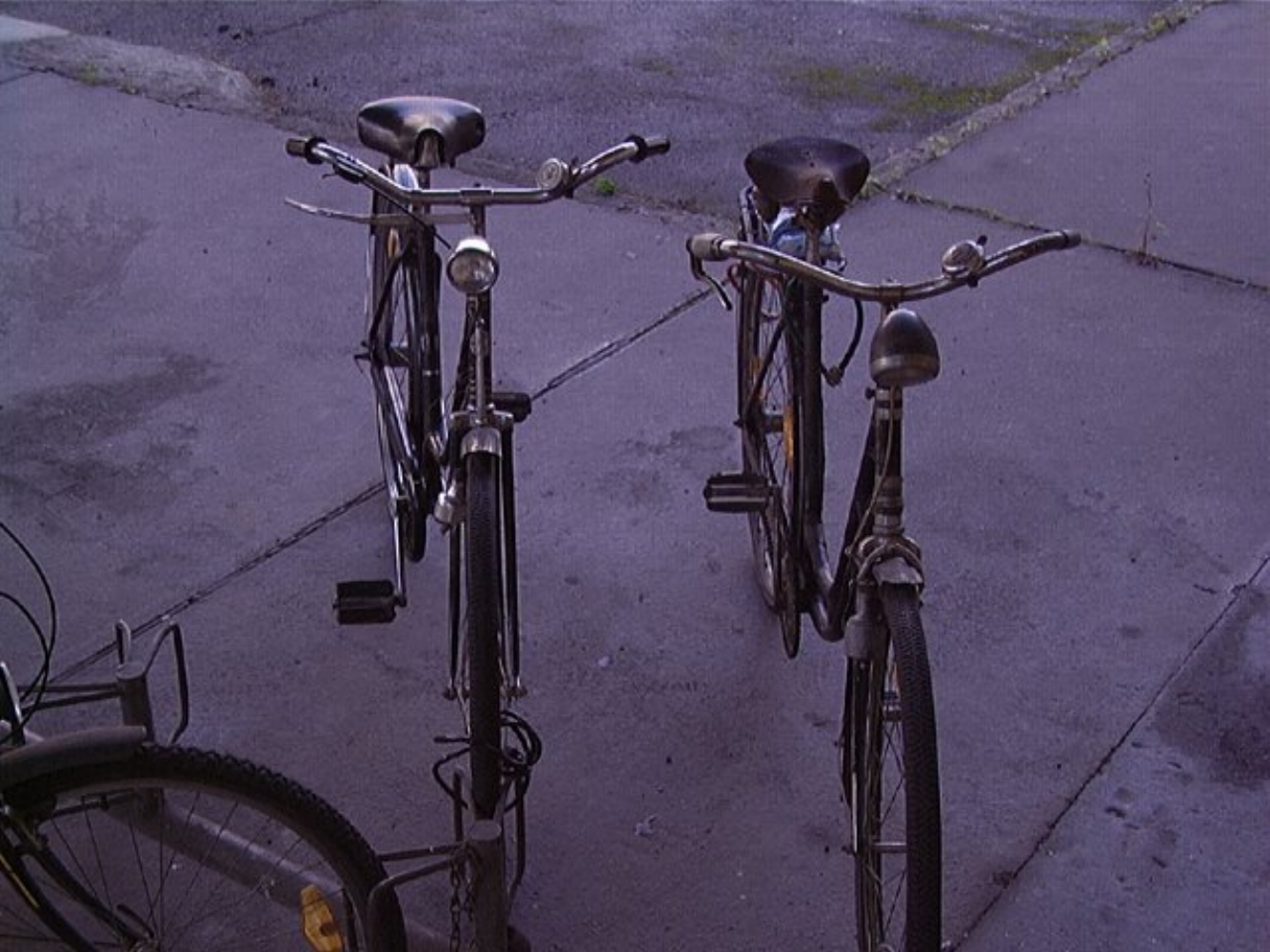}\vspace{2pt}
		\includegraphics[width=0.65in, height=0.6in]{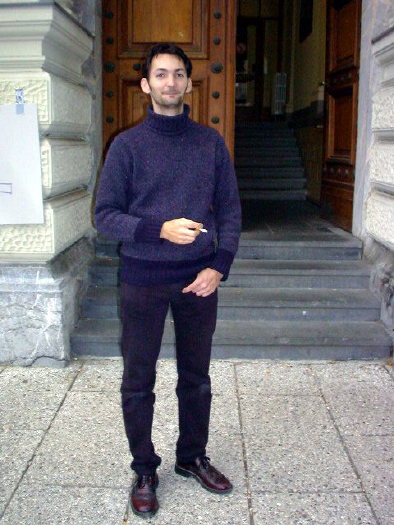}\vspace{2pt}
		\includegraphics[width=0.65in, height=0.6in]{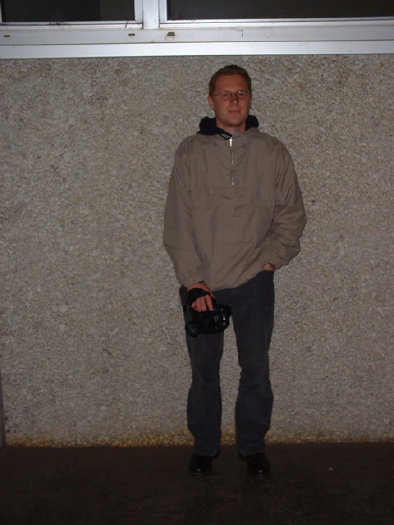}\vspace{2pt}
\caption{Testing}
        \end{subfigure}\hspace{1pt}
        \begin{subfigure}{0.65in}
        \includegraphics[width=0.65in, height=0.6in]{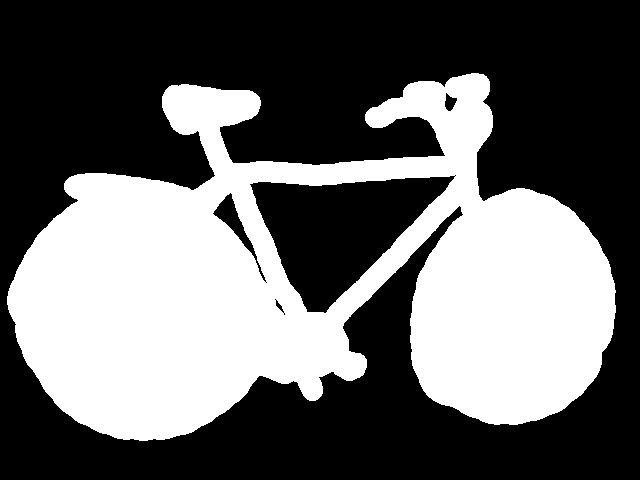}\vspace{2pt}
		\includegraphics[width=0.65in, height=0.6in]{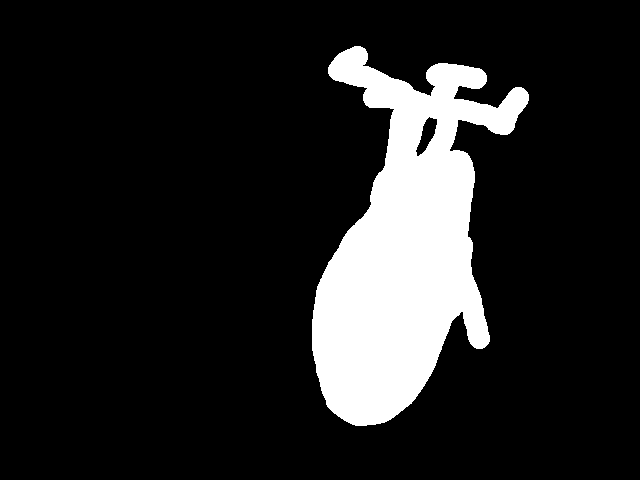}\vspace{2pt}
		\includegraphics[width=0.65in, height=0.6in]{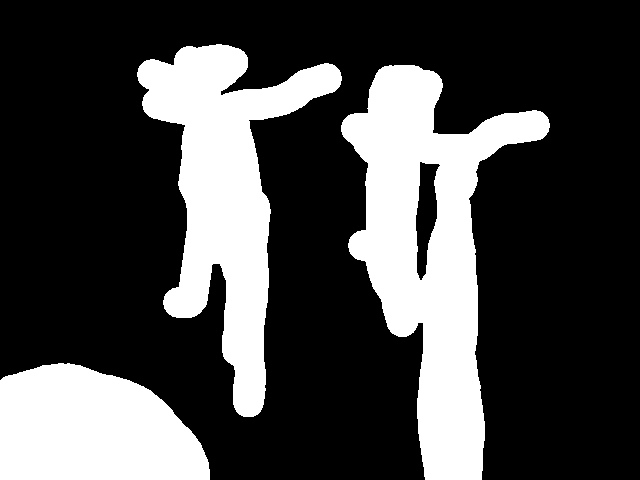}\vspace{2pt}
		\includegraphics[width=0.65in, height=0.6in]{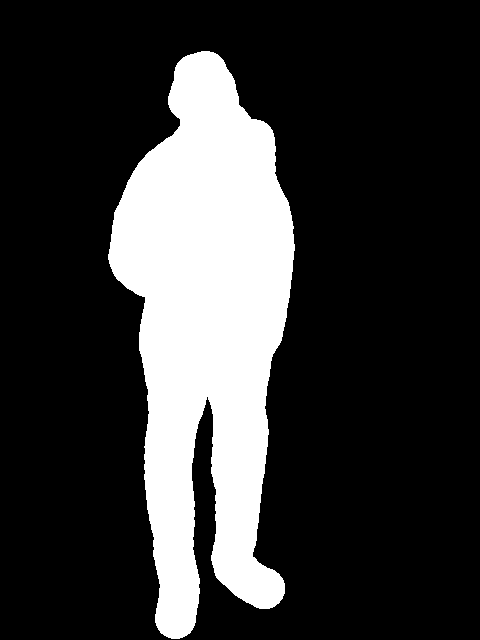}\vspace{2pt}
		\includegraphics[width=0.65in, height=0.6in]{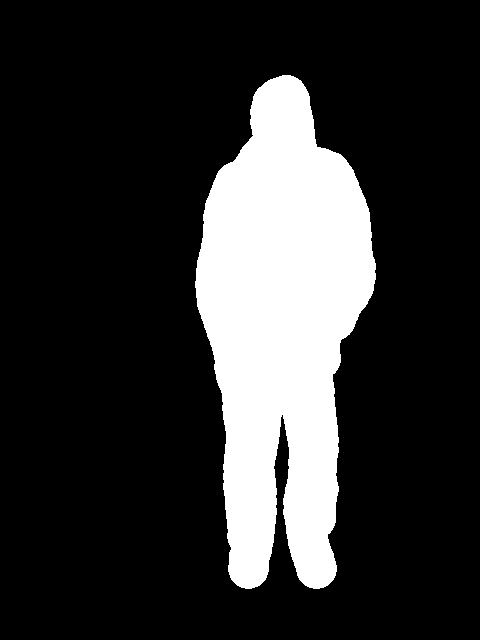}\vspace{2pt}
                \caption{Truth}
        \end{subfigure}\hspace{1pt}
        \begin{subfigure}{0.65in}
                \includegraphics[width=0.65in, height=0.6in]{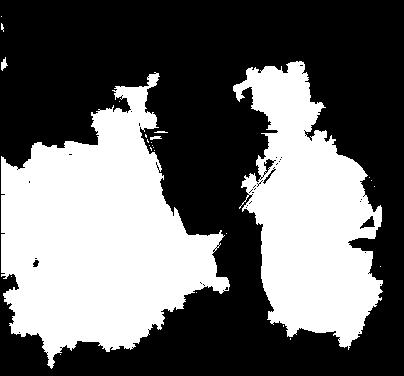}\vspace{2pt}
		\includegraphics[width=0.65in, height=0.6in]{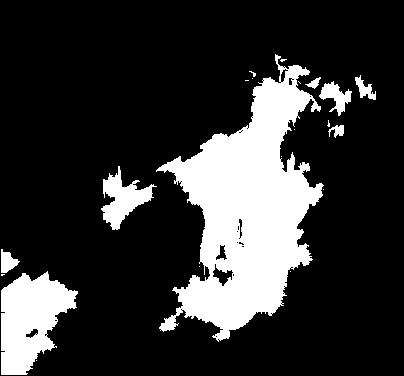}\vspace{2pt}
		\includegraphics[width=0.65in, height=0.6in]{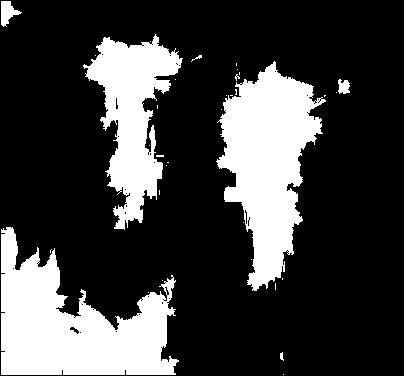}\vspace{2pt}
		\includegraphics[width=0.65in, height=0.6in]{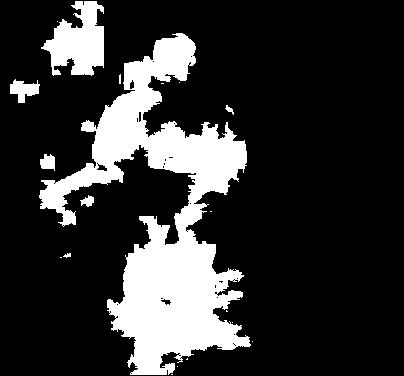}\vspace{2pt}
		\includegraphics[width=0.65in, height=0.6in]{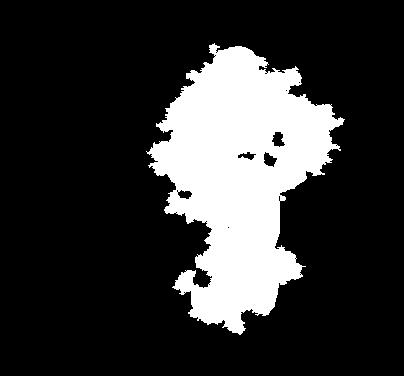}\vspace{2pt}
                \caption{AdaBoost}
        \end{subfigure}\hspace{1pt}
        \begin{subfigure}{0.65in}
                \includegraphics[width=0.65in, height=0.6in]{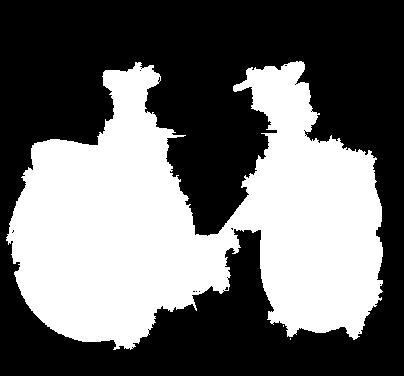}\vspace{2pt}
		\includegraphics[width=0.65in, height=0.6in]{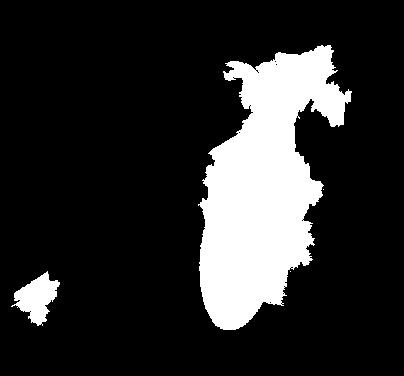}\vspace{2pt}
		\includegraphics[width=0.65in, height=0.6in]{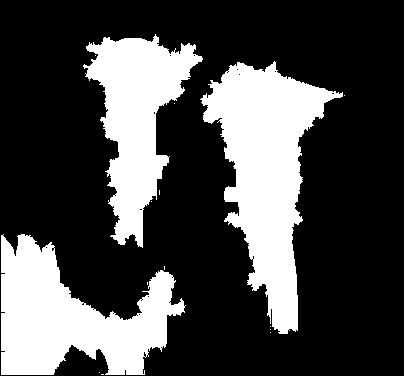}\vspace{2pt}
		\includegraphics[width=0.65in, height=0.6in]{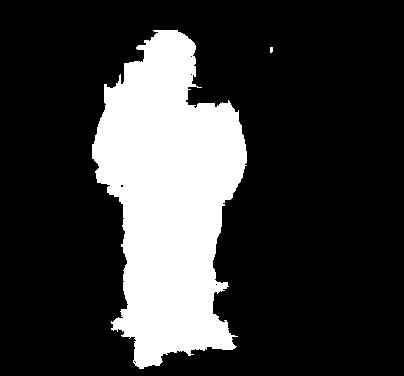}\vspace{2pt}
		\includegraphics[width=0.65in, height=0.6in]{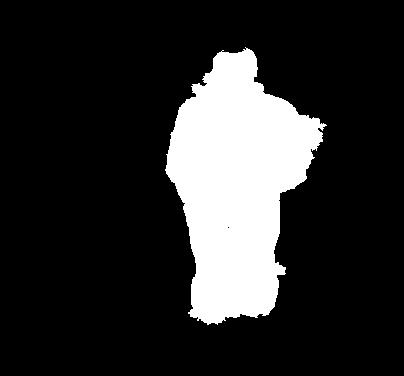}\vspace{2pt}
                \caption{SSVM}
        \end{subfigure}\hspace{1pt}
        \begin{subfigure}{0.65in}
                \includegraphics[width=0.65in, height=0.6in]{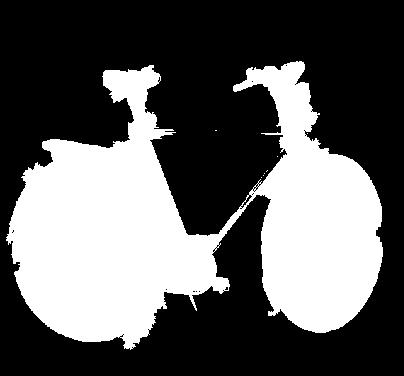}\vspace{2pt}
		\includegraphics[width=0.65in, height=0.6in]{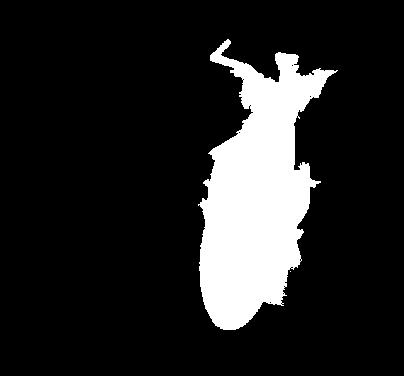}\vspace{2pt}
		\includegraphics[width=0.65in, height=0.6in]{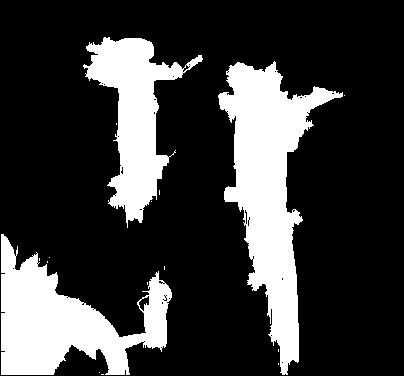}\vspace{2pt}
		\includegraphics[width=0.65in, height=0.6in]{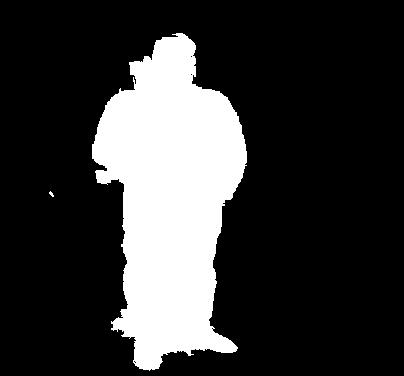}\vspace{2pt}
		\includegraphics[width=0.65in, height=0.6in]{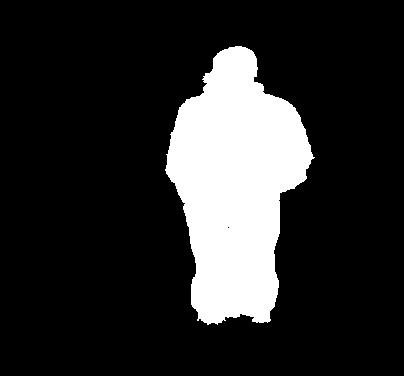}\vspace{2pt}
                \caption{StructBoost}
        \end{subfigure}\hfill\vrule\hfill
        \begin{subfigure}{0.65in}
		\includegraphics[width=0.65in, height=0.6in]{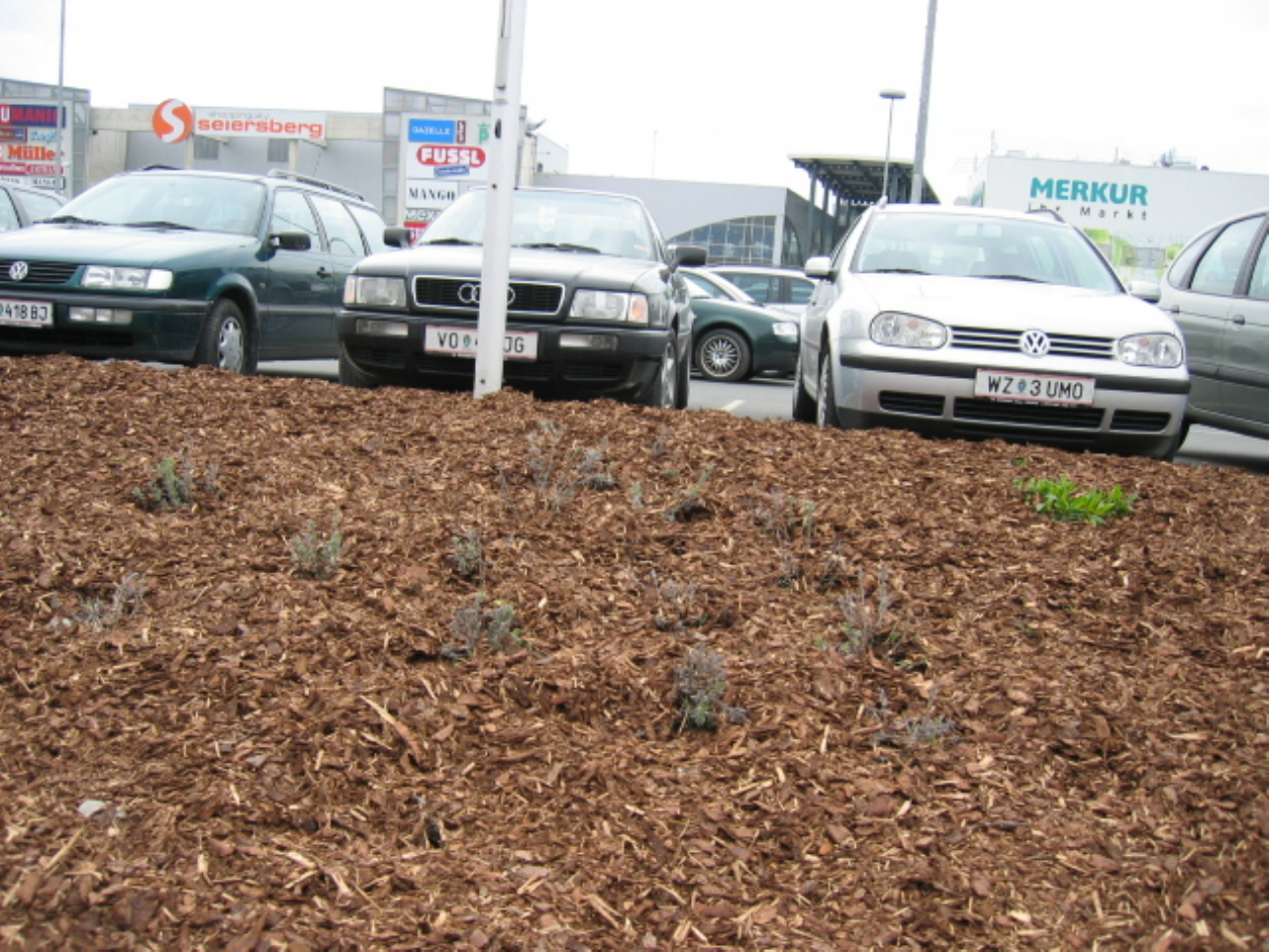}\vspace{2pt}
		\includegraphics[width=0.65in, height=0.6in]{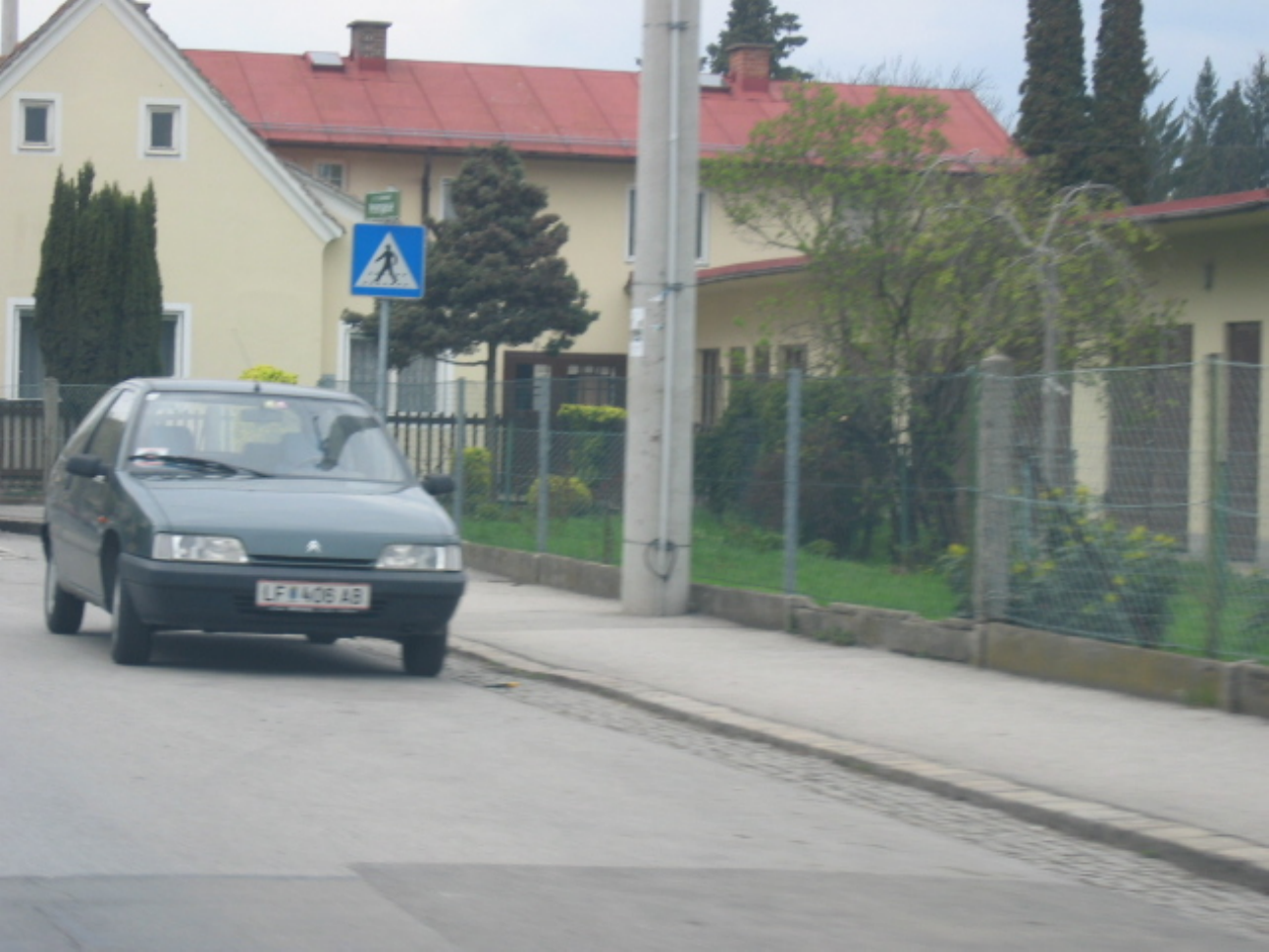}\vspace{2pt}
		\includegraphics[width=0.65in, height=0.6in]{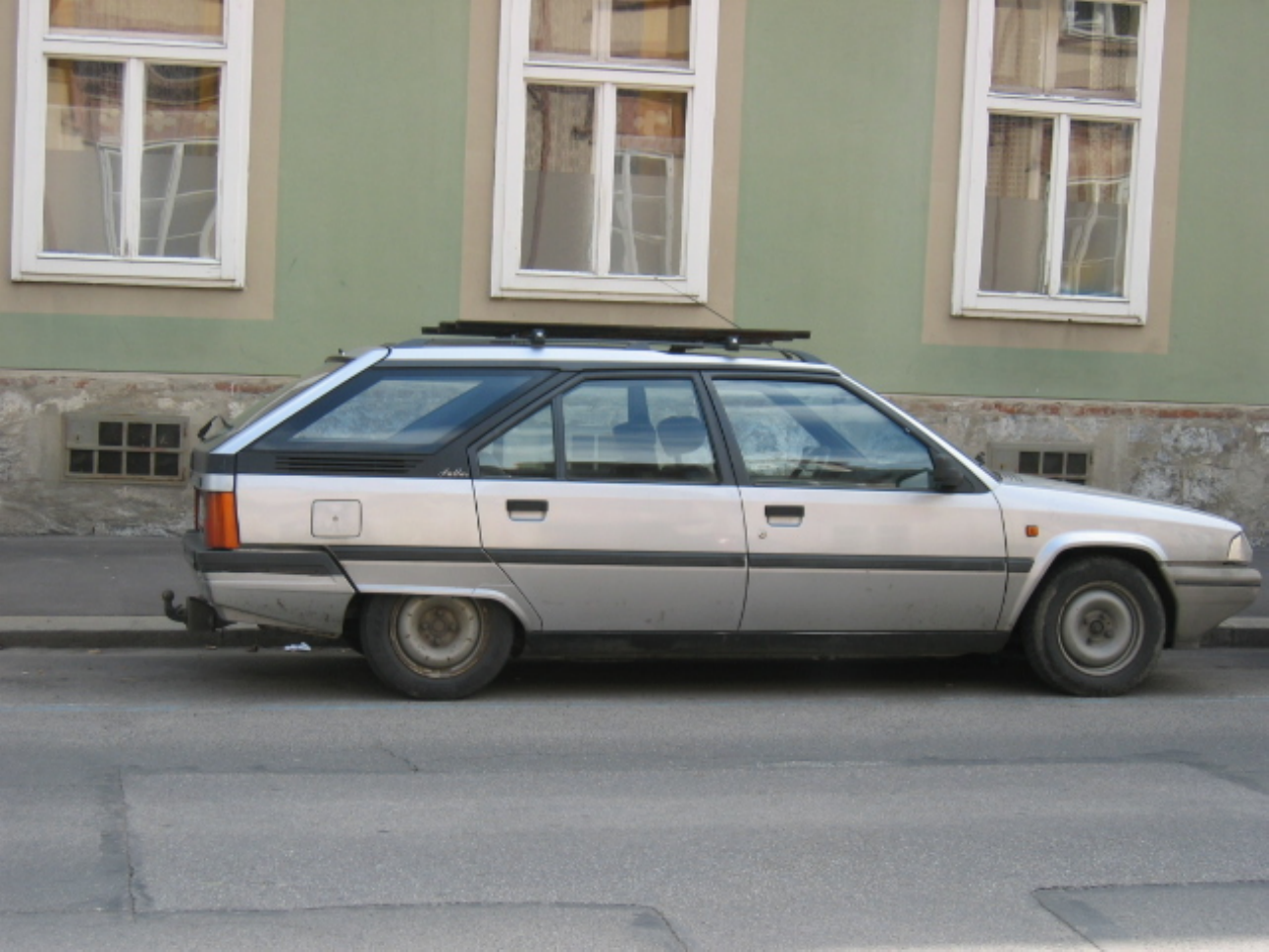}\vspace{2pt}
		\includegraphics[width=0.65in, height=0.6in]{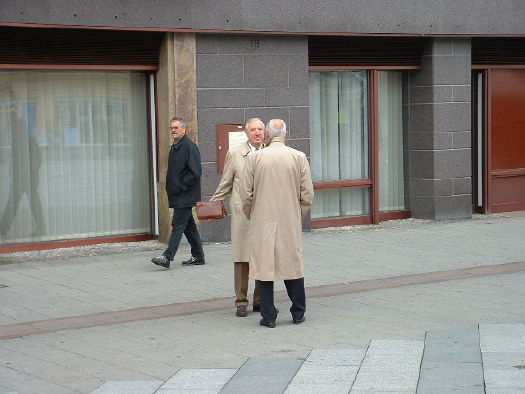}\vspace{2pt}
		\includegraphics[width=0.65in, height=0.6in]{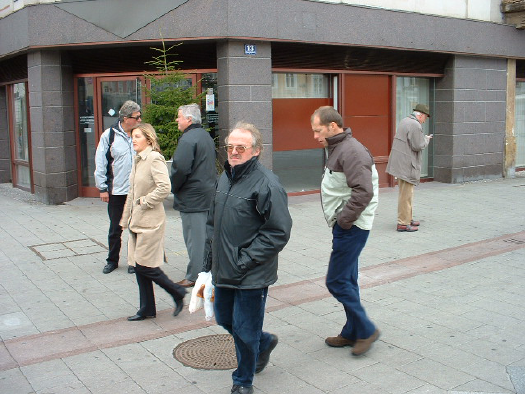}
                \caption{Testing}
        \end{subfigure}\hspace{1pt}
        \begin{subfigure}{0.65in}
		\includegraphics[width=0.65in, height=0.6in]{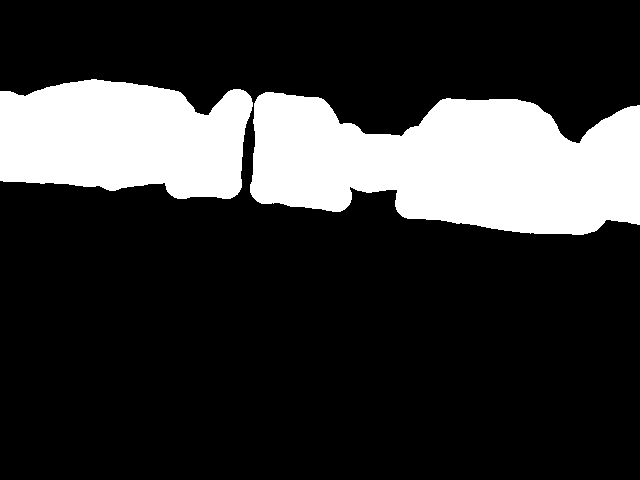}\vspace{2pt}
		\includegraphics[width=0.65in, height=0.6in]{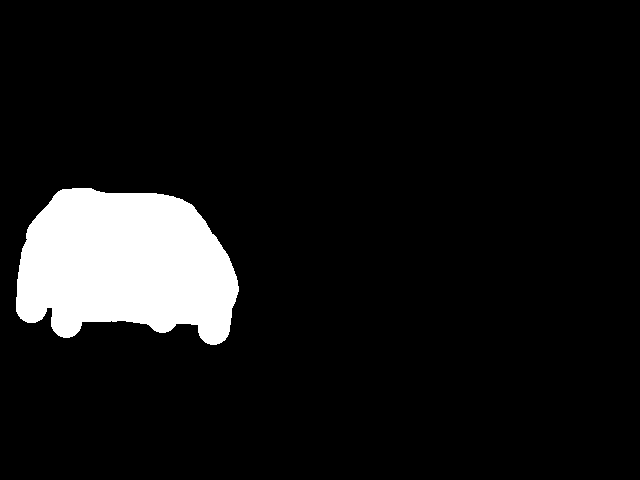}\vspace{2pt}
		\includegraphics[width=0.65in, height=0.6in]{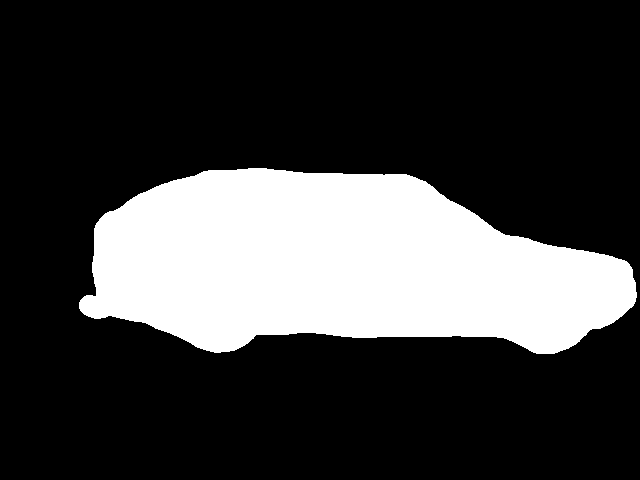}\vspace{2pt}
		\includegraphics[width=0.65in, height=0.6in]{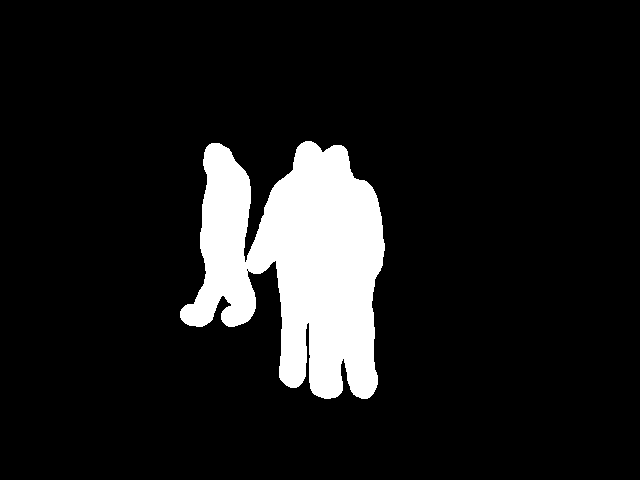}\vspace{2pt}
		\includegraphics[width=0.65in, height=0.6in]{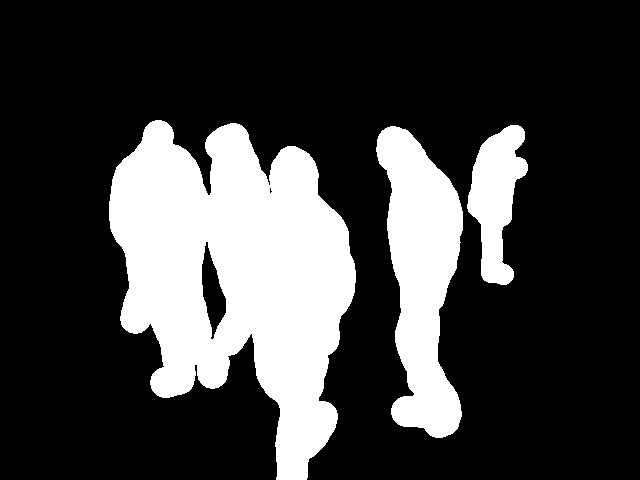}
                \caption{Truth}
        \end{subfigure}\hspace{1pt}
        \begin{subfigure}{0.65in}
		\includegraphics[width=0.65in, height=0.6in]{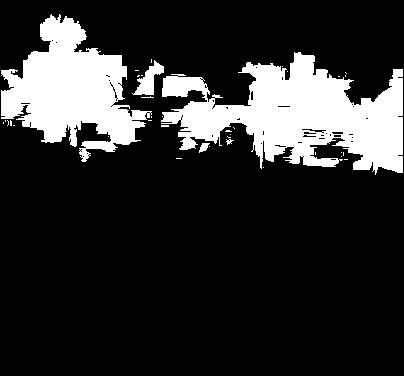}\vspace{2pt}
		\includegraphics[width=0.65in, height=0.6in]{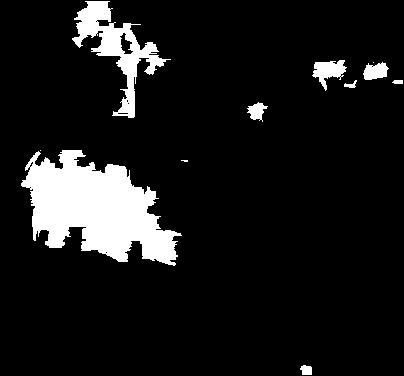}\vspace{2pt}
		\includegraphics[width=0.65in, height=0.6in]{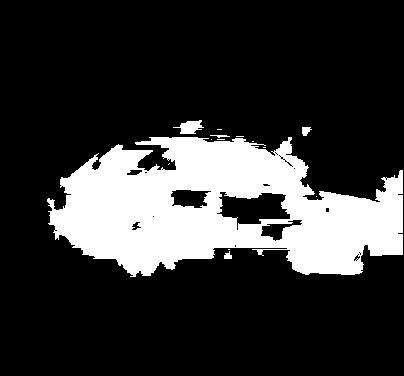}\vspace{2pt}
		\includegraphics[width=0.65in, height=0.6in]{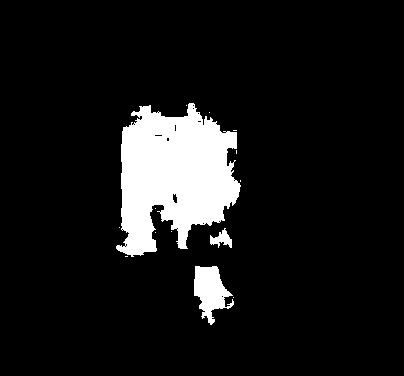}\vspace{2pt}
		\includegraphics[width=0.65in, height=0.6in]{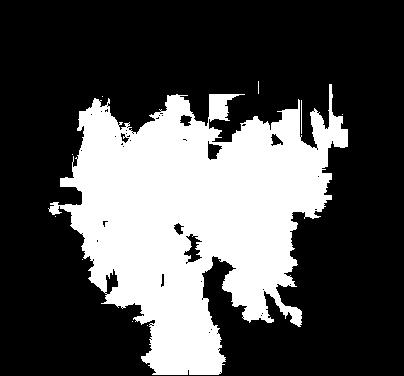}
                \caption{AdaBoost}
        \end{subfigure}\hspace{1pt}
        \begin{subfigure}{0.65in}
		\includegraphics[width=0.65in, height=0.6in]{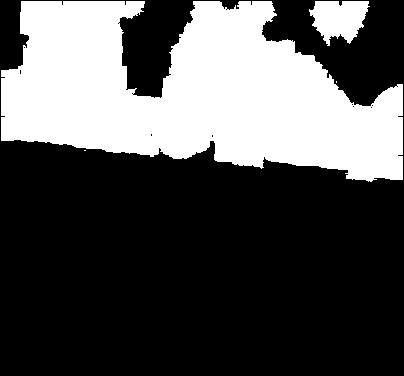}\vspace{2pt}
		\includegraphics[width=0.65in, height=0.6in]{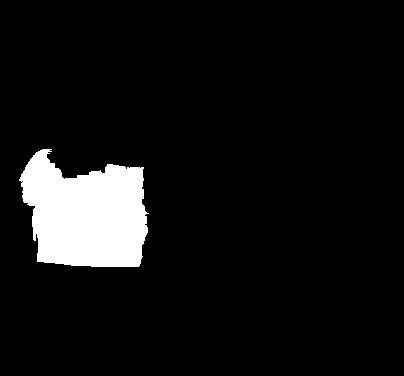}\vspace{2pt}
		\includegraphics[width=0.65in, height=0.6in]{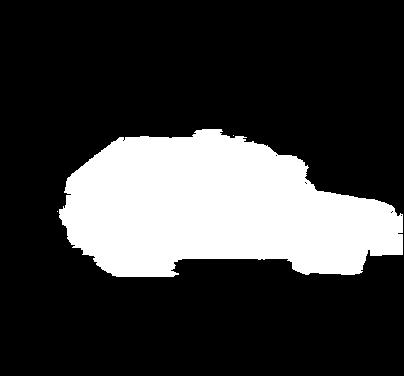}\vspace{2pt}
		\includegraphics[width=0.65in, height=0.6in]{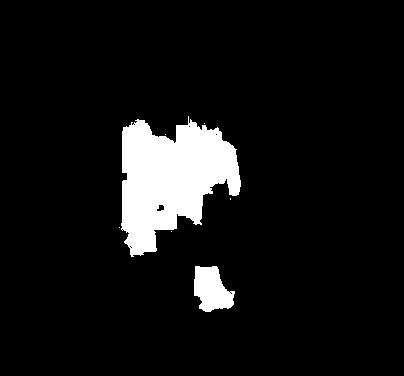}\vspace{2pt}
		\includegraphics[width=0.65in, height=0.6in]{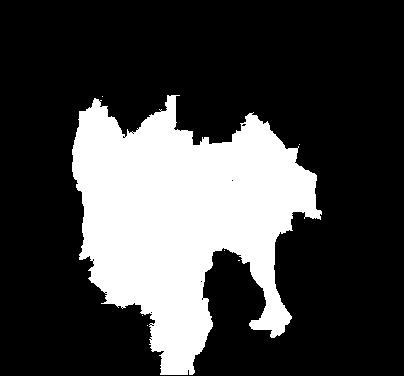}
                \caption{SSVM}
        \end{subfigure}\hspace{1pt}
        \begin{subfigure}{0.65in}
		\includegraphics[width=0.65in, height=0.6in]{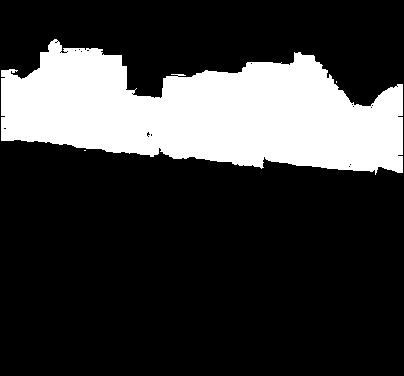}\vspace{2pt}
		\includegraphics[width=0.65in, height=0.6in]{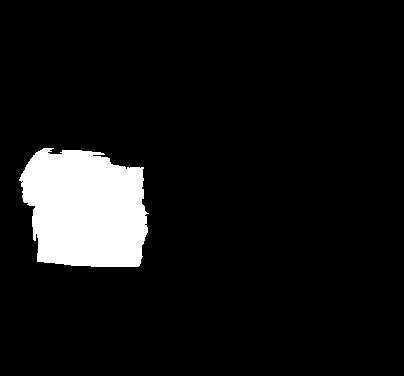}\vspace{2pt}
		\includegraphics[width=0.65in, height=0.6in]{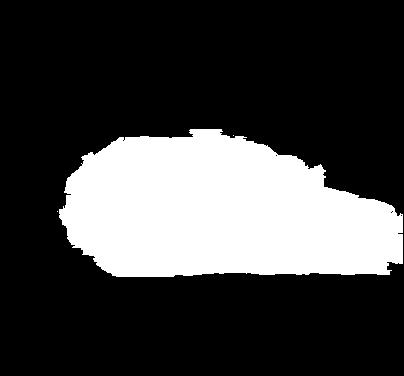}\vspace{2pt}
		\includegraphics[width=0.65in, height=0.6in]{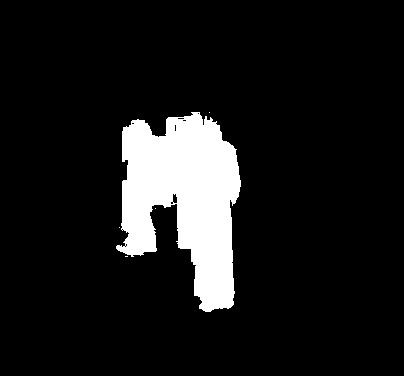}\vspace{2pt}
		\includegraphics[width=0.65in, height=0.6in]{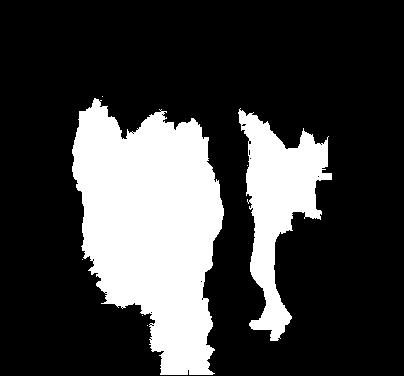}
                \caption{StructBoost}
        \end{subfigure}
	\caption{Some segmentation results on the Graz-02 dataset (bicycle, car and person).
Compared with AdaBoost, structured output learning methods (\structboost and \ssvm)
present sharper segmentation  boundaries,
and better spatial regularization. Compared with \ssvm, our \structboost with non-linear parameter
learning performs better, demonstrating more accurate foreground object boundaries
and cleaner backgrounds.}
	\label{fig:seg_examples_fix}
\end{figure*}

We use the intersection-union score, pixel accuracy (including the foreground and background) and
    $precision = recall$ value (as in \cite{fulkerson08localizing}) for evaluation.
Results are shown in Table \ref{tab:seg}. Some segmentation examples are shown in Figure \ref{fig:seg_examples_fix}.
   As shown in the results, both \structboost and \ssvm,
   which learn to combine
   different potential functions,
   are able to significantly outperform
   the simple binary models (AdaBoost and SVM).
\structboost  outperforms  \ssvm  since it learns  a non-linear combination of potentials.
Note that
SSVM learns a linear weighting for different potentials.
By employing nonlinear parameter
   learning, our method  gains further performance improvement over \ssvm.

\begin{table}
\caption{Image segmentation results on the Graz-02 dataset. The results show the
the pixel accuracy, intersection-union score (including the foreground and background) and $precision=recall$
value (as in \cite{fulkerson08localizing}). Our method \structboost for
nonlinear parameter learning performs better than SSVM and other methods.}
\centering
\resizebox{1\linewidth}{!}
  {
  \begin{tabular}{ r | c c c }
\hline
      & bike & car & people   \\ \hline
\hline
 & \multicolumn{3}{|c}{intersection/union  (foreground, background) (\%)} \\ \hline
    AdaBoost &   69.2 (57.6, 80.7) &  72.2 (51.7, 92.7) &  68.9 (51.2, 86.5)\\
    SVM &  65.2 (53.0, 77.4) & 68.6 (45.0, 92.3) & 62.9 (41.0, 84.8)\\
    \ssvm &  74.5 (64.4, 84.6) & 80.2 (64.9, 95.4) & 74.3 (58.8, 89.7) \\
    \structboost & {\bf 76.5} (66.3, 86.7) & {\bf 80.8} (66.1, 95.6) & {\bf 75.7} (61.0, 90.4) \\ \hline
\hline
 & \multicolumn{3}{|c}{pixel accuracy  (foreground, background) (\%)} \\ \hline
    AdaBoost &   84.4 (83.8, 85.1) &  82.9 (69.8, 96.0) &  81.0 (70.0, 92.1)\\
    SVM &  81.9 (81.8, 82.1) &  77.0 (57.2, 96.9) & 73.5 (53.8, 93.2)\\
    \ssvm &  {\bf 87.9} (87.9, 88.0) & 86.9 (75.8, 98.1)  & 83.5 (71.8, 95.2) \\
    \structboost & {87.4} (83.3, 91.5) & {\bf 87.6} (77.0, 98.1) & {\bf 84.6} (73.6, 95.6) \\ \hline
\hline
& \multicolumn{3}{|c}{$precision=recall$ (\%)} \\ \hline
    M.\ \& S. \cite{marszatek2007} &  61.8 & 53.8 & 44.1 \\
    F.\ et al.\ \cite{fulkerson08localizing} &  72.2 & 72.2 & 66.3 \\
    AdaBoost &  72.7 & 67.8 & 67.0 \\
    SVM &  68.3 & 63.4 & 61.2 \\
    \ssvm &77.3  &78.3  &74.4  \\
    \structboost & \bf78.9  & \bf79.3  & \bf75.9 \\ \hline
  \end{tabular}
  }
\label{tab:seg}
\end{table}

%% file: supplementary.tex
\clearpage

\onecolumn
\def\h{\psi}

\def\oneslack{$ 1 $-slack\xspace}

\def\bm{\bf}

\section{Supplementary---StructBoost: Boosting methods for predicting structured output variables
}

\subsection{Dual formulation of \mslack}
\label{sec:main}

The formulation of \structboost can be written as (\nslack primal):
\begin{subequations}
  \label{eq:structboost1}
\begin{align}
  \min_{ {\w \geq 0,\boldsymbol \xi \geq 0}}   \;\; &
        \wnorm +  {\tfrac{C}{m}} \, \one ^\T \bxi \\
  \st  \;\; &
    \w^\T \dwstructs_i(\y) \geq
    \loss ( \y_i, \y) - \xi_i, \notag \\
   & \quad\quad\quad
   \forall i=1, \dots, m;
   \text{ and }\forall \y \in  {\cal Y}. \label{EQ:1b}
\end{align}
\end{subequations}

The Lagrangian of the \mslack primal problem can be written as:
 \begin{align}
 \label{eq:la_mslack}
  L =   \;\; &
        \wnorm +  {\frac{C}{m}} \, \one ^\T \bxi -
       \sum_{i, \y} \mu_{ (  i, \y) } \cdot
        \biggl[
    \w^\T \dwstructs_i(\y) -
    \loss ( \y_i, \y) + \xi_i
        \biggr]
        -  {\boldsymbol \nu}^\T \w -  {\boldsymbol \beta}^\T \bxi,
 \end{align}
    where $ \bmu,  {\boldsymbol \nu},  {\boldsymbol
 \beta} $ are Lagrange multipliers: $ \bmu \geq 0,  {\boldsymbol
 \nu} \geq 0,  {\boldsymbol \beta} \geq 0 $. We denote by $ \mu_{
 ( i,\y ) }$ the Lagrange dual multiplier associated with the margin
 constraints \eqref{EQ:1b} for label  $ \y$
    and training pair $ (\x_i, \y_i ) $.
    At optimum, the first derivative of the Lagrangian w.r.t.\ the
    primal variables must vanish,
 \begin{align}
\frac{\partial L}{\partial \xi_i} = 0  \implies & \frac{C}{m} - \sum_{\y} \mu_{ (  i, \y) } -\beta_i = 0 \notag \\
 \implies &   0 \leq \sum_{ \y } \mu_{ (  i, \y) } \leq
 \frac{C}{m}; \notag
 \end{align}
and,
 \begin{align}
\frac{\partial L}{\partial \w} = 0  \implies & \one - \sum_{i, \y} \mu_{ (  i, \y)} \dwstructs_i( \y ) - {\boldsymbol \nu} = 0   \notag \\
 \implies &  \sum_{i, \y } \mu_{ (  i, \y) }\dwstructs_i( \y )
 \leq \one.   \notag
 \end{align}
    By putting them back into the Lagrangian \eqref{eq:la_mslack} and we can obtain
    the dual problem of the \mslack formulation in \eqref{eq:structboost1}:
    \begin{subequations}
      \label{EQ:struct-dual1}
        \begin{align}
        \max_{\bmu \geq 0}   \;\; &
        \sum_{i, \y } \mu_{ (  i, \y) } \loss( \y_i, \y )
         \\
        \st \;\; &
                      \textstyle \sum_{i, \y } \mu_{ (  i, \y)
                      }\dwstructs_i( \y )  \leq \one,
                      \\
                      &
                      0 \leq \textstyle \sum_{ \y  } \mu_{ (  i, \y)
                      } \leq
                      \frac{C}{m}, \forall i=1,\dots, m.
    \end{align}
    \end{subequations}

\subsection{Dual formulation of \oneslack}

The \oneslack formulation of \structboost can be written as:
\begin{subequations}
  \label{eq:structboost-oneslack}
\begin{align}
  \min_{\w \geq 0, \xi \geq 0}  \;\;
  &
  \wnorm + C \xi \\
  \st  \;\; &
   \tfrac{1}{m} \w^ \T
  \biggl[
             \sum_{i=1}^m c_i
            \cdot
                    \dwstructs_i ( \y )
    \biggr]
            \geq
            \frac{1}{ m }  \sum_{i=1}^m c_i\loss(\y_i,\y ) - \xi, \notag \\
    &
    \quad\quad
            \forall \c \in\{0,1\}^m;
  \forall \y \in \calY, i=1,\cdots, m.
\end{align}
\end{subequations}

The Lagrangian of the \oneslack primal problem can be written as:
 \begin{align}
  L = &  \wnorm + C \xi -  \sum_{ \c , \y }  \lambda_{ (\c, \y ) } \cdot \bigg\{   \frac{1}{m} \w^ \T
  \biggl[ \sum_{i=1}^m c_i  \cdot \dwstructs_i ( \y ) \biggr] - \notag \\ &
   \frac{1}{ m }  \sum_{i=1}^m c_i\loss(\y_i,\y ) + \xi \bigg\}
  -  {\boldsymbol \nu}^\T \w - \beta \xi,
 \end{align}
    where $\blambda, {\boldsymbol \nu}, { \beta}$ are Lagrange
    multipliers: $\blambda \geq 0, {\boldsymbol \nu} \geq 0, { \beta}
    \geq 0$.  We denote by $ \lambda_{( \c, \y)} $ the Lagrange
    multiplier associated with the inequality constraints for  $ \c
    \in \{ 0, 1 \}^m $ and label
    $ \y $.
At optimum, the first derivative of the Lagrangian w.r.t.\ the
primal variables must be zeros,
 \begin{align}
\frac{\partial L}{\partial \xi} = 0  \implies & C -  \sum_{ \c , \y }  \lambda_{ (\c, \y ) } -\beta = 0 \notag \\
 \implies &   0 \leq \sum_{\c, \y } \lambda_{ (  \c, \y) }
 \leq C; \notag
 \end{align}
and,
\begin{align}
\frac{\partial L}{\partial \w} = 0  \implies & \one - \frac{1}{m}
\sum_{ \c , \y }
    \lambda_{ (\c, \y ) } \cdot \biggl[ \sum_{i=1}^m c_i  \cdot \dwstructs_i ( \y ) \biggr]
    = {\boldsymbol \nu}.
 \notag
    \\
    \implies &
    \frac{1}{m}
    \sum_{ \c , \y }   \lambda_{ (\c, \y ) } \cdot
    \biggl[
        \sum_{i=1}^m c_i  \cdot \dwstructs_i ( \y )
    \biggr] \leq  \one.
    \label{EQ:9}
\end{align}
%
%
%
%
The dual problem of \eqref{eq:structboost-oneslack} can be written as:
\begin{subequations}
      \label{eq:structboost-oneslack-dual}
  \begin{align}
    \max_{ \blambda \geq 0}
    \;\;
    &
    \sum_{ \c , \y }
    \lambda_{ (\c, \y ) }
    \sum_{i=1}^m c_i  \loss( \y_i, \y ) \\
   \st \;\;
   &  \frac{1}{m} \sum_{ \c, \y } \lambda_{ (\c, \y) }
     \biggl[
          \sum_{i=1}^m c_i \cdot  \dwstructs_i( \y )
     \biggr]
         \leq \ones, \\
      &  0 \leq \textstyle \sum_{ \c, \y } \lambda_{ (\c, \y ) } \leq C.
\end{align}
\end{subequations}

\subsection{Convergence analysis of \structboost}

   The following result shows the convergence property of Algorithm 1.
   \begin{proposition}
       Algorithm 1 makes progress at each column
       generation iteration;
       i.e., the objective value decreases  at each iteration.
   \end{proposition}
   \begin{proof}
       Let us assume that the current solution is a finite subset of
       weak learners and their corresponding coefficients are
       $\w$. When we add a  weak learner that is not in the current
       subset and resolve the problem and the corresponding $ \hat w
       $  is zero, then the objective value and the solution keep
       unchanged. In this case, we can draw a conclusion that the
       current selected weak learner and the solution $ \w $ are
       optimal.

       Now let us assume that the optimality condition is violated.
       We want to show that we can find a weak learner $ {\hat \psi}( \cdot, \cdot )  $
       that is not in the current set of weak learners, such that
       its corresponding coefficient  $ \hat w > 0 $ holds.
       Assume that $ {\hat \h}( \cdot, \cdot )  $ is found by solving
       %
       %
       the weak learner generation subproblem
       and  the convergence condition
       $
       \frac{1}{m} \sum_{ \c, \y \neq \y_i} \lambda_{ (\c, \y) }
       \bigl[
           \sum_{i=1}^m c_i \cdot  \delta {\hat \h  }_i(\y)
       \bigr]  \leq 1
        $
        does not hold. In other words, we have
        $
       \frac{1}{m} \sum_{ \c, \y \neq \y_i} \lambda_{ (\c, \y) }
       \bigl[
           \sum_{i=1}^m  c_i \cdot  \delta {\hat \h  }_i(\y)
       \bigr]  >  1.
       $

       Now if this $ {\hat \h}( \cdot, \cdot )  $ is added into the
       master problem and the primal solution is not changed; i.e.,
       $  \hat w = 0  $, then we know that in
       \eqref{EQ:9}, $ \nu = 1 - \frac{1}{m} \sum_{ \c, \y \neq \y_i} \lambda_{ (\c, \y) }
       \bigl[
           \sum_{i=1}^m c_i \cdot  \delta {\hat \h  }_i(\y)
       \bigr] < 0 $. This contradicts the fact that the Lagrange
       multiplier $ \nu $ must be nonnegative.

       Therefore, after this weak learner is added into the master
       problem, its corresponding coefficient $ \hat w $ must be a
       non-zero positive value. It means that one more free variable
       is added into the master problem and re-solving the it must
       reduce the objective value. That means, a strict decrease in
       the objective is assured. Hence Algorithm 1 makes
       progress at each iteration.
   \end{proof}

At the $t$-th column generation iteration,
the objective of the primal in \eqref{eq:structboost1} can be written as:
\begin{align}
\label{eq:primal_t}
f(\w^{(t)})=\sum_{j=1}^t w_j^{(t)} + \frac{C}{m} \sum_{i=1}^m
\max_{\y} \biggl\{ \loss(\y_i, \y) - \sum_{j=1}^t w_j^{(t)} \biggl[ \wstruct_j(\x_i, \y_i)
-\wstruct_j(\x_i, \y) \biggr] \biggr\}.
\end{align}
Notice that the constraints: $\bxi \geq 0$ can be removed in the optimization \eqref{eq:structboost1}
because it is implicitly enforced by the first set of constraints.

Following the analysis of the boosting algorithm in \cite{Warmuth2008},
we have the following proposition which describes the progress of reducing the objective at
each boosting (column generation) iteration.

\begin{proposition}
The decrease of objective value between boosting iterations $(t-1)$ and $t$ is lower bounded as:
\begin{align}
f \left(\w^{(t-1)} \right) &- f\left(\w^{(t)} \right) \geq
\max_{\alpha \geq 0 } \bigg\lbrace -\alpha + \frac{\alpha C}{m} \sum_{i=1}^m
\biggl[
\wstruct_t \left( \x_i, \y_i\ \right)
-\wstruct_t \left( \x_i, \y_i^{(t)*}(\alpha) \right)
\biggr] \bigg\rbrace,
\end{align}
in which,
\begin{align}
\y_i^{(t)*}(\alpha)= \argmax_\y \biggl\{ \loss(\y_i, \y) + \sum_{j=1}^{t-1} w_j^{(t-1)} \wstruct_j(\x_i, \y) + \alpha \wstruct_t(\x_i, \y) \biggr\}.
\end{align}
\end{proposition}
\begin{proof}
We define the maximization solution in \eqref{eq:primal_t} as:
\begin{align}
\y_i^{(t)*}= \argmax_\y \biggl\{ \loss(\y_i, \y) + \sum_{j=1}^t w_j^{(t)} \wstruct_j(\x_i, \y) \biggr\}.
\end{align}
The objective of the $t$-th interation in \eqref{eq:primal_t} can be written as:
\begin{align}
\label{eq:primal_t2}
f\left(\w^{(t)}, \y^{(t)*}\right)=\sum_{j=1}^t w_j^{(t)} + \frac{C}{m} \sum_{i=1}^m
\biggl\{ \loss \left( \y_i, \y_i^{(t)*} \right) - \sum_{j=1}^t w_j^{(t)} \biggl[ \wstruct_j \left( \x_i, \y_i\ \right)
-\wstruct_j \left(\x_i, \y_i^{(t)*} \right) \biggr] \biggr\}.
\end{align}

Clearly, $\y^{(t)*}$ is a sub-optimal maximazation solution for the $(t-1)$-th
iteration. The decrease of the objective value between iterations  $(t-1)$ and $t$ can be lower
bounded as:
\begin{align}
f \left(\w^{(t-1)} \right) &- f\left(\w^{(t)} \right) =
f \left( \w^{(t-1)}, \y^{(t-1)*}\right) - f \left( \w^{(t)}, \y^{(t)*}\right)
\geq
f \left( \w^{(t-1)}, \y^{(t)*}\right) - f \left( \w^{(t)}, \y^{(t)*}\right).
\end{align}

Here we construct a sub-optimal solution (denoted as  $\w^{(t)'}$) for the $t$-th iteration, which takes the following form:
\begin{align}
\label{eq:sub_sol}
\w^{(t)'} =
\left[
    \begin{array}{c}
	  \w^{(t-1)}\\
      \alpha
    \end{array}
\right],
\end{align}
in which $\alpha \geq 0$. With this sub-optimial solution, the maximization solution is:
\begin{align}
\y_i^{(t)*}(\alpha)= \argmax_\y \biggl\{ \loss(\y_i, \y) + \sum_{j=1}^{t-1} w_j^{(t-1)} \wstruct_j(\x_i, \y) + \alpha \wstruct_t(\x_i, \y) \biggr\}.
\end{align}
Then the  objective decrease between iterations $(t-1)$ and $t$ can be further lower bounded as:
\begin{align*}
\label{eq:lower_bound}
f \left( \w^{(t-1)}, \y^{(t-1)*}\right) - f \left( \w^{(t)}, \y^{(t)*}\right)
& \geq
f \left( \w^{(t-1)}, \y^{(t)*}\right) - f \left( \w^{(t)}, \y^{(t)*}\right)
\\
& \geq
f \left( \w^{(t-1)}, \y^{(t)*}(\alpha) \right) - f \left( \w^{(t)'}, \y^{(t)*}(\alpha) \right).
\end{align*}
Substituting it  into the objective function,
the left part: $f \left( \w^{(t-1)}, \y^{(t)*} (\alpha)\right)$ can be  written as:
\begin{align*}
f\left(\w^{(t-1)}, \y^{(t)*}(\alpha)\right)=\sum_{j=1}^t w_j^{(t-1)} + \frac{C}{m} \sum_{i=1}^m
\biggl\{ \loss \left( \y_i, \y_i^{(t)*} (\alpha) \right) - \sum_{j=1}^t w_j^{(t-1)} \biggl[ \wstruct_j \left( \x_i, \y_i\ \right)
-\wstruct_j \left( \x_i, \y_i^{(t)*} (\alpha) \right) \biggr] \biggr\}.
\end{align*}
With the definition in \eqref{eq:sub_sol}, the right part: $f \left( \w^{(t)'}, \y^{(t)*} (\alpha) \right)$ is written as:
\begin{align*}
f\left(\w^{(t)'}, \y^{(t)*}(\alpha) \right)
 = &\sum_{j=1}^t w_j^{(t)'}
+ \frac{C}{m} \sum_{i=1}^m
\biggl\{ \loss \left( \y_i, \y_i^{(t)*}(\alpha) \right) - \sum_{j=1}^t w_j^{(t)'} \biggl[ \wstruct_j \left(\x_i, \y_i\ \right)
-\wstruct_j \left( \x_i, \y_i^{(t)*}(\alpha) \right) \biggr] \biggr\} \\
= &\sum_{j=1}^{t-1} w_j^{(t-1)} + \alpha
+ \frac{C}{m} \sum_{i=1}^m
\biggl\{ \loss \left( \y_i, \y_i^{(t)*}(\alpha) \right) \\
& - \sum_{j=1}^{t-1} w_j^{(t-1)} \biggl[ \wstruct_j \left( \x_i, \y_i\ \right) -\wstruct_j \left( \x_i, \y_i^{(t)*}(\alpha) \right) \biggr]
 - \alpha \biggl[ \wstruct_t \left( \x_i, \y_i\ \right)
-\wstruct_t \left( \x_i, \y_i^{(t)*}(\alpha) \right) \biggr]
\biggr\} \\
= & f\left(w^{(t-1)}, \y^{(t)*}(\alpha) \right) + \alpha
- \frac{\alpha C}{m} \sum_{i=1}^m \biggl\{
\biggl[
\wstruct_t \left( \x_i, \y_i\ \right)
-\wstruct_t \left( \x_i, \y_i^{(t)*}(\alpha) \right)
\biggr] \biggr\}. \\
\end{align*}
From the above, the lower bound in
\eqref{eq:lower_bound} can be wrriten as:
\begin{align}
f \left(w^{(t-1)}, \y^{(t)*} (\alpha) \right) & - f\left(\w^{(t)'}, \y^{(t)*} (\alpha)\right) \geq
-\alpha + \frac{\alpha C}{m} \sum_{i=1}^m
\biggl[
\wstruct_t \left( \x_i, \y_i\ \right)
-\wstruct_t \left( \x_i, \y_i^{(t)*}(\alpha) \right)
\biggr].
\end{align}
Finally, the  objective decrease is lower bounded as:
\begin{align}
f \left(\w^{(t-1)} \right) &- f\left(\w^{(t)} \right) \geq
\max_{\alpha \geq 0} \bigg\lbrace -\alpha + \frac{\alpha C}{m} \sum_{i=1}^m
\biggl[
\wstruct_t \left( \x_i, \y_i\ \right)
-\wstruct_t \left( \x_i, \y_i^{(t)*}(\alpha) \right)
\biggr] \bigg\rbrace,
\end{align}
in which,
\begin{align}
\y_i^{(t)*}(\alpha)= \argmax_\y \biggl\{ \loss(\y_i, \y) + \sum_{j=1}^{t-1} w_j^{(t-1)} \wstruct_j(\x_i, \y) + \alpha \wstruct_t(\x_i, \y) \biggr\}.
\end{align}

This concludes the proof.
\end{proof}